\documentclass[10pt, letter, onecolumn]{arxiv}

\makeatletter
\renewcommand\normalsize{\linespread{1.0}\@setfontsize\normalsize{10.0}{11.0}}
\makeatother

\usepackage{dm-colors}
\definecolor{softblue}{HTML}{136783}   

\usepackage{amsmath,amssymb,amsfonts,mathtools,bm,bbm}

\usepackage{multirow}
\usepackage{array}
\usepackage{longtable}
\usepackage{booktabs}
\usepackage{colortbl}
\usepackage[table]{xcolor}
\usepackage{tabularx}
\usepackage{threeparttable}

\usepackage[section]{placeins}
\usepackage{tikz}
\usetikzlibrary{shapes.geometric, arrows, positioning}

\usepackage[most]{tcolorbox}

\usepackage[pagebackref=false,breaklinks=true,
            colorlinks=true,bookmarks=true,
            citecolor=ourdarkblue,
            urlcolor=ourdarkblue,
            linkcolor=ourdarkblue]{hyperref}
\usepackage[noabbrev,capitalize]{cleveref}

\usepackage{lineno}

\usepackage{listings}
\usepackage{soul}
\usepackage{xspace}
\usepackage{enumitem}
\usepackage{etoolbox}
\patchcmd{\thebibliography}{\itemsep\z@}{\itemsep 2pt plus 1pt minus 1pt}{}{}

\AtBeginDocument{%
  \let\origcite\cite
  \renewcommand{\cite}[1]{\mbox{\origcite{#1}}}%
}

\renewcommand{\absfont}{\normalsize}

\definecolor{gray10}{gray}{0.90}

\titleformat{\paragraph}[runin]
  {\bfseries\itshape\fontsize{10}{11}\selectfont}
  {}
  {0em}
  {#1}
  [.]
\titlespacing*{\paragraph}{0em}{.20em}{0.5em}

\titleformat{\section}
  {\bfseries\fontsize{15}{17}\selectfont}{}
  {0.45em}{#1}[]
\titleformat{name=\section,numberless}
  {\bfseries\fontsize{15}{17}\selectfont}{}
  {0.em}{#1}[]
\titleformat{\subsection}
  {\bfseries\fontsize{13}{14}\selectfont}{}
  {0.35em}{#1}[]
\titleformat{name=\subsection,numberless}
  {\bfseries\fontsize{13}{14}\selectfont}{}
  {0.em}{#1}[]

\renewcommand{\titlefont}{\linespread{1.3}\color{black}\normalfont\bfseries\fontsize{18}{20}\selectfont}

\renewcommand\Affilfont{\linespread{1.0}\normalfont\fontsize{10.0}{11.5}\selectfont}

\renewcommand{\absfont}{\color{black}\linespread{1.0}\fontsize{9.0}{10}\selectfont}
\title{GreenRFM: Learning a resource-efficient radiology vision-language foundation model via supervision-centric pre-training}

\author[1,2]{Yingtai Li$^\dagger$}
\author[3]{Shuai Ming$^\dagger$}
\author[4]{Qiuli Wang$^\dagger$}
\author[1,2]{Mingyue Zhao}
\author[3]{Hongchun Zhang}
\author[1,2]{Yuhe Tian}
\author[1,2]{Haoran Lai}
\author[1,2]{Rongsheng Wang}
\author[1,2]{Rui Zhou}
\author[1,2]{Rundong Wang}
\author[5,6]{Yujia Li}
\author[7]{Zhiyang He}
\author[7]{Xiaodong Tao}
\author[4]{Wei Chen$^\ddagger$}
\author[3]{Wei Wei$^\ddagger$}
\author[1,2,8,9,10]{Shaohua Kevin Zhou$^\ddagger$}

\makeatletter
\renewcommand\AB@affilsepx{\\[2pt]\protect\Affilfont}
\makeatother
\affil[1]{School of Biomedical Engineering, Division of Life Sciences and Medicine, University of Science and Technology of China (USTC), Hefei, Anhui, 230026, China}
\affil[2]{Center for Medical Imaging, Robotics, Analytic Computing \& Learning (MIRACLE), Suzhou Institute for Advanced Research, USTC, Suzhou, Jiangsu, 215123, China}
\affil[3]{Department of Radiology, The First Affiliated Hospital of USTC, Division of Life Sciences and Medicine, USTC, Hefei, Anhui, 230001, China}
\affil[4]{7T Magnetic Resonance Translational Medicine Research Center, Department of Radiology, The First Affiliated Hospital (Southwest Hospital) of Army Medical University, Chongqing, 400033}
\affil[5]{Key Laboratory of Intelligent Information Processing of Chinese Academy of Sciences (CAS), Institute of Computing Technology, CAS}
\affil[6]{University of Chinese Academy of Sciences}
\affil[7]{Anhui IFLYHealth Co., Ltd.}
\affil[8]{Jiangsu Provincial Key Laboratory of Multimodal Digital Twin Technology, Suzhou, Jiangsu, 215123, China}
\affil[9]{State Key Laboratory of Precision and Intelligent Chemistry, USTC, Hefei, Anhui, 230026, China}
\affil[10]{Biomedical Basic Research Center (BBRC) of Jiangsu Province, Suzhou, Jiangsu, China 215123}

\date{}

\begin{document}

\begin{abstract}

Radiology foundation models (RFMs) have largely inherited the scale-first recipe of natural-image vision--language pre-training. This recipe is difficult to deploy in 3D radiology, where training corpora are smaller, reports vary across institutions, and receiving hospitals often need local adaptation under privacy and compute constraints. We ask whether routine radiology reports can instead be converted into auditable diagnostic supervision that shapes the image encoder, text encoder, aligned space, and local-adaptation procedure. We develop GreenRFM, a supervision-centric pre-training framework organized around four empirical principles: \textit{\underline{M}ore distilled}, \textit{\underline{U}biquitous}, \textit{\underline{S}emantic-enforcing}, and \textit{\underline{T}ask-aligning} (MUST) supervision. These principles convert noisy reports into structured diagnostic signals and use them to learn discriminative unimodal encoders plus an aligned image--text space for diagnosis-centered multimodal use. GreenRFM requires 24 GPU-hours on a single 24GB GPU (lightweight variant: 6GB VRAM, 4~hours) and reaches a zero-shot CT-RATE AUC of $84.8$. Evaluations using more than $200{,}000$ volumes from six institutions and two modalities show transfer to private clinical cohorts and to musculoskeletal MRI. On a local institutional cohort, computationally feasible retraining raises macro-AUC from $70.5$ to $82.1$. The aligned space also improves hepatocellular-carcinoma microvascular-invasion prediction and trans-arterial chemoembolization response analysis over established clinical scores. These results support supervision-centric pre-training as a practical route to resource-efficient, locally adaptable, diagnosis-centered radiology vision--language representations.

\end{abstract}
\maketitle

{%
\renewcommand{\thefootnote}{\fnsymbol{footnote}}%
\setcounter{footnote}{2}%
\footnotetext{These authors contributed equally to this work.}%
\setcounter{footnote}{3}%
\footnotetext{Correspondence: Shaohua Kevin Zhou (\href{mailto:skevinzhou@ustc.edu.cn}{skevinzhou@ustc.edu.cn}), Wei Wei (\href{mailto:weiweill@ustc.edu.cn}{weiweill@ustc.edu.cn}) and Wei Chen (\href{mailto:landcw@tmmu.edu.cn}{landcw@tmmu.edu.cn}).}%
}

\section*{Introduction}

\begin{figure}[!t]
    \centering
    \includegraphics[width=\textwidth]{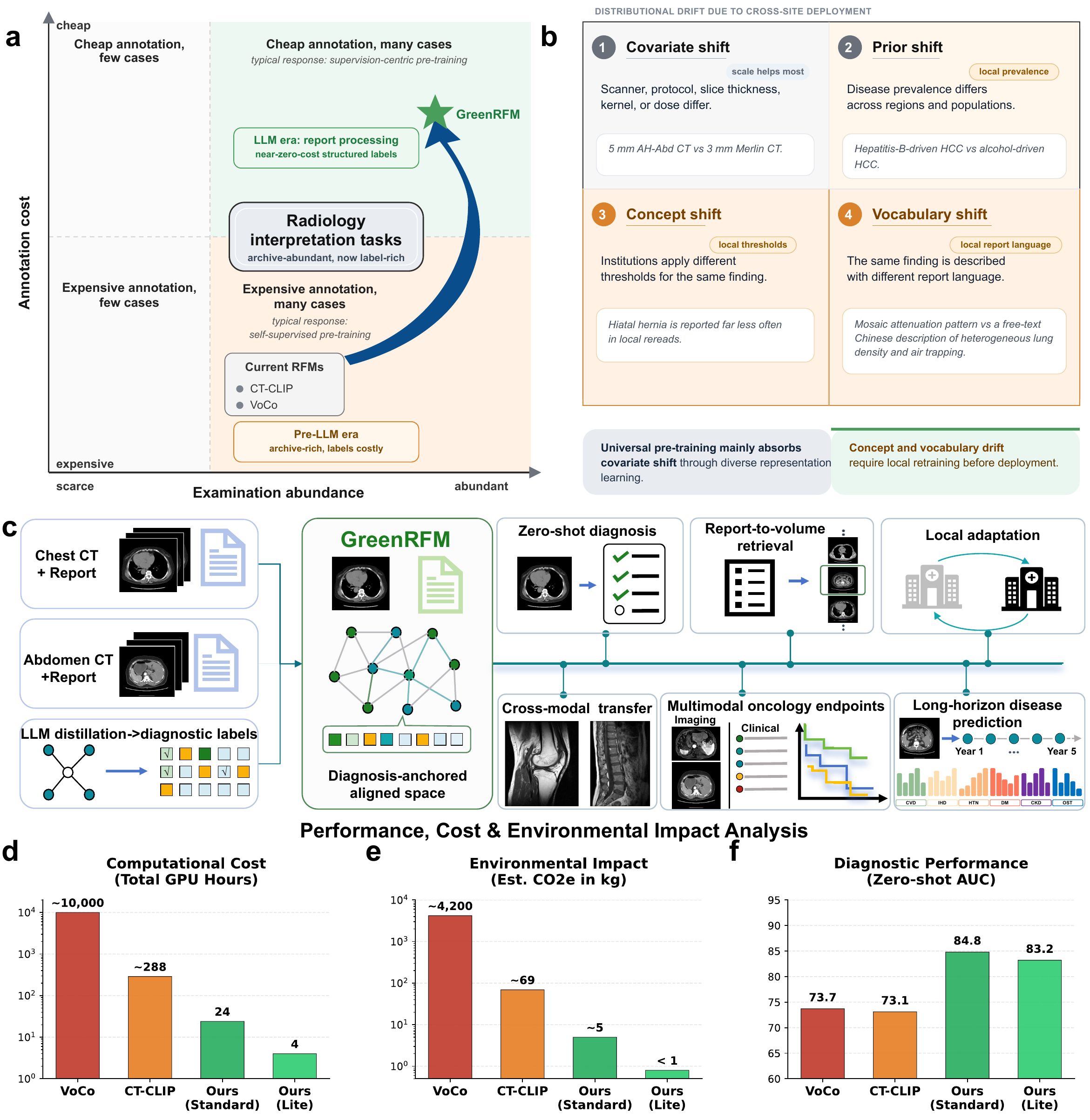}
    \caption{\textbf{From structural mismatch to deployable radiology foundation models.}
    \textbf{a} \,|\, The radiology interpretation problem is changing: for many routine findings, LLMs can convert reports into low-cost structured labels. This creates a more label-rich setting and motivates a stronger role for supervision-centric pre-training alongside self-supervised objectives.
    \textbf{b} \,|\, Deployment can be limited by data drift that scaling model size and compute alone cannot reliably resolve. Larger training runs may help with scanner, protocol and prevalence differences, but concept and vocabulary shifts reflect local diagnostic thresholds and report language, making local retraining an important route before deployment.
    \textbf{c} \,|\, GreenRFM leverages report-derived diagnostic supervision and learns a radiology vision--language representation. Such a representation is foundational so that it can be reusable, transferable, and locally adaptable to support a variety of downstream tasks. 
    \textbf{d} \,|\, Computational cost is substantially reduced relative to scaling-oriented baselines, from $\sim$10{,}000 GPU-hours for VoCo and $\sim$288 for CT-CLIP to 24 for standard GreenRFM and 4 for the lightweight variant.
    \textbf{e} \,|\, Estimated environmental impact is lower in parallel, from $\sim$4{,}200 kg CO$_2$e for VoCo and $\sim$69 kg CO$_2$e for CT-CLIP to $\sim$5 kg CO$_2$e for standard GreenRFM and $<1$ kg CO$_2$e for the lightweight variant.
    \textbf{f} \,|\, Diagnostic performance is retained and improved in this setting: GreenRFM improves zero-shot AUC over VoCo and CT-CLIP, reaching 84.8 with the standard model and 83.2 with the lightweight model versus 73.7 and 73.1 for the scaling-oriented baselines.}
    \label{fig:teaser}
\end{figure}

Volumetric imaging now anchors clinical decisions ranging from cancer staging to emergency triage, and Computed Tomography (CT) and Magnetic Resonance Imaging (MRI) provide the three-dimensional context needed for many of these decisions~\cite{ardila2019end,cao2023large,hu2025ai,wang2024screening,hermans2024rsna}. Yet CT and MRI use has outpaced radiologist capacity for interpretation~\cite{smith2012use}. Foundation models are attractive in this setting because they learn reusable representations from large, heterogeneous datasets~\cite{bommasani2021opportunities,paschali2025foundation}. For radiology, however, reuse alone is not the deployment target. A model trained at one site must also tolerate differences in scanner protocols, patient populations, and reporting conventions at another site. When local practice differs, the receiving hospital must be able to adapt the model without exporting data or using specialized large-scale infrastructure. The practical target is therefore a diagnostic representation that is broadly reusable and locally adaptable.

Radiology foundation models (RFMs) pursue this target by training on large collections of scans and reports. The resulting representations can support language-based diagnosis, semantic case retrieval, and integration with the reasoning capabilities of large language models (LLMs)~\cite{alayrac2022flamingo,li2023blip2,liu2023visual}. Recent RFMs have therefore assembled large volumetric image--report corpora for vision--language pre-training~\cite{beeche2025pan,hamamci2024ctrate,wang2025simcrop}. However, many current RFMs still transplant generic vision--language pre-training into medical imaging: paired images and reports are treated as generic cross-modal data, optimized with CLIP-style contrastive objectives~\cite{radford2021clip,zhang2022contrastive,hamamci2024ctrate}, and improved mainly by increasing model size and training compute~\cite{hamamci2024ctrate}. This scale-first route rests on two assumptions that were once reasonable but now need to be re-examined: dense structured supervision is too costly to obtain, and deployment heterogeneity can be absorbed by larger centralized training.

The first assumption was once justified. Before modern LLMs, radiology interpretation occupied an archive-abundant but label-costly regime (Fig.~\ref{fig:teaser}a): hospitals had many image--report pairs, but structured diagnostic labels remained expensive. Under that condition, contrastive and self-supervised objectives were natural responses because they extracted signal from archives without manual labels~\cite{tiu2022expert}. Modern LLMs have changed this operating regime for many routine findings. Free-text reports can now be distilled into structured supervision at low marginal cost~\cite{bigolin2025leavs}. The design question therefore shifts from how to invent proxy tasks under label scarcity to how to use abundant, imperfect, clinically grounded supervision.

The second assumption is limited by data governance and by the absence of a unified global reporting convention. Larger centralized training may improve average robustness to scanner, protocol, and prevalence differences, but clinical data cannot be pooled freely across hospitals~\cite{rieke2020future,kaissis2020secure}. Deployed radiology models also face \emph{concept} shift, such as institution-specific diagnostic thresholds, and \emph{vocabulary} shift~\cite{finlayson2021clinician,brady2018radiology} (Fig.~\ref{fig:teaser}b). For example, when CT-RATE cases labelled positive for hiatal hernia were reread by Chinese radiologists according to local reporting practice, only $33.9\%$ were regarded as reportable hiatal hernia~(Supplementary Table~\ref{tab:hiatal_hernia_reread}); similarly, CT-RATE reports may use the term ``mosaic attenuation pattern'', whereas AnHui Provincial Hospital reports may describe patchy regional lung-density differences with air trapping. These shifts encode how local communities read and report images. They cannot be fully removed by larger centralized training and instead require privacy-preserving adaptation on local archives.

These two changes define the question addressed here: once routine reports can be distilled into diagnostic labels, where should that supervision enter an RFM, and can the resulting representation remain useful after local adaptation? We introduce GreenRFM, a supervision-centric pre-training framework for building diagnosis-centered radiology vision--language representations. GreenRFM places \emph{supervision design} alongside model scale as a primary design lever and organizes that design around four testable principles, abbreviated as MUST (detailed in Methods). \textit{\underline{M}ore distilled supervision} predicts that LLM-distilled labels remain useful when residual report-distillation errors are predominantly conservative rather than hard false positives. \textit{\underline{U}biquitous supervision} predicts that direct supervision of the image encoder, text encoder, and aligned space yields positive marginal gains. \textit{\underline{S}emantic-enforcing supervision} predicts that alignment is limited by the weaker unimodal representation, motivating supervised image- and text-encoder pre-training before cross-modal alignment. \textit{\underline{T}ask-aligning supervision} predicts that reducing train--inference mismatch improves downstream performance. This study builds on our preliminary work~\cite{li2025more} and extends it into a supervision-centric RFM framework with explicit MUST principles, broader external and private validation, downstream multimodal-fusion analyses, and local-adaptation studies.

GreenRFM tests these principles in a two-stage pipeline (Fig.~\ref{fig:method_overview}). LLM-distilled labels first supervise the image and text encoders, and the encoders are then aligned while diagnostic supervision is retained. This design places the model on the efficient side of Fig.~\ref{fig:teaser}d,e: standard and lightweight GreenRFM require 24 and 4 GPU-hours, respectively, with estimated emissions of $\sim$5 and $<1$ kg CO$_2$e. The lower footprint coincides with higher accuracy in our benchmarks (Fig.~\ref{fig:teaser}f). On CT-RATE zero-shot diagnosis, standard and lightweight GreenRFM reach AUCs of 84.8 and 83.2, compared with 73.7 for VoCo and 73.1 for CT-CLIP. Across public and private CT benchmarks, GreenRFM achieves higher diagnosis and retrieval performance than the baselines evaluated here using a $33$M-parameter $3$D ResNet-18, and it matches the performance of $1.2$B-parameter VoCo using $1\%$ of the data. This low computational cost makes the local-adaptation question testable: when reporting conventions shift, hospitals can retrain GreenRFM on local data. On the held-out 20\% AH-Abd local-adaptation split, the same universal checkpoint is remeasured at macro-AUC $70.5$ and improves to $82.1$ after local fine-tuning.

GreenRFM also keeps the text branch as a downstream asset rather than using it only as training scaffolding for the vision encoder. Within this diagnosis-centered scope, the aligned text encoder provides measurable gains beyond the image backbone alone in hepatocellular-carcinoma microvascular-invasion prediction, where multimodal fusion raises AUC by $+6.4$ over the image-only model. In trans-arterial chemoembolization response analysis, GreenRFM multimodal fine-tuning improves non-complete-response prediction and yields a decision-curve integrated net benefit of $0.302$---$1.47\times$ that of the next-best clinical score. In the survival association analysis, adding the numeric 6-and-12 clinical score to GreenRFM imaging features raises overall-survival C-index from $0.717$ to $0.754$ and progression-free-survival C-index from $0.758$ to $0.815$. Together, these results support a bounded claim: supervision-centric pre-training can turn auditable report-derived labels into resource-efficient, locally adaptable, diagnosis-centered radiology vision--language representations.

\section*{Results}

We first tested whether LLM-distilled supervision is reliable and improves representation learning. We then evaluated whether these gains translate into robust zero-shot performance, transfer, and representation quality across datasets, modalities, and evaluation settings. Finally, we asked whether GreenRFM reduces deployment bottlenecks beyond cross-institutional transfer through local retraining and downstream clinical use.

\subsection*{Principled supervision design improves representation learning}
\label{sec:principled_supervision}

\begin{figure}[!t]
    \centering
    \includegraphics[width=\textwidth]{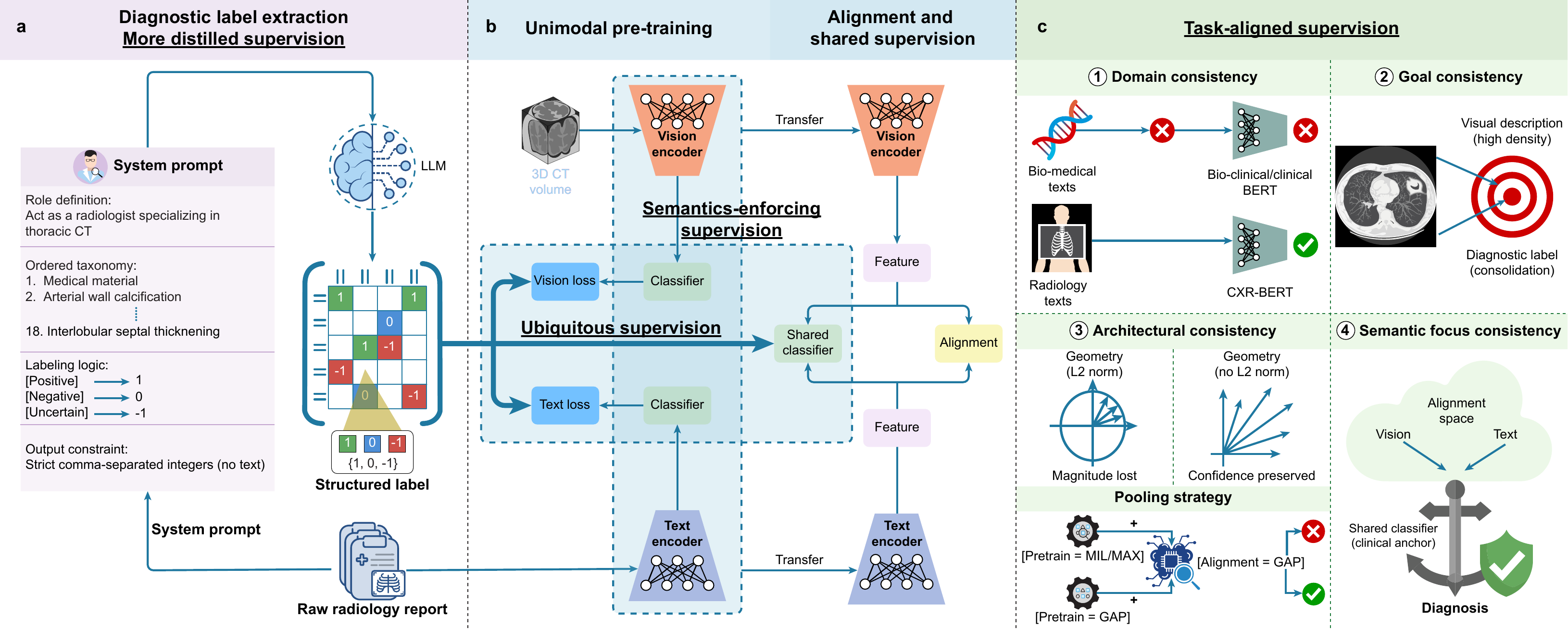}
    \caption{\textbf{MUST framework overview.} The framework is summarized by four empirically tested design principles. \textbf{a, More distilled supervision:} An LLM distills noisy radiology reports into structured diagnostic labels (present, absent, uncertain), creating a scalable source of supervision. \textbf{b, Ubiquitous \& semantics-enforcing supervision:} Supervision is applied throughout the pipeline: to the vision encoder, the text encoder, and the shared alignment space. Vision and text encoders are first pre-trained independently using the structured labels and then transferred to the alignment stage. \textbf{c, Task-aligning supervision:} The training pipeline is matched to downstream diagnosis through (1) domain consistency via a radiology-specific text encoder, (2) goal consistency by prioritizing diagnostic labels, (3) architectural consistency by aligning pooling strategies and removing $L_2$ normalization, and (4) semantic-focus consistency by anchoring the alignment space with a shared classifier.}
\label{fig:method_overview}
\end{figure}

GreenRFM implements the MUST framework as a two-stage image--report pre-training pipeline (Fig.~\ref{fig:method_overview}; Methods). This first results section tests the four principles in order: more distilled supervision by label quality, expert audit, uncertainty-aware training, and synthetic noise stress tests; ubiquitous and semantics-enforcing supervision by component-wise ablations and sequential-versus-joint optimization; and task-aligning supervision by domain-, supervision-, and architecture-alignment ablations.

\begin{figure}[!t]
    \centering
    \includegraphics[width=\textwidth]{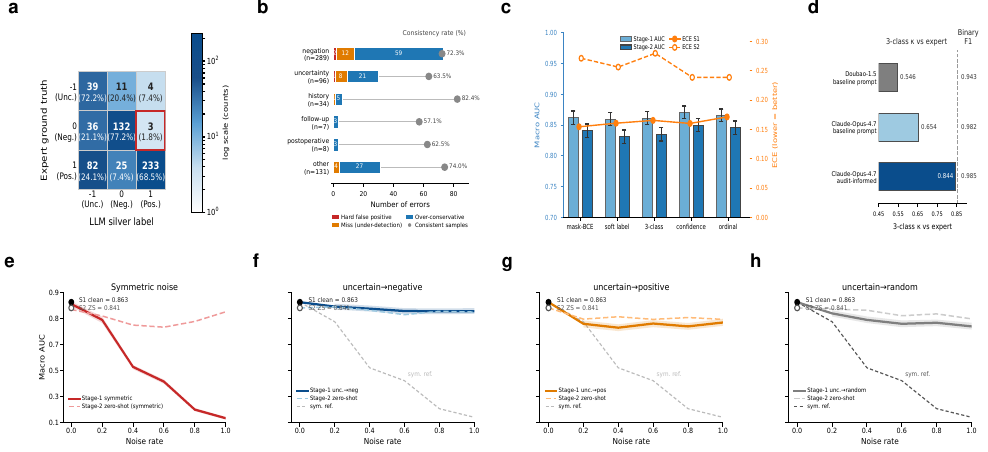}
    \caption{\textbf{Auditing and absorbing LLM silver-label noise.} Panel \textbf{a} shows the board-certified expert re-read confusion matrix ($n\!=\!565$), panel \textbf{b} stratifies residual errors by sentence semantic context, panel \textbf{c} compares uncertainty-aware training strategies on Merlin, panel \textbf{d} summarizes model- and prompt-level noise reduction, and panels \textbf{e--h} report Stage~1 and Stage~2 AUC under symmetric label flips and uncertainty-to-negative, uncertainty-to-positive, and uncertainty-to-random corruption. Together, these analyses show that the supervision pipeline is auditable and empirically robust under the tested uncertainty and noise settings.}
\label{fig:audit}
\end{figure}

\subsubsection*{LLM-distilled diagnostic labels provide effective supervision}
Principle~M predicts that LLM-distilled label supervision remains effective when residual errors are sufficiently small and predominantly conservative. We therefore distilled the reports into structured diagnostic labels and evaluated both label quality and downstream utility (Methods). The exact diagnostic label-extraction prompts are provided in Supplementary Appendix~\ref{supp:diagnostic_label_prompts}.

Against the official Merlin labels, agreement was high once uncertain cases were excluded ($\kappa=0.94$, F1 $0.96$), whereas the remaining confusion was concentrated between uncertain and definitive states rather than between definitive positive and negative labels (Fig.~\ref{fig:S1}; Supplementary Table~\ref{tab:S_label_eval}).

To test whether the silver labels retained enough diagnostic signal for training, we compared a vision encoder trained on LLM-derived labels with the same encoder trained on curated labels. The silver-label model reached AUC $87.41\%$ on the Merlin test set versus $85.78\%$ for curated-label training (Supplementary Table~\ref{tab:S_label_eval}). Under class imbalance, the decision boundary is shaped primarily by the minority positive class~\cite{japkowicz2002class,buda2018systematic}, so the dominant noise---confusion between negative and uncertain states---appears to have limited impact on learned representations in this setting.

\subsubsection*{Expert audit indicates that residual silver-label errors are predominantly conservative}

To identify sentence contexts most prone to LLM labelling errors, we built a linguistic risk map by classifying all $115{,}076$ disease--sentence pairs from $25{,}494$ Merlin reports into six contextual risk classes ($\approx\!26\%$ designated high-risk; Methods). A radiologist then re-read $n\!=\!565$ stratified samples drawn from this map (Fig.~\ref{fig:audit}a, Supplementary Tables~\ref{tab:audit_3x3}), yielding high binary agreement ($\kappa\!=\!0.848$, sensitivity $0.903$ [$0.866$--$0.937$], specificity $0.978$ [$0.949$--$1.000$], F1 $0.943$ [$0.921$--$0.963$]) and a hard false-positive rate of $3/565\!=\!0.53\%$. Stratifying errors by sentence semantic context showed that residual errors are systematic and predominantly conservative (Fig.~\ref{fig:audit}b). \emph{Negation} sentences---the context most often suspected of generating false positives---produced $2/289$ hard false positives ($0.7\%$), arguing against the worst-case scenario of widespread negation mishandling. \emph{Uncertainty-hedge} sentences had low three-class agreement ($63.5\%$, $96$ samples), but their failures were almost exclusively conservative: the LLM labelled $21/35$ errors as uncertain ($-1$) when the expert judged them positive or negative, with only one hard false positive. \emph{History} sentences had the highest agreement ($82.4\%$). Across all context types, $82$ expert-positive samples were conservatively labelled $-1$ (e.g., when the report writes ``renal cyst'', the radiologist scored the associated hypodense lesion as a positive finding but the LLM abstained under the implicit description). Masked-BCE training masks these cases rather than back-propagating them as noise.

\subsubsection*{Residual noise can be reduced and is selectively tolerated}
All main GreenRFM results use the original Doubao-extracted labels; the following analyses test how much residual label noise can be reduced or absorbed.

\paragraph{Model- and prompt-level reduction} Residual label error is not an unavoidable ceiling and can be reduced in two steps (Fig.~\ref{fig:audit}d, Supplementary Table~\ref{tab:prompt_tuning}). First, replacing Doubao with Claude Opus~4.7 under the same baseline prompt raised three-class $\kappa$ from $0.546$ to $0.654$ and binary F1 from $0.943$ to $0.982$. Second, applying six linguistic rules derived from Doubao's audit failure patterns to the Claude Opus~4.7 prompt further raised $\kappa$ to $0.844$ and binary F1 to $0.985$, reducing most of the remaining gap to curated labels in this audit sample.

\paragraph{Robustness to uncertainty-treatment strategy} GreenRFM's reported results use masked-BCE throughout. To test whether this conservative choice is adequate, we compared five uncertainty-treatment strategies on Merlin (Fig.~\ref{fig:audit}c; Supplementary Table~\ref{tab:uncertainty_strategy_ci}). All five produced Stage~1 Macro AUC values in the range $0.860$--$0.871$ and Stage~2 values in $0.831$--$0.850$, showing that training is relatively insensitive to the exact uncertainty treatment in this experiment. Across discrimination (AUC), calibration (ECE), and high-sensitivity operating point (Spec@Sens$_{0.95}$), confidence-weighted BCE showed consistent marginal advantages over masked-BCE (AUC $+0.009$, ECE $-0.03$, Spec@Sens$_{0.95}$ $+0.014$), providing a promising path for future pipeline optimization.

\paragraph{Selective robustness under synthetic stress} A synthetic noise stress test separated clinically realistic uncertainty corruption from arbitrary label inversion (Fig.~\ref{fig:audit}e-h). At Stage~1, even reassigning all uncertain labels to negative, positive, or random targets still yielded Macro AUCs of $0.828$, $0.785$, and $0.770$, respectively, relative to the clean baseline of $0.856$. By contrast, symmetric random flips caused a sharp performance drop to $0.529$ at $40\%$ noise and $0.135$ at $100\%$ noise. After alignment, uncertainty-type corruption kept zero-shot AUC in the $0.797$--$0.845$ range (Supplementary Table~\ref{tab:noise_robust_stage2}), indicating that GreenRFM is more tolerant of the structured uncertainty-like noise observed in reports than of arbitrary relabelling, consistent with prior evidence that deep neural networks can remain robust to substantial label noise when the corrupted labels still preserve task-relevant signal~\cite{rolnick2017deep}. Together, the agreement, audit, and stress-test analyses indicate that the distilled labels retain enough diagnostic signal for the explicit supervision terms tested below to be informative in this setting.

\subsubsection*{Ubiquitous and semantics-enforcing supervision improves representation quality and alignment dynamics}

\begin{figure}[!t]
    \centering
  \includegraphics[width=\textwidth]{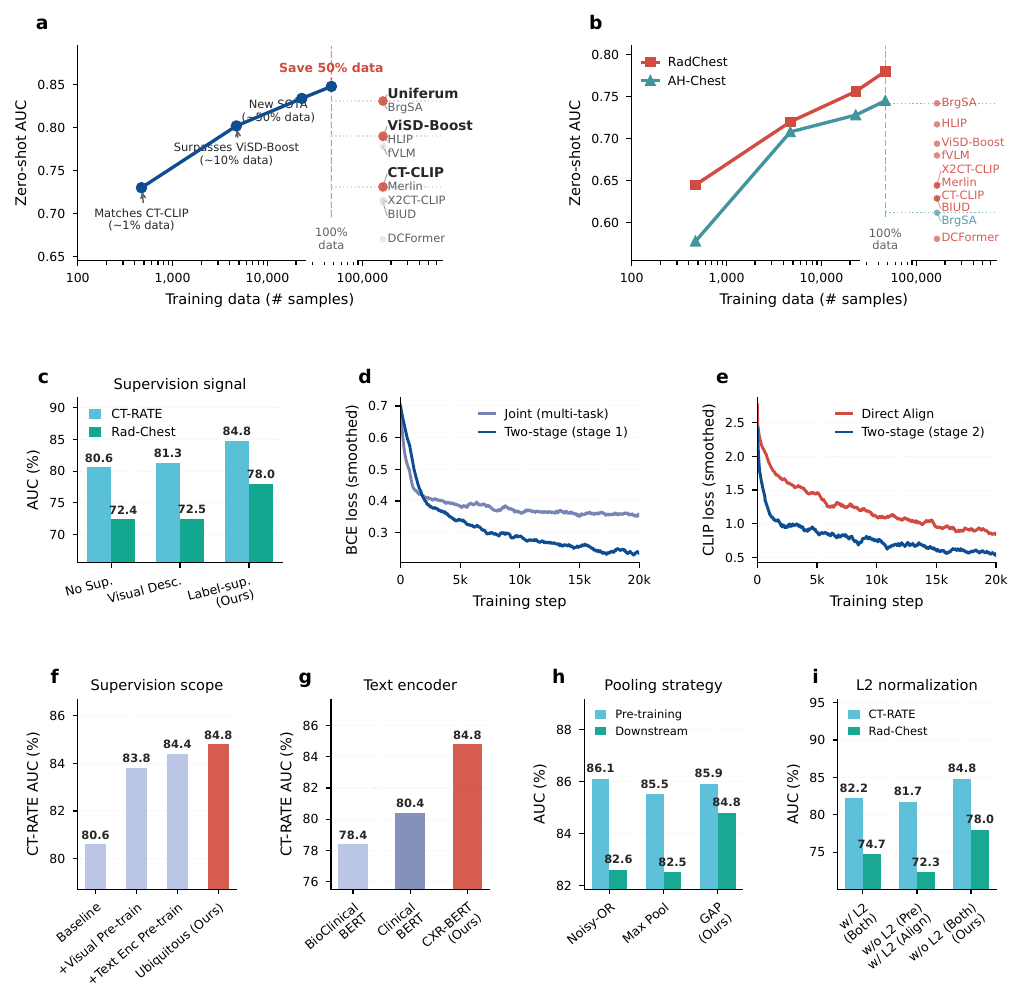}
    \caption{\textbf{Ablation evidence for MUST: data scaling, supervision signal, and task-aligned design.} \textbf{a} \,|\, Data scaling efficiency. GreenRFM exhibits a steeper scaling curve than prior methods, reaching prior benchmark performance with substantially less training data. \textbf{b} \,|\, Generalization scaling law. Zero-shot performance on RAD-ChestCT and AH-Chest improves consistently with training-data scale. \textbf{c} \,|\, Impact of supervision signal. Diagnostic label supervision outperforms visual-description supervision and no supervision on both CT-RATE and RAD-ChestCT. \textbf{d} \,|\, Two-stage versus joint multi-task representation learning. The Stage~1 vision encoder reaches lower BCE loss when trained independently than under joint optimization. \textbf{e} \,|\, Impact of two-stage training on alignment. The two-stage model converges faster and reaches lower alignment loss than direct alignment. \textbf{f} \,|\, Impact of supervision scope. Explicit supervision of the image encoder, text encoder, and aligned space yields the highest zero-shot performance among the tested variants. \textbf{g} \,|\, Impact of text encoder. Radiology-specific CXR-BERT outperforms broader biomedical encoders. \textbf{h} \,|\, Impact of pooling strategy. Global average pooling (GAP) performs best among the tested pooling strategies for downstream diagnosis. \textbf{i} \,|\, Impact of $L_2$ normalization. Removing $L_2$ normalization retains feature geometry and improves performance in this setting.}
  \label{fig:ablation_study}
\end{figure}

Ubiquitous and semantics-enforcing supervision predict that adding direct supervision to more components should yield positive marginal gains and that establishing unimodal diagnostic structure before alignment should yield more favorable optimization dynamics than earlier-alignment training schedules. We therefore compared alignment-only training, component-wise supervision variants, and a joint multi-task baseline using matched data and architecture (Fig.~\ref{fig:ablation_study}; Methods).

Relative to an alignment-only CLIP-style baseline (AUC $80.6\%$), adding direct supervision to the image encoder increased CT-RATE AUC by $+3.2$ points, adding text-encoder supervision added $+0.6$, and preserving supervision in the aligned space added $+0.4$, yielding a final AUC of $84.8\%$ (Fig.~\ref{fig:ablation_study}f). All three marginal gains were positive, with the vision-encoder term the largest.

Relative to the tested joint multi-task baseline, building unimodal diagnostic structure before alignment showed more favorable optimization dynamics. The Stage~1 vision model reached lower classification loss than the joint baseline (Fig.~\ref{fig:ablation_study}d), and the subsequent alignment stage converged faster and to a lower validation alignment loss (Fig.~\ref{fig:ablation_study}e), consistent with the practical prediction of Principle~S that alignment works best after unimodal diagnostic structure has already been established.

\subsubsection*{Task-aligned design improves zero-shot diagnostic utility}
Principle~T predicts that reducing train--inference mismatch in text domain, supervision target, pooling operator, and feature geometry should improve zero-shot diagnosis. We tested these mismatches one at a time in Fig.~\ref{fig:ablation_study}.

\paragraph{Radiology-specific text encoders improve alignment}
Consistent with Principle~T, the radiology-specific text encoder CXR-BERT outperformed broader biomedical encoders, reaching AUC $84.8\%$ versus $78.4\%$ for BioClinicalBERT and $80.4\%$ for ClinicalBERT (Fig.~\ref{fig:ablation_study}g).

\paragraph{Diagnostic targets outperform visual descriptions}
Using diagnostic labels as the pre-training target also outperformed visual-description supervision, yielding CT-RATE AUC $84.8\%$ versus $81.3\%$ (Fig.~\ref{fig:ablation_study}c). The exact prompt used to extract visual-description supervision is provided in Supplementary Appendix~\ref{supp:visual_description_prompt}.

\paragraph{Matched feature geometry improves diagnostic utility}
Removing $L_2$ normalization increased Stage~1 supervised AUC from $83.9\%$ to $85.9\%$; maintaining this removal during the alignment stage likewise increased Stage~2 zero-shot AUC from $81.7\%$ to $84.8\%$ (Fig.~\ref{fig:ablation_study}i). Using GAP throughout pre-training and evaluation yielded AUC $84.8\%$ versus $82.6\%$ for MIL-based pooling (Fig.~\ref{fig:ablation_study}h).

\paragraph{Maintaining semantic focus in the aligned space yields additional gains}
Adding the shared classifier during alignment increased CT-RATE zero-shot AUC from $84.4\%$ to $84.8\%$ (Fig.~\ref{fig:ablation_study}f), completing the matched design predicted by Principle~T.

\subsection*{GreenRFM generalizes efficiently across benchmarks}

Having established that principled supervision is reliable and improves representation learning, we next evaluated whether these gains translate into model-level performance, transfer, and representation quality across datasets, modalities, and evaluation settings.

\subsubsection*{GreenRFM improves zero-shot diagnosis with high data efficiency}

We validated GreenRFM on three large-scale public benchmarks: CT-RATE (internal), RAD-ChestCT (external), and Merlin (internal). Table~\ref{tab:comparison} summarizes split-matched literature comparisons on the official benchmark splits and label spaces, together with bootstrap uncertainty for GreenRFM predictions. Within this comparison, GreenRFM reached the highest central values across the listed metrics, including against methods that rely on larger backbones, computationally intensive pre-training, or complex auxiliary tasks.

\begin{table}[htbp]
    \centering
    \caption{\textbf{Zero-shot abnormality diagnosis performance comparison.} Comparison across internal (CT-RATE, Merlin) and external (RAD-ChestCT) validation benchmarks. \textbf{Bold} numbers denote the best central value within each column. Prior-method values are quoted from the original publications under the official benchmark splits and label spaces; they are therefore reported as split-matched literature comparisons rather than unified re-implementations. For our methods, we report the mean $\pm$ 95\% confidence interval estimated via bootstrap resampling of the model's predictions on the held-out test set.}
    \label{tab:comparison}
    \small
    \resizebox{\textwidth}{!}{
    \setlength{\tabcolsep}{4pt}
    \begin{tabular}{@{}l cccc cccc c@{}}
    \toprule
    Method
      & \multicolumn{4}{c}{CT-RATE}
      & \multicolumn{4}{c}{RAD-ChestCT}
      & \multicolumn{1}{c}{Merlin} \\
    \cmidrule(lr){2-5} \cmidrule(lr){6-9} \cmidrule(lr){10-10}
      & AUC(\%) & ACC(\%) & F1(\%) & Prec.(\%)
      & AUC(\%) & ACC(\%) & F1(\%) & Prec.(\%)
      & F1(\%) \\
    \midrule
    OpenCLIP \cite{ilharco2021openclip} & - & - & - & - & - & - & - & - & 27.6 $\pm$ 1.4 \\
    BioMedCLIP \cite{zhang2025multimodal} & - & - & - & - & - & - & - & - & 28.5 $\pm$ 1.1 \\
    DCFormer \cite{ates2025dcformer} & 67.0 & 62.4 & 66.8 & 28.6 & 58.1 & 55.8 & 60.8 & 29.9 & - \\
    X2CT-CLIP \cite{you2025x2ct} & 71.6 & - & - & - & 64.5 & - & - & - & - \\
    CT-CLIP \cite{hamamci2024ctrate}   & 73.1 & 66.8 & 70.7 & 32.3 & 62.9 & 59.5 & 64.2 & 33.6 & - \\
    VoCo \cite{wu2024large} & 73.7 & - & - & - & - & - & - & - & - \\
    BIUD \cite{cao2024bootstrapping}      & 71.3 & 68.1 & 71.6 & 33.8 & 62.9 & 60.6 & 65.2 & 33.7 & - \\
    Merlin \cite{blankemeier2024merlin}    & 72.8 & 67.2 & 70.9 & 33.7 & 64.4 & 61.9 & 66.3 & 34.8 & 74.1 $\pm$ 1.4 \\
    fVLM \cite{shui2025largescale}       & 77.8 & 71.8 & 75.1 & 37.9 & 68.0 & 64.7 & 68.8 & 37.4 & - \\
    HLIP \cite{zhao2025towards} & 78.7 & 72.4 & 75.5 & 38.4 & 71.7 & 67.7 & 71.4 & 39.8 & - \\
    ViSD-Boost \cite{cao2025boosting} & 79.0 & 73.1 & 75.9 & 38.7 & 69.4 & 65.2 & 69.3 & 34.2 & - \\
    BrgSA \cite{lai2025bridged}      & 82.9 & 77.0 & 79.3 & 43.2 & 74.2 & 68.6 & 72.0 & 42.2 & - \\
    Uniferum \cite{lee2025unified} & 83.1 & - & - & - & - & - & - & - & - \\

    \textbf{GreenRFM-L (ours)}
               & 83.2
               & 77.1
               & 79.4
               & 43.2
               & 76.5
               & 71.0
               & 74.1
               & 44.6
               & 82.1 \\
    \rowcolor{gray10}
    \textbf{GreenRFM (ours)}
               & \textbf{84.8 $\pm$ 0.8}
               & \textbf{77.7 $\pm$ 0.9}
               & \textbf{80.0 $\pm$ 0.8}
               & \textbf{44.4 $\pm$ 1.6}
               & \textbf{78.0 $\pm$ 0.9}
               & \textbf{73.0 $\pm$ 0.9}
               & \textbf{75.7 $\pm$ 0.8}
               & \textbf{46.1 $\pm$ 1.3}
               & \textbf{84.3 $\pm$ 1.4} \\
    \bottomrule
    \end{tabular}
    }
\end{table}

On CT-RATE, GreenRFM reached AUC $84.8\%$, exceeding the two closest competitors Uniferum ($83.1\%$) and BrgSA ($82.9\%$), as well as all other prior methods. The lightweight GreenRFM-L ($83.2\%$) likewise outperformed all published baselines. Both variants use a $33$M-parameter 3D ResNet-18 (Methods).

On the external RAD-ChestCT dataset, GreenRFM reached an AUC of 78.0\%, supporting generalization to unseen data distributions. On Merlin, GreenRFM reached an F1 score of 84.3\%, compared with the previous best reported F1 of 74.1\% (Table~\ref{tab:comparison}).

\paragraph{Comparison scope and statistical uncertainty} The baseline rows in Table~\ref{tab:comparison} quote the values reported in the original publications on the official CT-RATE/Merlin benchmark splits and label spaces. Because per-sample predictions and replicate-level outputs are unavailable for these literature rows, paired significance tests against those baselines cannot be computed. For GreenRFM, we report the mean and 95\% confidence interval (CI) estimated by bootstrap resampling of held-out predictions ($n{=}1000$ resamples). For the AUC columns in Table~\ref{tab:comparison}, the closest competing central estimates fall below the lower bound of GreenRFM's 95\% CI, which contextualizes the observed margins without treating the literature rows as paired statistical tests.

\subsubsection*{Performance scales efficiently with data}
As shown in Figure~\ref{fig:ablation_study}a, GreenRFM exhibits high data efficiency compared to previous works~\cite{shui2025largescale,blankemeier2024merlin}. It matches the previous best reported performance using less than 50\% of the training data, and the resulting scaling curve remains approximately log-linear, indicating that performance has not yet saturated. This trend extends to external validation: zero-shot performance on RAD-ChestCT and AH-Chest improves consistently with training-data scale (Fig.~\ref{fig:ablation_study}b), suggesting that the same data-efficiency pattern translates to out-of-distribution generalization.

\subsubsection*{Performance is robust to prompt variation}
To test whether the performance gain depends on prompt wording, we evaluated five positive/negative prompt templates:
\begin{itemize}
    \item \textbf{Prompt 1:} ``\textit{\{pathology\}.}'' vs. ``\textit{not \{pathology\}.}''
    \item \textbf{Prompt 2:} ``\textit{\{pathology\}.}'' vs. ``'' (empty string).
    \item \textbf{Prompt 3:} ``\textit{There is \{pathology\}.}'' vs. ``\textit{There is no \{pathology\}.}''
    \item \textbf{Prompt 4:} ``\textit{\{pathology\} is present.}'' vs. ``\textit{\{pathology\} is not present.}''
    \item \textbf{Prompt 5:} ``\textit{Findings are compatible with \{pathology\}.}'' vs. ``\textit{Findings are not compatible with \{pathology\}.}''
\end{itemize}

Across these prompt templates, the lowest CT-RATE AUC remained $82.5\%$, indicating limited prompt sensitivity.

\subsection*{GreenRFM transfers across institutions, modalities, and retrieval settings}

\subsubsection*{Transfer under covariate shift on private clinical benchmarks}
We next evaluated zero-shot robustness under covariate shift on the AH-Chest and AH-Abd cohorts, isolating cross-site generalization before any site-specific retraining.

GreenRFM achieved zero-shot AUC $74.5$ on AH-Chest and $73.2$ on AH-Abd, exceeding the modality-matched cross-site baselines BrgSA ($61.2$) and Merlin ($60.3$), respectively (Table~\ref{tab:private_benchmarks}).
Per-category diagnostic results for AH-Chest and AH-Abd are provided in Supplementary Tables~\ref{tab:S5} and~\ref{tab:S6}.

\begin{table}[htbp]
    \centering
    \begin{threeparttable}
    \caption{\textbf{Generalization beyond the source CT diagnosis benchmarks.} Unified summary of cross-site CT transfer, report-to-volume retrieval, and cross-modality MRI transfer.}
    \label{tab:transfer_retrieval_summary}
    \label{tab:private_benchmarks}
    \label{tab:report_to_volume_retrieval}
    \label{tab:mri_results}
    \footnotesize
    \setlength{\tabcolsep}{2.8pt}
    \renewcommand{\arraystretch}{1.12}
    \begin{tabular}{@{}>{\raggedright\arraybackslash}p{1.55cm}>{\raggedright\arraybackslash}p{2.95cm}*{8}{>{\centering\arraybackslash}p{0.88cm}}@{}}
        \toprule
        Dataset & Method & \multicolumn{4}{c}{Zero-shot diagnosis} & \multicolumn{4}{c}{Report-to-volume retrieval} \\
        \cmidrule(lr){3-6} \cmidrule(lr){7-10}
        & & AUC(\%) & ACC(\%) & F1(\%) & Prec.(\%) & R@5 & R@10 & R@50 & R@100 \\
        \midrule
        \multicolumn{10}{@{}l}{\textbf{Cross-site CT transfer before local retraining}} \\
        \addlinespace[1pt]
        AH-Chest & BrgSA \cite{lai2025bridged} & 61.2 & 74.3 & 74.9 & 16.2 & 0.02 & 0.03 & 0.12 & 0.23 \\
        \rowcolor{gray10}
         & \textbf{GreenRFM (ours)} & \textbf{74.5} & 71.3 & 74.7 & \textbf{40.2} & \textbf{0.11} & \textbf{0.21} & \textbf{0.80} & \textbf{1.30} \\
        AH-Abd & Merlin \cite{blankemeier2024merlin} & 60.3 & 59.9 & 67.2 & 23.4 & 0.22 & 0.38 & 1.6 & 2.9 \\
        \rowcolor{gray10}
         & \textbf{GreenRFM (ours)} & \textbf{73.2} & \textbf{87.4} & \textbf{85.4} & \textbf{40.3} & \textbf{0.70} & \textbf{1.30} & \textbf{4.0} & \textbf{6.2} \\
        \midrule
        \multicolumn{10}{@{}l}{\textbf{CT report-to-volume retrieval on source and external cohorts}} \\
        \addlinespace[1pt]
        CT-RATE & Random & \multicolumn{4}{c}{---} & 0.1 & 0.4 & 2.0 & 3.4 \\
         & CT-CLIP \cite{hamamci2024ctrate} & \multicolumn{4}{c}{---} & 2.9 & 5.0 & 18.0 & 28.7 \\
         & Merlin \cite{blankemeier2024merlin} & \multicolumn{4}{c}{---} & 1.5 & 2.7 & 7.7 & 12.7 \\
         & BrgSA \cite{lai2025bridged} & \multicolumn{4}{c}{---} & 5.8 & 10.1 & 28.6 & 42.0 \\
        \rowcolor{gray10}
         & \textbf{GreenRFM (ours)} & \multicolumn{4}{c}{---} & \textbf{9.5} & \textbf{16.0} & \textbf{39.9} & \textbf{54.3} \\
        Merlin & Random & \multicolumn{4}{c}{---} & 0.1 & 0.2 & 1.0 & 1.8 \\
         & Merlin \cite{blankemeier2024merlin} & \multicolumn{4}{c}{---} & 2.9 & 4.8 & 14.6 & 23.0 \\
        \rowcolor{gray10}
         & \textbf{GreenRFM (ours)} & \multicolumn{4}{c}{---} & \textbf{22.5} & \textbf{32.9} & \textbf{62.2} & \textbf{74.8} \\
        \midrule
        \multicolumn{10}{@{}l}{\textbf{Cross-modality MRI transfer}} \\
        \addlinespace[1pt]
        AH-Knee & MRCLIP & 72.9 & 85.5 & 83.7 & 41.7 & 17.2 & 28.7 & 78.2 & 95.1 \\
        \rowcolor{gray10}
         & \textbf{GreenRFM (ours)} & \textbf{79.0} & \textbf{87.3} & \textbf{86.1} & \textbf{46.3} & \textbf{26.1} & \textbf{41.2} & \textbf{84.7} & \textbf{97.6} \\
        AH-Spine & MRCLIP & 70.5 & 87.8 & 85.7 & 35.0 & 7.2 & 13.4 & 42.2 & 65.1 \\
        \rowcolor{gray10}
         & \textbf{GreenRFM (ours)} & \textbf{80.8} & \textbf{89.4} & \textbf{88.5} & \textbf{44.5} & \textbf{10.3} & \textbf{18.3} & \textbf{53.0} & \textbf{72.9} \\
        \bottomrule
    \end{tabular}
    \begin{tablenotes}[flushleft]
        \footnotesize
        \item The left block reports zero-shot diagnosis and the right block report-to-volume retrieval. CT-RATE and Merlin contribute retrieval-only rows here because their public zero-shot diagnosis comparison is summarized in Table~\ref{tab:comparison}.
    \end{tablenotes}
    \end{threeparttable}
\end{table}

\subsubsection*{Cross-modal retrieval supports semantic transfer}
Beyond diagnostic classification, we evaluated semantic alignment through report-to-volume retrieval on CT-RATE, Merlin, AH-Chest, and AH-Abd (Table~\ref{tab:report_to_volume_retrieval}). This task requires identifying the correct CT volume from a large candidate pool based solely on a textual description. GreenRFM achieved the highest Recall@$K$ values in this comparison on every dataset: on CT-RATE, R@50/R@100 reached $39.9/54.3$ versus $28.6/42.0$ for BrgSA and $18.0/28.7$ for CT-CLIP; on Merlin, R@5 reached $22.5$ versus $2.9$ for the Merlin baseline; and on AH-Abd, R@100 reached $6.2$ versus $2.9$ for Merlin.

\subsubsection*{Extension to MRI supports modality-level generality}

To test whether MUST supervision principles remain effective when applied to a different imaging modality, we applied the GreenRFM training pipeline directly to two musculoskeletal MRI cohorts (AH-Knee and AH-Spine), training from scratch on MRI image--report pairs and comparing against a matched CLIP-style baseline (MRCLIP) trained on the same splits.
GreenRFM outperformed MRCLIP on both cohorts across both evaluation tasks (Table~\ref{tab:mri_results}). For zero-shot diagnosis, AUC improved from $72.9$ to $79.0$ on AH-Knee ($+6.1$) and from $70.5$ to $80.8$ on AH-Spine ($+10.3$). For report-to-volume retrieval, R@10 improved from $28.7$ to $41.2$ on AH-Knee and from $13.4$ to $18.3$ on AH-Spine. The consistent gains across both cohorts and both task types indicate that the supervision design is applicable to other 3D imaging modalities.
Per-category MRI diagnostic results are reported in Supplementary Tables~\ref{tab:S7} and~\ref{tab:S8}.

\subsection*{GreenRFM preserves semantic structure beyond zero-shot diagnosis}

\begin{figure}[!t]
    \centering
    \includegraphics[width=\textwidth]{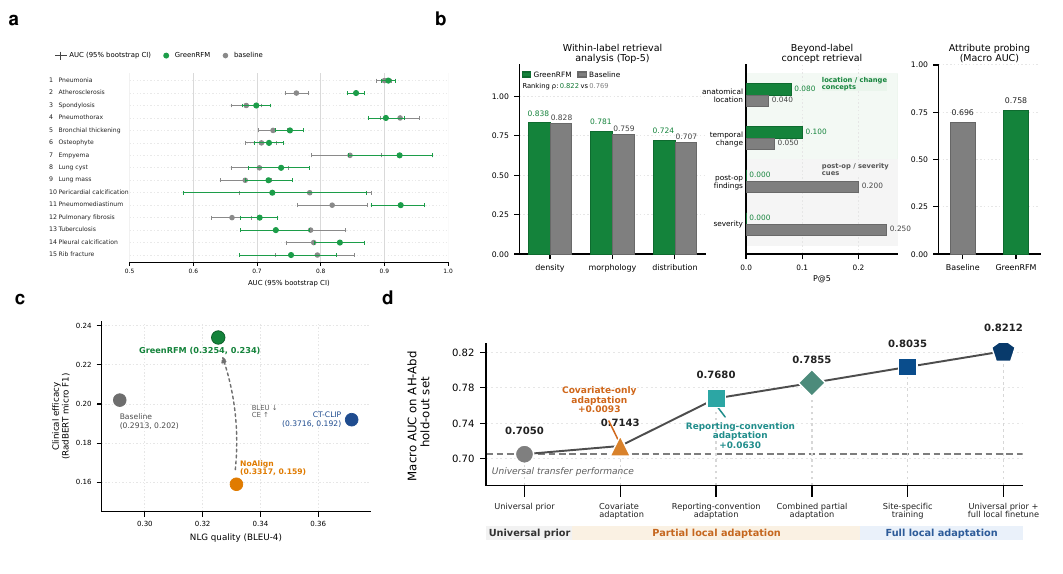}
    \caption{\textbf{Semantic preservation and deployment adaptation of GreenRFM.}
    \textbf{a} \,|\, Long-tail findings recognition on the 15-pathology CT-RATE benchmark of Lai \emph{et al.}~\cite{lai2025bridged}. Per-class zero-shot AUCs are shown with 95\% bootstrap confidence intervals; GreenRFM improves macro AUC to $\approx\!0.90$ versus $0.77$ for the matched baseline and achieves higher AUC on $11/15$ out-of-distribution findings.
    \textbf{b} \,|\, Semantic probing within and beyond the 18-label training space on CT-RATE. Left: within-label image-to-image retrieval at Top-5, scored against LLM-derived density, morphology, and distribution attributes; GreenRFM improves attribute matching and yields stronger agreement between attribute-correlated dimensions and full-embedding neighbourhood structure ($\rho=0.822$ vs.\ $0.769$). Middle: beyond-label concept retrieval (P@5) for clinical concept families absent from the training labels, including anatomical location, temporal change, post-operative findings, and severity. Right: linear attribute probes on $n\!=\!47{,}149$ CT-RATE embeddings show higher macro AUC for GreenRFM than for the baseline ($0.758$ vs.\ $0.696$).
    \textbf{c} \,|\, Encoder-swap report-generation probe. Each point represents one frozen visual encoder inserted into a Reg2RG-style LLaMA-2-7B decoder and evaluated in BLEU-4 versus RadBERT micro-F1 space. GreenRFM ($0.3254$, $0.234$) preserves stronger clinical content than CT-CLIP ($0.3716$, $0.192$), the weakly aligned baseline ($0.2913$, $0.202$), and NoAlign ($0.3317$, $0.159$), indicating that linguistic fluency and clinical efficacy are not identically ranked across encoders.
    \textbf{d} \,|\, Deployment intervention ladder on AH-Abd, ordered by increasing local information. Macro AUC rises from the universal zero-shot prior remeasured on the held-out 20\% local-adaptation split ($0.7050$, dashed line) to covariate-only batch-normalization affine adaptation ($0.7143$), frozen-backbone local linear probing ($0.7680$), combined partial adaptation ($0.7855$), site-specific two-stage training from scratch ($0.8035$), and full local fine-tuning from the cross-institutional checkpoint ($0.8212$).}

    \label{fig:robustness_probing}
\end{figure}

To test whether diagnostic supervision compressed the representation or narrowed GreenRFM to a fixed diagnostic-label space, we evaluated rare findings, within-label and beyond-label semantics, and report-generation encoder transfer (Fig.~\ref{fig:robustness_probing}).

We compared GreenRFM against a CLIP-style baseline under identical architectures and training data in a coordinated probing suite.

\subsubsection*{Long-tail finding recognition} On a 15-pathology CT-RATE long-tail benchmark~\cite{lai2025bridged}, GreenRFM improved zero-shot macro AUC to approximately $0.90$ versus $0.77$ for the baseline, with gains visible across 11/15 rare findings (Fig.~\ref{fig:robustness_probing}a). Difficult categories such as pleural calcification and rib fracture still favored GreenRFM ($0.85$ vs.\ $0.74$ and $0.83$ vs.\ $0.66$, respectively), suggesting that diagnostic supervision can extend to long-tail categories outside the pre-training label space rather than benefiting only in-distribution findings.

\subsubsection*{Within-label retrieval} Within each label's positive cases, image-to-image nearest-neighbour retrieval was scored against three LLM-derived attribute groups (density, morphology, distribution) independent of the diagnostic labels. GreenRFM's Top-5 attribute match rate exceeded the CLIP baseline at every $K$ (e.g., morphology: $0.781$ vs.\ $0.759$; distribution: $0.724$ vs.\ $0.707$ at $K\!=\!5$; Fig.~\ref{fig:robustness_probing}b and Supplementary Table~\ref{tab:within_label_retrieval}). To understand which part of the embedding drives within-label ordering, we partitioned all 768 embedding dimensions into those whose correlation was stronger with visual attributes than with diagnostic labels ($n\!=\!88$ for GreenRFM vs.\ $n\!=\!72$ for baseline). The Spearman rank correlation between the ordering produced by this attribute-correlated subset and the full-embedding ordering was $\rho\!=\!0.822$ for GreenRFM versus $\rho\!=\!0.769$ for the baseline, indicating that GreenRFM's within-label neighbourhoods are more aligned with visual attributes in this analysis.

\subsubsection*{Beyond-label concept retrieval} We encoded 24 clinical concept queries (across five categories: anatomical location, temporal change, post-operative/device findings, severity, texture/appearance) that do not appear in the 18-label training set and ranked them against CT images (Fig.~\ref{fig:robustness_probing}b and Supplementary Table~\ref{tab:beyond_label_retrieval}). Both methods retrieved above-chance information in every category. The results split by concept type: GreenRFM was superior on temporal change (P@10: $0.175$ vs.~$0.075$) and anatomical location (P@5: $0.080$ vs.~$0.040$), where identifying a finding requires representing disease state and spatial context. The CLIP baseline was superior on post-operative findings, severity, and texture/appearance (e.g., severity P@5: $0.250$ vs.~$0.000$).

\subsubsection*{Attribute probing} To test whether diagnostic supervision compresses non-diagnostic information, we linear probe the models on full CT-RATE training embeddings ($n\!=\!47{,}149$) and evaluated on the validation set (Fig.~\ref{fig:robustness_probing}b). GreenRFM's mean attribute AUC was $0.758$ versus $0.696$ for the CLIP baseline ($\Delta\!=\!+0.062$), indicating that GreenRFM's representations encode more fine-grained visual attribute information than the CLIP baseline. Per-attribute AUCs are reported in Supplementary Table~\ref{tab:attr_probe_perlabel}. Centered Kernel Alignment and mutual-information analyses were concordant: GreenRFM's diagnostic CKA increased by $+0.050$ relative to baseline, while its attribute CKA changed by only $-0.006$ ($0.162$ vs.\ $0.168$), and attribute mutual information was virtually identical ($0.079$ vs.\ $0.078$~nats; Supplementary Table~\ref{tab:attribute_probing}), suggesting that LLM-distilled diagnostic supervision strengthened the diagnostic representation without evidence of reduced attribute information.

\subsubsection*{Report-generation probe supports preserved report-level semantics}
To test whether the representations also transfer to generative tasks and encode clinically specific content beyond predefined categories, we replaced only the frozen visual encoder in a Reg2RG-style LLaMA-2-7B report generator~\cite{chen2025large}, holding all other components fixed, and evaluated generated reports on two complementary axes: linguistic fluency (BLEU-4) and clinical accuracy (CE: RadBERT micro F1 for 18 disease categories) (Fig.~\ref{fig:robustness_probing}c). GreenRFM achieved the highest CE micro F1 among the four encoders ($0.234$), outperforming CT-CLIP ($0.192$) and the matched CLIP-style baseline ($0.202$). CT-CLIP led in linguistic fluency (BLEU-4 $0.3716$) but ranked lower in clinical efficacy. The pre-alignment NoAlign Stage~1 encoder scored lowest (CE F1 $0.159$), consistent with alignment contributing to clinical content accuracy; the divergence between BLEU-4 and CE rankings across encoders further indicates that linguistic fluency and clinical content accuracy reflect distinct properties of the learned representation.

\subsection*{Deployment beyond cross-site transfer}

The preceding analyses established how far GreenRFM generalizes as a cross-site model across datasets, modalities, and evaluation settings. We next ask what happens at the deployment boundary, where cross-site pre-training may no longer be sufficient and a hospital may need to retrain the model locally under realistic compute constraints.

\subsubsection*{Local retraining decomposes the deployment gap across shift components}

AH-Abd represents a deployment site that differs from the Merlin source corpus along the shift axes introduced in Fig.~\ref{fig:teaser}b, so we evaluated a deployment intervention ladder that progressively injects local information (Fig.~\ref{fig:robustness_probing}d). This design is not intended to isolate single causal factors, but to provide an operational decomposition of which forms of local adaptation most directly reduce the deployment gap. The Merlin-trained checkpoint started at macro AUC $0.7050$ under local reporting conventions, which we treat as the zero-shot starting point.

Updating only batch-normalization affine parameters raised AUC to $0.7143$ ($+0.93$\,pt), indicating that acquisition-driven covariate adaptation helped, but only modestly. Freezing the backbone and training a local linear classifier raised AUC to $0.7680$ ($+5.37$\,pt over the BN rung), suggesting that local class priors, diagnostic thresholds, and label semantics explained a larger share of the deployment gap than scanner or protocol differences. Combining these two partial adaptations further improved AUC to $0.7855$, but still fell short of full local training and fine-tuning: site-specific training reached $0.8035$, whereas full fine-tuning initialized from the Merlin checkpoint reached $0.8212$.

The ordered gains therefore match the qualitative decomposition in Fig.~\ref{fig:teaser}b: covariate adaptation helped modestly, local prior and concept re-mapping produced the largest partial gain, and report-aware joint refinement closed much of the remaining gap. GreenRFM's low training cost makes these local regimes computationally feasible, converting compute efficiency from a training-side advantage into a practical deployment route.

\subsection*{GreenRFM representations transfer to downstream clinical tasks}

\begin{figure}[!t]
    \centering

    \includegraphics[width=\textwidth]{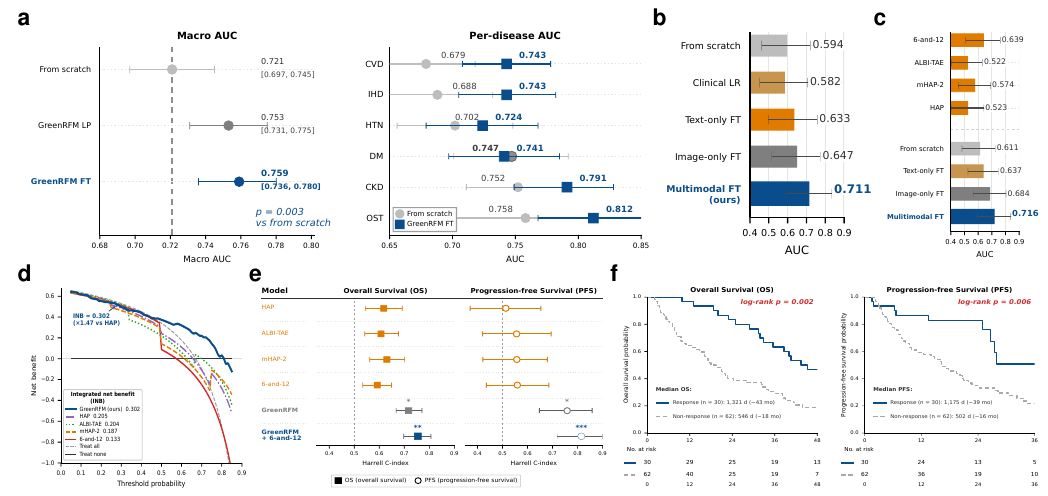}
    \caption{\textbf{Translational downstream evaluation.}
    \textbf{a} \,|\, Macro AUC for five-year incident-disease prediction across six categories; bars show $95\%$ bootstrap CIs. GreenRFM Stage-1 fine-tuning gives $+0.038$ over from-scratch ($p\!=\!0.003$), and a frozen Stage~2 linear probe reaches $0.753$ ($95\%$ CI $[0.731, 0.775]$).
    \textbf{b} \,|\, HCC microvascular-invasion prediction across image-only, text-only, and image $+$ text configurations; multimodal fusion through the aligned text encoder is the best-performing configuration in this cohort.
    \textbf{c} \,|\, TACE non-complete-response prediction AUC across eight model configurations on WAW-TACE (positive class: LR-TR equivocal/viable disease); upper group shows clinical scoring systems (6-and-12, ALBI-TAE, mHAP-2, HAP) and lower group shows learned models (from-scratch, text-only fine-tuning, image-only fine-tuning, and GreenRFM multimodal fine-tuning). GreenRFM multimodal fine-tuning achieves AUC 0.716 ($95\%$ CI $[0.593, 0.837]$), exceeding all clinical baselines and image-only models; error bars show 95\% bootstrap CIs.
    \textbf{d} \,|\, Decision-curve analysis on WAW-TACE: GreenRFM has higher net benefit than HAP, mHAP-2, ALBI-TAE, 6-and-12, and treat-all/treat-none across the evaluated threshold range.
    \textbf{e} \,|\, Cox C-indices for overall survival (OS) and progression-free survival (PFS) on WAW-TACE; error bars show $1{,}000$-replicate bootstrap $95\%$ CIs. The GreenRFM $\!+\!$ 6-and-12 fusion has non-overlapping CIs against all individual clinical scores for PFS; for OS, its CI is non-overlapping against HAP, ALBI-TAE, and 6-and-12, with only a slight overlap against mHAP-2(0.003).
    \textbf{f} \,|\, Kaplan--Meier curves stratified by true complete-response status on WAW-TACE confirm that the response endpoint is itself prognostic (OS cohort, $n\!=\!92$, log-rank $p\!=\!0.002$; PFS subset, $n\!=\!43$, log-rank $p\!=\!0.006$), supporting the clinical relevance of improving this endpoint while motivating prospective validation.}
    \label{fig:downstream}
\end{figure}

We evaluated whether GreenRFM's two downstream assets---discriminative image representations and the aligned text encoder---transfer to image-only long-horizon disease prediction and multimodal oncology endpoints analysis (Fig.~\ref{fig:downstream}). Unlike vision-centric pipelines that treat the text encoder as disposable scaffolding, GreenRFM retains both for downstream use; we test each regime below.

\subsubsection*{Long-horizon disease prediction (Merlin 5-year disease prediction subset, $n\!=\!2{,}488$)} GreenRFM attained five-year incident-disease macro AUC $0.759$ ($95\%$ CI $[0.736, 0.780]$), versus $0.721$ for from-scratch ($p\!=\!0.003$, $10{,}000$-permutation test). More importantly, with the GreenRFM encoder frozen, linear probing already reached macro AUC $0.753$ ($95\%$ CI $[0.731, 0.775]$). This is comparable to Merlin's own reported multi-disease 5-year prediction benchmark (AUROC $0.757$, $95\%$ CI $[0.743, 0.772]$)~\cite{blankemeier2024merlin}. By contrast, from-scratch linear probing fell to AUC $\approx\!0.59$ under the same protocol. The small gap between linear probing and full fine-tuning indicates that the prognostic signal is already organized in the learned representation rather than being created only during downstream adaptation.

\subsubsection*{Hepatocellular-carcinoma microvascular invasion prediction (HCC MVI, $n\!=\!190$, 7-sequence Gd-EOB-DTPA-enhanced MRI)} MVI prediction is a cross-modal-transfer challenge ($\text{CT}\!\to\!\text{MRI}$). Fine-tuning GreenRFM trained on Merlin abdominal CT reached macro AUC $0.711$ ($95\%$ CI $[0.591, 0.835]$), exceeding from-scratch ($0.594$), clinical logistic regression ($0.582$), text-only fine-tuning ($0.633$), and image-only fine-tuning ($0.647$)~(Fig.~\ref{fig:downstream}b). The text encoder received a structured natural-language string encoding age, sex, key laboratory values (albumin, bilirubin, ALT, AST, PT-INR) and hepatic function indices (ALBI grade, FIB-4)~(Supplementary Appendix~\ref{supp:downstream_detail}); its \texttt{[CLS]} embedding was projected and concatenated with the image feature before the classification head. The multimodal advantage over image-only fine-tuning ($+0.064$) indicates that the aligned text encoder can provide complementary information that is not captured by image features alone.

\subsubsection*{TACE response and survival (WAW-TACE, $n\!=\!229$, four-phase CT)} Fine-tuning GreenRFM trained on Merlin attained AUC $0.716$ ($95\%$ CI $[0.593, 0.837]$) for predicting the non-complete-response endpoint (label 1; LR-TR equivocal/viable disease), exceeding all four clinical scoring systems (6-and-12~\cite{wang2019development}: $0.639$; mHAP-2~\cite{park2016addition}: $0.574$; ALBI-TAE~\cite{lee2019new}: $0.522$; HAP~\cite{kadalayil2013simple}: $0.523$) and all three learned baselines including image-only fine-tuning ($0.684$), text-only fine-tuning ($0.637$), and from-scratch training ($0.611$) in this cohort (Fig.~\ref{fig:downstream}c). Decision-curve analysis showed that GreenRFM had higher net benefit than all comparators across the evaluated threshold range, yielding an integrated net benefit of $0.302$---$1.47\!\times\!$ the next-best HAP score ($0.205$) in this cohort (Fig.~\ref{fig:downstream}d). For survival association analysis on the held-out test set ($n\!=\!92$; PFS subset $n\!=\!43$), Cox C-indices for OS and PFS were $0.717$ and $0.758$ for GreenRFM features alone, rising to $0.754$ ($95\%$ CI $[0.698, 0.806]$) and $0.815$ ($95\%$ CI $[0.719, 0.899]$) when the numeric 6-and-12 score was added to the GreenRFM-feature Cox model. The PFS CI was non-overlapping against every individual clinical baseline, whereas the OS CI was non-overlapping against HAP, ALBI-TAE, and 6-and-12 and overlapped only slightly with mHAP-2 (Fig.~\ref{fig:downstream}e; Supplementary Appendix~\ref{supp:downstream_detail}). Kaplan--Meier analysis confirmed that complete-response status was itself prognostic: in the OS cohort ($n\!=\!92$), complete responders (label 0; $n\!=\!30$) achieved a median OS of $1{,}321$ days (${\approx}43$ months) versus $546$ days (${\approx}18$ months) for non-complete responders (label 1; $n\!=\!62$; log-rank $p\!=\!0.002$). In the PFS-annotated subset ($n\!=\!43$), complete responders ($n\!=\!28$) achieved a median PFS of $1{,}175$ days (${\approx}39$ months) versus $502$ days (${\approx}16$ months) for non-complete responders ($n\!=\!15$; log-rank $p\!=\!0.006$) (Fig.~\ref{fig:downstream}f), supporting the clinical relevance of improving prediction of this endpoint while requiring prospective validation before clinical use.

\section*{Discussion}

GreenRFM starts from a changed premise for radiology foundation models. Clinical archives remain abundant, but modern LLMs can now convert many routine reports into low-cost structured diagnostic labels, moving radiology toward a more label-rich setting for common findings. In this regime, the central question is no longer only how to extract representations from largely unlabeled image--report pairs. It is also how clinically grounded supervision should be obtained, injected into unimodal encoders, preserved during alignment, and adapted at deployment. The results support a supervision-centric technical route for diagnosis-centered RFMs: scale remains useful, but carefully designed supervision can make scale more data-efficient, computationally accessible, and locally deployable.

The methodological contribution is a two-stage training framework based on the MUST principles. We do not present MUST as necessary-and-sufficient theory; rather, it is an empirical supervision-design program that integrates choices prior efforts have touched but pursued in isolation~\cite{blankemeier2024merlin,lai2025bridged,shui2025largescale,silva2025reality}. GreenRFM distils report supervision, applies it to the image encoder, text encoder, and aligned space, strengthens each unimodal representation before alignment, and matches the training pipeline to the diagnostic task. Across RAD-ChestCT, AH-Chest, and AH-Abd, GreenRFM transfers under external scanner, protocol, and cohort differences, suggesting that domain-grounded supervision can improve robustness without relying on scale alone.

A natural concern is that supervised pre-training might compress radiology semantics into the predefined label set and overlook information such as lesion size, shape, or texture. The representation-probing suite argues against this concern in our setting. Attribute linear-probe AUC reaches $0.758$ for GreenRFM versus $0.696$ for the CLIP baseline, and this $+0.062$ gain exceeds the corresponding diagnostic gain ($+0.046$). CKA gives the same interpretation: supervision enriches diagnostic structure ($0.456$ vs.\ $0.406$) without displacing attribute structure ($0.162$ vs.\ $0.168$). These results support a different mechanism: diagnostic supervision anchors the representation around clinically meaningful states while preserving axes for finer-grained attributes. They also show alignment cannot compensate for weak unimodal encoders. In medical vision--language pre-training, the quality of unimodal representations bounds what cross-modal alignment can recover. This lens further exposes a practical pitfall: small train--inference inconsistencies, including mismatched feature normalization or pooling strategies, cost more than $2\%$ in downstream accuracy in our ablations, often exceeding the gains produced by substantially larger training subsets.

GreenRFM also changes the role of the text branch. Many RFMs treat the text encoder as an auxiliary training signal for the vision encoder and then discard or ignore it at inference. GreenRFM instead treats the aligned text encoder as a reusable clinical-information encoder within a diagnosis-centered scope. In the TACE survival task, fusing GreenRFM features with the 6-and-12 clinical score raises the PFS Cox C-index from $0.559$ for the score alone to $0.815$, with non-overlapping bootstrap confidence intervals in this cohort (Fig.~\ref{fig:downstream}).

Deployment exposes the boundary of universal pre-training, a known risk for clinical AI systems under dataset shift~\cite{finlayson2021clinician}. When institutional thresholds and reporting vocabulary diverge from the source corpus, broader pre-training alone may be insufficient. On AH-Abd, the universal model plateaus at macro AUC $0.7050$, whereas full local fine-tuning from the universal checkpoint reaches $0.8212$ (Fig.~\ref{fig:robustness_probing}). The intervention ladder clarifies the source of this gap: batch-normalization adaptation raises AUC only to $0.7143$, whereas adding local semantic supervision through a linear probe reaches $0.7680$, and full local retraining closes most of the remaining deficit. Thus, scanner and protocol covariates matter, but reporting-convention and concept shifts appear to dominate this deployment setting. Because GreenRFM derives supervision from reports and retains it during alignment, local archives can provide both the labels and the adaptation signal needed to re-tune the model within a local reporting ecosystem. Clinically connected hospital networks, such as Chinese medical consortia~\cite{nhc2020yiliantimanagement}, are therefore plausible settings for GreenRFM-style local adaptation, where reporting templates, referral pathways, and diagnostic conventions may be more closely aligned than across unrelated institutions; prospective multi-centre validation remains necessary before such network-level deployment.

If local adaptation is needed for deployment, its computational cost becomes part of the deployment problem. If adaptation requires specialized large-scale infrastructure, medical AI remains concentrated in the few institutions able to afford large training runs. This constraint is especially important for academic and clinical groups that must develop methods under limited data and hardware. GreenRFM is designed for this regime. The standard variant requires $24$ GPU-hours on a single $24$-GB GPU, versus tens of thousands of high-performance GPU-hours for VoCo~\cite{wu2024large} and Pillar-0~\cite{agrawal2025pillar}; the lightweight variant trains in $4$ hours on $6\,$GB VRAM. Substituting high-quality supervision for raw computational scale therefore supports a more sustainable and broadly accessible path for RFM development~\cite{paschali2025foundation}.

Three limitations qualify these conclusions. First, the study is retrospective. Second, LLM-extracted labels were not exhaustively verified by humans at full dataset scale. A related concern is that performance might reflect overfitting to a particular LLM's labelling biases rather than clinical understanding. Three checks argue against that interpretation. The LLM labels agree strongly with Merlin's human-verified findings, and a board-certified re-read of $565$ stratified samples yielded binary F1 $=0.943$ ($\kappa\!=\!0.848$, hard false-positive rate $0.53\%$; Fig.~\ref{fig:audit}); residual errors were mostly conservative, and an audit-informed prompt raised binary F1 to $0.985$. A dual-LLM consensus protocol further filters model-specific hallucinations from the private datasets. Finally, GreenRFM transfers to external benchmarks with independent, human-curated labels, including RAD-ChestCT, AH-Knee, and AH-Spine. Third, the present implementation is diagnosis-centered. It supports diagnosis, retrieval, multimodal fusion, generation-oriented transfer, and local adaptation, but dense prediction tasks such as segmentation would require spatial supervision, task-specific heads, and different alignment choices rather than direct reuse of the shared-classifier design. These tests strengthen retrospective evidence but do not replace prospective validation; multi-centre trials remain a necessary next step~\cite{stewart2024evidence}.

Taken together, the audit, uncertainty, ablation, probing, fusion, and local-training studies support a clear conclusion: in the LLM era, supervision is no longer merely a scarce annotation resource, but a central design object for radiology foundation models. GreenRFM organizes this route into a testable program for obtaining, injecting, preserving, and adapting clinical supervision. These results suggest that future RFMs can benefit not only from more data, but also from better supervision: supervision that is clinically grounded, auditable, retained across modalities, aligned with downstream tasks, and feasible to adapt within the clinical institutions that need it.

\section*{Methods}

\subsection*{Study overview and Methods roadmap}

\paragraph{Representation-learning problem} Let $\mathcal{X} \subset \mathbb{R}^{H \times W \times D}$ denote 3D medical volumes and $\mathcal{R}$ free-text radiology reports. Given paired data $\mathcal{D}=\{(x_i,r_i)\}_{i=1}^N$, with $x_i \in \mathcal{X}$ and $r_i \in \mathcal{R}$, GreenRFM aims to learn clinically grounded image and text representations for diagnosis, retrieval, multimodal fusion, and generation-oriented transfer.

\paragraph{MUST-guided pipeline roadmap} We use MUST as a shorthand for the empirical design principles tested in GreenRFM, not as a complete theory of RFM training. GreenRFM implements them as a two-stage image--report pre-training pipeline: LLMs distill noisy reports into structured diagnostic labels, these labels supervise the image and text encoders, and diagnostic supervision is retained during image--text alignment. The Methods follow the same order as the \hyperref[sec:principled_supervision]{Results} tests and Fig.~\ref{fig:method_overview}: more distilled supervision for label distillation and audit, ubiquitous and semantics-enforcing supervision for the two-stage training strategy, and task-aligning supervision for task-aligned design choices. The uncertainty-treatment and synthetic-noise protocols used to further test label distillation are specified in dedicated sections below.

\subsection*{MUST hypotheses, dependencies, and failure modes}

\paragraph{Principle M: more distilled supervision} LLM-distilled label supervision should remain effective when residual report-distillation errors are sufficiently small and predominantly conservative. Conservative errors include uncertain labels or abstentions for expert-positive findings, which can be masked or downweighted rather than used as definitive positive-negative inversions. This principle predicts useful silver-label training. It is unsupported if silver-label training fails relative to curated labels.

\paragraph{Principle U: Ubiquitous supervision} Holding architecture and data fixed, adding direct supervision to the vision encoder, text encoder, and aligned space should yield positive marginal gains over alignment-only training. The principle is unsupported if any added term is non-positive or if the full system underperforms the alignment-only baseline.

\paragraph{Principle S: semantics-enforcing supervision} Let $A_I$ and $A_T$ denote the unimodal diagnostic utility of the image and text encoders after Stage~1. For cross-modal transfer, post-alignment utility should be constrained by the weaker unimodal encoder: alignment can expose diagnostic structure already present in paired representations, but should not be expected to create stronger discrimination de novo. This reflects a relaxed data-processing intuition, not a necessary-and-sufficient theorem. Operationally, delaying alignment until after unimodal pre-training should yield lower Stage~1 classification loss and faster or lower Stage~2 alignment loss than earlier-alignment schedules. We test this prediction against a one-stage joint multi-task baseline; the principle is challenged if that baseline matches or exceeds sequential training on these loss-based comparisons.

\paragraph{Principle T: task-aligning supervision} Let $\Delta_{\mathrm{task}}$ denote train--inference mismatch in text domain, supervisory target, pooling/normalization strategy, and semantic focus of the aligned space. Zero-shot diagnostic utility should improve as $\Delta_{\mathrm{task}}$ decreases. Thus, radiology-specific text encoders, diagnostic labels, GAP, non-normalized alignment, and a shared diagnostic classifier should outperform mismatched alternatives. The principle is challenged if reducing these mismatches fails to improve, or harms, downstream diagnosis.

\paragraph{Inter-principle dependencies} More distilled supervision supplies the labels used by ubiquitous and semantics-enforcing supervision. The benefits of ubiquitous and semantics-enforcing supervision therefore depend on the distilled labels retaining enough task-relevant signal: if distillation noise is too large, explicit supervision need not outperform direct alignment. When label quality is adequate, semantics-enforcing pre-training enables ubiquitous supervision during alignment by ensuring that unimodal diagnostic structure is established before alignment. Task-aligning supervision reduces train--inference mismatch after supervision has injected those diagnostic signals.

\subsection*{Report-to-label distillation, audit, and label harmonization}

\paragraph{Structured LLM label extraction} To instantiate Principle~M, we distilled noisy radiology reports into structured diagnostic label vectors that serve as scalable silver-label supervision. A structured prompt template instructed the LLM to extract abnormality labels from each report. The system prompt defined each category and enforced a strict output format of either 18 or 30 comma-separated integers, requiring each condition to be marked as present (1), absent (0), or uncertain/missing (-1). Exact diagnostic label-extraction prompts are provided in Supplementary Appendix~\ref{supp:diagnostic_label_prompts}. Following prior radiology label-extraction practice that treats uncertainty as an explicit state~\cite{irvin2019chexpert}, we retained uncertain or missing findings as an explicit label state rather than forcing them into the negative class, allowing subsequent training to mask or confidence-weight ambiguous supervision. For the audited Merlin silver-label analyses, the primary GreenRFM analyses used the original Doubao-extracted labels; the Claude Opus~4.7 and audit-informed prompt variants were used only for the residual-noise reduction analysis.

\paragraph{Automatic validation and retry} Because malformed LLM outputs would create avoidable label noise, every response was automatically checked before use. The verifier required exactly 18 or 30 comma-separated tokens and required each token to be one of \{``1'', ``0'', ``-1''\}. If the check failed, the report was re-queried until a maximum number of attempts was reached; unresolved cases were flagged for manual review (Supplementary Fig.~\ref{fig:llm_workflow}). For example, when a report contained ``nodule in both Lungs'', the LLM could occasionally output ``2'' rather than ``1'' for the ``Nodule'' category, which triggered this validation step.

\paragraph{Institution-specific label-space harmonization} For \textbf{AH-Chest} and \textbf{AH-Abd}, we applied the same LLM-driven label extraction pipeline used for CT-RATE and Merlin, extracting 18 and 30 abnormality labels respectively. Labels were retained only when two independent LLMs (Qwen~\cite{bai2023qwen} and Doubao~\cite{doubao2023}) agreed on more than 90\% of cases. Abnormality categories were finalized in consultation with radiologists from AnHui Provincial Hospital, as some categories in the public datasets (e.g., ``Mosaic Attenuation Pattern'') are not routinely reported at this site.
The resulting institutional label distributions for AH-Chest and AH-Abd are reported in Supplementary Tables~\ref{tab:S2} and~\ref{tab:S3}.

For \textbf{AH-Knee} and \textbf{AH-Spine}, the same pipeline extracted 13 and 8 abnormality labels respectively, targeting conditions such as meniscal tears and ligament injuries (knee) and disc herniation, spinal stenosis, and vertebral fractures (spine). Categories were selected in consultation with radiologists. All labels in the held-out test set were manually reviewed. Since musculoskeletal MRI reports typically describe findings in detail, uncertain or missing labels ($-1$) were excluded.

\paragraph{Silver-label agreement and downstream utility} We evaluated the distilled labels in two ways, matching the Principle~M analyses in Results. First, we compared the LLM-derived labels with the official Merlin labels, reporting agreement metrics with uncertain cases excluded and inspecting the full three-state confusion pattern among present, absent, and uncertain labels. Second, we trained matched vision encoders using either LLM-derived labels or curated labels and compared their held-out diagnostic AUC, testing whether the silver labels retained sufficient signal for representation learning.

\paragraph{Linguistic-risk expert audit sampling} To audit the main failure modes of report distillation, we enriched the expert sample for linguistically risky contexts rather than sampling only at random. We first performed a linguistic risk-map analysis over all $115{,}076$ disease--sentence pairs extracted from $25{,}494$ Merlin reports, assigning each pair to one or more of six contextual risk classes: negation, hedge, history, follow-up, post-operative, and other. Pairs assigned to at least one class were designated high-risk ($\approx\!26\%$ of all pairs). We then drew a stratified audit sample from the Doubao-annotated Merlin label file and the corresponding CT report texts.

For each of the 30 abnormality categories, we independently sampled up to 8 LLM-positive ($+1$), 6 LLM-uncertain ($-1$), and 6 LLM-negative ($0$) instances, targeting approximately 20 samples per disease and 600 samples in total. A sample was eligible only if the original CT report contained at least one sentence matching the disease keyword with a length exceeding 10 characters; the matched sentence together with its immediate neighbours ($\pm1$ sentence, capped at 3 sentences and 600 characters in total) was extracted to provide the annotating radiologist with sufficient textual evidence. 
When a disease--class stratum contained fewer eligible samples than the target quota, all available samples were included (adaptive sampling). 

This procedure yielded $565$ samples spanning all 30 disease categories. A board-certified radiologist independently assigned each sample a label of present ($+1$), absent ($0$), or uncertain ($-1$) based on the extracted report context. (Supplementary Table~\ref{tab:audit_3x3}).

\paragraph{Audit-informed prompt tuning} To test whether residual label errors were reducible, we repeated label extraction on the expert-audit sample under three settings: the original Doubao labels used for the primary GreenRFM analyses, Claude Opus~4.7 with the same diagnostic-label prompt, and Claude Opus~4.7 with six linguistic rules derived from the Doubao audit failure patterns. This analysis was used to quantify model- and prompt-level label-noise reduction, not to replace the labels used in the main GreenRFM results.

\subsection*{Uncertainty-aware supervision ablations}

To evaluate the robustness of LLM-distilled supervision under different treatments of uncertain/missing ($-1$) labels, we compared five training strategies on the Merlin dataset, holding all other hyperparameters fixed.

\paragraph{Masked BCE baseline} Uncertain labels are excluded from the binary cross-entropy loss via a sample mask; only confirmed positive ($+1$) and negative ($0$) instances contribute gradients.

\paragraph{Equal-probability soft labels} Uncertain labels are mapped to $[0.5,0.5]$ and included in the loss with unit weight, treating clinical ambiguity as equal probability of presence and absence.

\paragraph{Three-class uncertainty modelling} Labels $\{-1,0,+1\}$ are treated as three discrete classes under a cross-entropy objective, expanding the per-disease head to three outputs.

\paragraph{Confidence-weighted BCE} Uncertain labels are mapped to the negative class (0) and included in the binary loss with a downweighting factor of 0.3, reducing the influence of ambiguous samples while preserving gradient signal.

\paragraph{Ordinal cumulative-label regression} Each pathology head outputs two cumulative logits predicting $P(y \geq \text{uncertain})$ and $P(y \geq \text{present})$, trained with the sum of two BCE losses against binary targets $(1,0)$ for uncertain and $(1,1)$ for present. This explicitly encodes the natural ordering absent $<$ uncertain $<$ present without treating uncertain as a hard binary label.

Each strategy was applied at Stage~1 and Stage~2 independently.

\subsection*{Synthetic label-noise stress test}

To separate the effect of clinically realistic uncertainty-type errors from arbitrary label corruption, we applied four noise regimes to the Merlin training labels in a controlled factorial design. 

\textbf{Confirmed-label symmetric flip.} Each confirmed positive ($+1$) or negative ($0$) label is independently flipped to the opposite class with a fraction $p \in \{0.00, 0.20, 0.40, 0.60, 0.80, 1.00\}$; uncertain ($-1$) labels are never flipped and always masked.

\textbf{Uncertain-to-negative ($u_{-}$).} A fraction $p$ of uncertain labels (where $p \in \{0.20, 0.40, 0.60, 0.80, 1.00\}$) are reassigned to negative (0), simulating conservative under-annotation.

\textbf{Uncertain-to-positive ($u_{+}$).} A fraction $p$ of uncertain labels (where $p \in \{0.20, 0.40, 0.60, 0.80, 1.00\}$) are reassigned to positive (+1), simulating aggressive over-annotation.

\textbf{Uncertain-to-random ($u_{\pm}$).} A fraction $p$ of uncertain labels (where $p \in \{0.20, 0.40, 0.60, 0.80, 1.00\}$) are reassigned to either negative (0) or positive (1) with equal probability (0.5 each).

Each condition was trained under the default masked-BCE loss, and evaluated on the clean Merlin held-out test set.

\subsection*{Two-stage supervised pre-training for ubiquitous and semantics-enforcing supervision}

Ubiquitous and semantics-enforcing supervision determine where and when the distilled labels enter training. We first use the structured labels to create discriminative unimodal encoders, then align the encoders while retaining diagnostic supervision in the shared space.

\textbf{Stage 1: supervised unimodal semantic learning.} We pre-train the image and text encoders independently before contrastive alignment. The vision encoder is supervised by the LLM-distilled diagnostic labels via binary cross-entropy (BCE) loss ($\mathcal{L}_{\text{BCE}}$); when anatomical segmentation masks are available (e.g., for CT datasets), a lightweight decoder additionally predicts coarse anatomical segmentation masks $M$~\cite{li2025supervisionscaling} (derived from TotalSegmentator~\cite{wasserthal2023totalsegmentator}) via cross-entropy loss ($\mathcal{L}_{\text{seg}}$), giving a total vision loss $\mathcal{L}_{\text{vision}} = \mathcal{L}_{\text{BCE}} + \mathcal{L}_{\text{seg}}$. For the text encoder, a classification head is attached to the \texttt{[CLS]} token embedding and trained with BCE ($\mathcal{L}_{\text{text}}$), encouraging the encoder to extract explicit diagnostic signals.

\textbf{Stage 2: non-normalized contrastive alignment with retained supervision.} In the second stage, we align the pre-trained encoders. Let $z_i$ and $t_i$ denote the projected image and text embeddings for the $i$th image--report pair. We optimize a joint objective combining the contrastive loss (without $L_2$ normalization) and the shared classifier loss. The contrastive loss is defined as:
\begin{equation}
	\mathcal{L}_{\text{CLIP}} = -\frac{1}{2N}\sum_{i=1}^N \left[\log\frac{\exp(z_i^\top t_i)}{\sum_{j=1}^N \exp(z_i^\top t_j)} + \log\frac{\exp(z_i^\top t_i)}{\sum_{j=1}^N \exp(z_j^\top t_i)}\right].
\end{equation}
The total objective is:
\begin{equation}
	\mathcal{L}_{\text{total}} = \mathcal{L}_{\text{CLIP}} + \lambda (\mathcal{L}_{\text{cls\_img}} + \mathcal{L}_{\text{cls\_txt}}),
\end{equation}
where $\mathcal{L}_{\text{CLIP}}$ is computed on non-normalized embeddings, and $\mathcal{L}_{\text{cls\_img}}$ and $\mathcal{L}_{\text{cls\_txt}}$ encourage image and text embeddings to remain discriminative for the diagnostic labels throughout alignment.

\textbf{Optimization details.} We use the AdamW optimizer with a learning rate of $1\times10^{-4}$ for supervised pre-training of the image encoder, $1\times10^{-5}$ for supervised pre-training of the text encoder, and $1\times10^{-5}$ for cross-modal alignment. The batch size was set to 10 for both stages and for both CT-RATE and Merlin, and positive/negative samples are weighted equally in the BCE loss.

\subsection*{Task-aligned architecture and alignment choices}

Task-aligning supervision treats downstream zero-shot diagnosis as a protocol-matching problem. We therefore aligned four dimensions of the pre-training design with the diagnostic inference setting: domain consistency through a radiology-specific text encoder, goal consistency through diagnostic labels, architectural and feature-geometry consistency through GAP and removal of $L_2$ normalization, and semantic-focus consistency through the shared classifier.

\textbf{Text-domain consistency: CXR-BERT.} We select CXR-BERT~\cite{boecking2022cxrbert} over generic clinical language models to maintain domain consistency. Unlike general biomedical models, CXR-BERT is pre-trained specifically on radiology reports with a specialized tokenizer, making the semantic space more closely aligned with radiology terminology than with general medical literature.

\textbf{Supervisory-goal consistency: diagnostic labels.} We prioritize diagnostic labels over generic visual descriptions to align the pre-training objective with the downstream diagnostic task. While visual descriptions (e.g., texture, shape) capture morphological details, they do not directly map to downstream tasks. By supervising the model with explicit diagnostic concepts (e.g., ``calcification''), we encourage the encoder to learn features associated with pathology, reducing the gap between the pre-training proxy task and the final diagnostic application.

\textbf{Vision-architecture consistency: 3D ResNet-18 and GAP.} We employ a 3D ResNet-18 (initialized with Kinetics-400 weights~\cite{kay2017kinetics}) as our vision encoder $f_I$. We offer two configurations to balance performance and resource efficiency. For the resource-efficient ``lightweight'' version, we modify the initial stem convolution to use a kernel size and stride of $(4, 4, 4)$, which downsamples the input volume to reduce computational overhead and memory usage. To maintain consistency with the alignment stage and downstream zero-shot evaluation---which operates on global volume representations---we use Global Average Pooling (GAP) to aggregate the 3D feature maps into a vector $z \in \mathbb{R}^c$:
\begin{equation}
	z = \text{GAP}(F) = \frac{1}{D'H'W'} \sum_{d,h,w} F_{d,h,w}.
\end{equation}
This prevents performance from degrading when training and inference protocols differ structurally.

\textbf{Feature-geometry consistency: no $L_2$ normalization.} Standard CLIP training enforces $L_2$ normalization, projecting features onto a hypersphere. However, diagnostic confidence may be encoded in feature magnitude~\cite{wang2017normface,meng2021magface}. Enforcing $L_2$ normalization during alignment would discard this potential confidence signal. To maintain consistency between pre-training and inference, we remove the $L_2$ normalization constraint in the contrastive loss calculation, allowing the model to preserve semantic magnitude.

\begin{equation}
	\text{sim}(z, t) = z^\top t = \|z\| \|t\| \cos\theta,
\end{equation}
where $z$ and $t$ represent the image and text embeddings, respectively. Unlike standard cosine similarity, which only captures angular distance ($\cos\theta$), this formulation allows the embedding norm to preserve the confidence or severity learned in supervised pre-training.

\textbf{Semantic-focus consistency: shared classifier.} To encourage the cross-modal alignment to serve the diagnostic objective, we introduce a shared linear classifier that operates on both the projected image and text embeddings. This anchors the alignment space to high-level pathological concepts, and is the task-alignment role of the retained classifier used during Stage~2.

\subsection*{Ethics approval and data governance}
This retrospective study was approved by the Institutional Review Board (IRB) of the participating hospitals, with informed consent waived. 

\subsection*{Cohorts, splits, and task roles}
In total, our study involves diverse modalities and anatomies. Datasets are organized by role: two large-scale public CT datasets serve as pre-training corpora; three CT datasets provide external generalization validation; and six datasets spanning CT and MRI support downstream task evaluation.
Cohort-level sample counts, patient counts where available, age distributions, and sex distributions are summarized in Supplementary Table~\ref{tab:S1}.
For foundation-model pre-training, the target population is the available image--report archive rather than a disease-specific clinical cohort; therefore, we used all available image--report pairs in the source datasets according to their official or predefined splits, without additional diagnosis-specific patient-level inclusion or exclusion criteria. 

\medskip\noindent\textbf{Pre-training corpora.}

\textbf{CT-RATE}~\cite{hamamci2024ctrate} serves as a pre-training and validation corpus, containing 50,188 reconstructed volumes, radiology reports and 18 abnormality labels from 25,692 scans of 21,304 patients. We follow the official training/validation splits.

\textbf{Merlin}~\cite{blankemeier2024merlin} is a high-quality abdominal CT dataset from Stanford University Medical Center, containing 25,528 scans from 18,321 patients. It provides labels for 30 abnormalities, including explicit markers for missing or uncertain findings. We report results on the official held-out test set (5,137 scans).

\medskip\noindent\textbf{External CT generalization cohorts.}

We use \textbf{RAD-ChestCT}~\cite{draelos2021radchestct} and two institutional CT cohorts~(\textbf{AH-Chest} and \textbf{AH-Abd}) as external validation sets.

The \textbf{RAD-ChestCT} dataset comprises 36,316 non-contrast chest CTs from Duke University (2012-2017). We use the publicly available subset of 3,630 scans to test generalization to unseen distributions.

\textbf{AH-Chest} and \textbf{AH-Abd} are plain chest CT and contrast-enhanced abdominal CT scans collected from The First Affiliated Hospital of the University of Science and Technology of China~(AnHui Provincial Hospital) between August 2021 and June 2023, containing 29,377 and 14,038 scans respectively. The AH-Chest contains 63,378 reconstructed images.
The plain chest CT scans usually contain 1--3 images, with raw data reconstructed at different slice thicknesses ($0.625\,\mathrm{mm}$, $5\,\mathrm{mm}$) or different reconstruction kernels (sharp kernel, smooth kernel). We retain all reconstructed images for analysis.
The contrast-enhanced abdominal CT scans usually contain a plain phase and a contrast-enhanced phase, with many containing more than one contrast-enhanced phase. For examinations with only two phases, we select the contrast-enhanced phase for analysis. For examinations with more than two phases, we select the portal venous phase if available; otherwise, we select the arterial phase. All scans are de-identified before processing and analysis.
Scanner manufacturers, scanner models, slice thicknesses, and tube-voltage ranges for the institutional cohorts are reported in Supplementary Table~\ref{tab:S4}.

\medskip\noindent\textbf{Downstream transfer cohorts.}

\textbf{Merlin 5-year disease prediction subset.} For clarity, we refer to the longitudinal subset of the Merlin dataset used for incident-disease prediction as the Merlin 5-year disease prediction subset. It comprises $2{,}488$ scans from patients with five-year clinical follow-up records, split into training ($n\!=\!656$), validation ($n\!=\!589$), and test ($n\!=\!1{,}243$) sets. Each scan is paired with incident-disease labels for six endpoints: cardiovascular disease (CVD), ischaemic heart disease (IHD), hypertension (HTN), diabetes mellitus (DM), chronic kidney disease (CKD), and osteoporosis (OST). Disease status is encoded as an ordinal variable ($0=$ no disease at baseline or follow-up; $1=$ incident within five years; $2$--$3=$ prevalent disease at baseline); subjects with prevalent disease at baseline (labels $2$--$3$) are masked from training and evaluation to isolate incident-prediction from detection of pre-existing pathology. Preprocessing follows the abdominal CT protocol described for Merlin.

\textbf{AH-Knee} and \textbf{AH-Spine} are musculoskeletal MRI datasets concurrently acquired from the same institution. \textbf{AH-Knee} comprises 3,662 studies (15,431 images) focusing on knee pathologies, while \textbf{AH-Spine} includes 9,203 studies (28,260 images) targeting spinal conditions. Each examination typically consists of multi-sequence (PD, T1, T2) and multi-planar (Sagittal, Coronal, Transverse) acquisitions, though specific sequences or planes may be absent in some cases. Each dataset is split in a 9:1 ratio for training and validation.

\textbf{RadGenome-ChestCT}~\cite{zhang2025development} is a region-guided chest CT report-generation dataset built on CT-RATE. It provides 25,692 region-grounded CT--report pairs from 21,304 patients, partitioned into a training split of 24,128 pairs and an evaluation split of 1,564 pairs. Each volume includes anatomical segmentation masks for 10 thoracic structures---abdomen, bones, breasts, oesophagus, heart, lungs, trachea and bronchi, mediastinum, pleura, and thyroid, and region-aligned diagnostic descriptions. This dataset is used for the report-generation encoder-swap analysis.

\textbf{XN-Liver} is an institutional multi-sequence liver MRI dataset collected from XiNan Hospital (Southwest Hospital), comprising 190 patients who underwent hepatectomy, split into training ($n\!=\!114$) and held-out test ($n\!=\!76$) sets (MVI-positive rate $\approx 43\%$). MVI status is confirmed by post-operative histopathological examination. For each patient, a single pre-operative examination is selected: seven hepatobiliary MRI sequences stored as independent NIfTI files (Table~\ref{tab:mvi_channels}).

\begin{table}[htbp]
\centering

\caption{\textbf{Seven hepatobiliary MRI sequences constituting the multi-channel input for HCC MVI prediction.} Channels are stacked in fixed order to produce a $(7, 64, 224, 224)$ input tensor.}
\label{tab:mvi_channels}
\small
\setlength{\tabcolsep}{5pt}
\begin{tabular}{@{}clll@{}}
\toprule
Ch. & Sequence & Phase & Description \\
\midrule
0 & T1 VIBE-FS (2.5\,mm) & Unenhanced & T1-w, fat-suppressed, thin-slice \\
1 & Arterial & Contrast-enhanced & T1, early arterial phase \\
2 & T1 VIBE-FS (5\,min) & Contrast-enhanced & T1, portal-venous / hepatobiliary phase \\
3 & T1 VIBE-FS (15\,min, FLIP~30\textdegree) & Contrast-enhanced & T1, delayed / hepatobiliary phase \\
4 & T2 HASTE (axial, p2) & Unenhanced & T2-weighted, breath-hold \\
5 & DWI ($b\!=\!50/400/800\,\mathrm{s/mm^2}$) & Unenhanced & Diffusion-weighted imaging \\
6 & ADC map & Derived & Apparent diffusion coefficient \\
\bottomrule
\end{tabular}
\end{table}

\textbf{WAW-TACE} is a public four-phase contrast-enhanced abdominal CT dataset comprising 229 patients with hepatocellular carcinoma who underwent trans-arterial chemoembolization. The released cohort was analysed with a training partition ($n\!=\!137$) and a held-out test set for downstream evaluation ($n\!=\!92$; PFS subset $n\!=\!43$). The binary endpoint was coded as label 1 for LR-TR equivocal or viable disease, indicating non-complete response with residual viable or equivocal tumour, and as label 0 for LR-TR nonviable disease, indicating complete response. The label-1/non-complete-response rate was $154/229$ ($67.0\%$) in the full cohort and $62/92$ ($67.4\%$) in the held-out test set. The four pre-TACE CT phases are: unenhanced, arterial, portal-venous, and delayed. TACE response labels are determined by mRECIST-based post-TACE imaging review~\cite{lencioni2010modified}. Overall survival (OS) endpoints were available for all $92$ held-out test cases; progression-free survival (PFS) endpoints were available for the progression-annotated subset ($n\!=\!43$).

\subsection*{Image preprocessing by cohort}
For \textbf{CT-RATE}, \textbf{RAD-ChestCT} and \textbf{AH-Chest}, all chest CT volumes are resampled to an in-plane spacing of $1.5\,\mathrm{mm} \times 1.5\,\mathrm{mm}$ and slice thickness of $3.0\,\mathrm{mm}$ via trilinear interpolation. Hounsfield Units are clipped to a lung-relevant window ([-1000, 200] HU) and linearly normalized to the range [-1, 1]. Volumes are first padded or cropped to a fixed canvas of $240 \times 240 \times 120$ voxels, then randomly cropped to $192 \times 192 \times 96$ during training. At evaluation time, we use a single center crop of size $192 \times 192 \times 96$.

For the abdominal \textbf{Merlin} dataset, we adopt a similar pipeline but with abdomen-appropriate settings. Volumes are resampled to $1.5\,\mathrm{mm} \times 1.5\,\mathrm{mm} \times 3.0\,\mathrm{mm}$, intensities are clipped to a wider range ([-1000, 1000] HU) and normalized to [-1, 1]. Each volume is padded or cropped to $280 \times 280 \times 180$, followed by random crops of $224 \times 224 \times 144$ for training and a center crop of the same size for evaluation.

For AH-Abd, as the images in AnHui Provincial Hospital are mostly reconstructed with $5\,\mathrm{mm}$ slice thickness, we resample the volumes to an in-plane spacing of $1.5\,\mathrm{mm} \times 1.5\,\mathrm{mm}$ and slice thickness of $5.0\,\mathrm{mm}$. The other preprocessing steps are the same as in the Merlin dataset. CT scans with fewer than 50 slices are excluded.

For the MRI datasets (\textbf{AH-Knee} and \textbf{AH-Spine}), we adopt a multi-view multi-sequence preprocessing pipeline. We group images by plane (Sagittal, Coronal, Transverse) and sequence (PD, T1, T2). Images within the same plane are resampled to match the spacing of the PD sequence. We compute a joint bounding box using Otsu thresholding and crop the volumes with a margin. Intensity values are normalized using Z-score normalization based on foreground voxels. Finally, the images are padded to a fixed size and resized to $240 \times 240 \times 24$. During training, we apply random cropping of size $192 \times 192 \times 20$.

For \textbf{RadGenome-ChestCT}, preprocessing is applied separately to the whole-volume input and to each region-masked input. Whole-volume CT volumes are foreground-cropped using a threshold of $-1000$ HU to remove background air and resized to $256 \times 256 \times 64$ voxels. To match each encoder's original training distribution, the HU window is encoder-specific: GreenRFM volumes are clipped to $[-1000, 200]$ HU and normalized as $(x + 400) / 600$, whereas CT-CLIP volumes are clipped to $[-1000, 1000]$ HU and linearly mapped to $[-1, 1]$. For each of the 10 anatomical region inputs, the corresponding SAT-derived binary mask is first applied to the CT volume, with voxels outside the mask set to $-1024$ HU, before the same foreground-crop, resize, clip, and normalization pipeline is applied. Regions with entirely zero masks and samples lacking region-specific report text are excluded from training and evaluation.

For the \textbf{XN-Liver} multi-sequence MRI (HCC MVI prediction), each of the seven sequences (Table~\ref{tab:mvi_channels}) is preprocessed independently. Volumes are spatially resampled from $(1.25, 1.25, 2.5)\,\mathrm{mm}$ to a target spacing of $(1.5, 1.5, 3.0)\,\mathrm{mm}$ via linear interpolation. Each volume is normalized via percentile clipping: the 1st and 99th percentiles ($p_1$, $p_{99}$) are computed over non-zero voxels; values are clipped to $[p_1, p_{99}]$ and linearly remapped to $[-1, 1]$. Missing sequences are filled with $-1.0$ at the reference shape. The seven channels are stacked into a single tensor. Volumes are padded to $(80, 280, 280)$ voxels and randomly cropped to $(64, 224, 224)$ during training; at inference, a center crop of $(64, 224, 224)$ is applied.

For the \textbf{WAW-TACE} four-phase CT (TACE analysis), each phase is resampled to $(1.5, 1.5, 3.0)\,\mathrm{mm}$ via linear interpolation. CT HU values are clipped to $[-1000, 1000]$ and linearly mapped to $[-1, 1]$. Missing phases are filled with $-1.0$ at the reference shape. The four phases (unenhanced, arterial, portal-venous, delayed) are stacked in fixed order. Volumes are padded to $(160, 280, 280)$ and randomly cropped to $(144, 224, 224)$ during training; at inference, a center crop of $(144, 224, 224)$ is used.

\subsection*{Evaluation metrics and clinical endpoints}
We used task-specific metrics throughout; multi-label results were macro-averaged across disease categories unless otherwise stated, and 95\% confidence intervals were obtained by non-parametric bootstrap resampling of held-out predictions.

\paragraph{Diagnostic classification and zero-shot diagnosis} Classification performance was assessed by the area under the receiver operating characteristic curve (AUC), accuracy, precision, recall (sensitivity), specificity, and F1-score. This metric set was applied uniformly across zero-shot diagnosis, downstream clinical prediction, and site-adaptation analyses. For zero-shot diagnosis, the probability of disease $c$ given image $I$ was computed as a softmax over dot-product similarities between the volume embedding $v=f_I(I)$ and paired positive and negative prompt embeddings $t_c^+=f_T(T_c^+)$, $t_c^-=f_T(T_c^-)$:
\begin{equation}
	P(c\mid I) = \frac{\exp(\mathrm{sim}(v, t_c^+))}{\exp(\mathrm{sim}(v, t_c^+)) + \exp(\mathrm{sim}(v, t_c^-))},
\end{equation}
where $\mathrm{sim}(v,t)=v^\top t$. Macro AUC served as the primary summary metric in long-tail and data-efficiency analyses. Agreement between LLM-derived labels and expert annotations was additionally quantified by Cohen's $\kappa$. Uncertainty-aware training experiments further reported expected calibration error (ECE) \cite{guo2017calibration}, and specificity at sensitivity $\geq 0.95$ (Spec@Sens$_{0.95}$).

\paragraph{Retrieval and representation metrics} Retrieval and representation quality were evaluated through several complementary measures. Report-to-volume retrieval used Recall@$K$ for $K\in\{1,5,10,50,100\}$, defined as the fraction of query reports whose paired volume ranked in the top-$K$ candidates under similarity $\mathrm{sim}(t_i,v_j)$, where $t_i=f_T(T_i)$ and $v_j=f_I(I_j)$. Precision@$K$ was used for concept-level retrieval probes and within-label attribute retrieval, measuring the fraction of top-$K$ retrieved items satisfying the queried concept or attribute group. Within-label attribute alignment was further characterised by the Spearman rank correlation ($\rho$) between attribute-dominant embedding dimensions and the full-embedding neighbour order. Representation geometry was characterised by centred kernel alignment (CKA) \cite{kornblith2019similarity}.

\paragraph{Report-generation metrics} Report generation quality was measured by standard natural-language generation metrics--BLEU-4~\cite{papineni2002bleu}--computed against reference radiology reports. Clinical content fidelity was assessed by passing generated reports through a frozen RadBERT classifier~\cite{yan2022radbert} and computing micro-averaged F1-score over the 18 disease categories.

\paragraph{Survival and decision-curve analysis} Cox proportional-hazards models \cite{cox1972regression} were evaluated by Harrell's C-index~\cite{harrell1996multivariable} with bootstrap 95\% confidence intervals, and Kaplan--Meier analyses \cite{kaplan1958nonparametric} were summarised by median overall survival (OS) and progression-free survival (PFS) with log-rank $p$-values. Clinical utility for TACE-response prediction was quantified by decision-curve analysis (DCA) \cite{vickers2006decision}, reporting net benefit over the prespecified threshold range $[0.05,\,0.85]$ and the corresponding integrated net benefit.

\subsection*{Grad-CAM visualization}
To interpret model predictions qualitatively, we use Gradient-weighted Class Activation Mapping (Grad-CAM \cite{selvaraju2017grad}). Given an input volume $I$ and a target diagnostic prompt $T$ describing a specific pathology (e.g., ``There is {pathology}.''), we compute the similarity score $S = \text{sim}(f_I(I), f_T(T))$.
We then backpropagate the gradients of this score with respect to the feature maps $A^k$ of the last convolutional layer (layer4) of the 3D ResNet-18 image encoder.
The importance weights $\alpha_k$ for each feature map $k$ are obtained by global average pooling of the gradients:
\begin{equation}
	\alpha_k = \frac{1}{Z} \sum_{i} \sum_{j} \sum_{l} \frac{\partial S}{\partial A_{ijl}^k},
\end{equation}
where $Z$ is the number of elements in the feature map. The localization map $L_{\text{Grad-CAM}}$ is then computed as a weighted combination of the feature maps, followed by a ReLU activation to isolate features that have a positive influence on the target class:
\begin{equation}
	L_{\text{Grad-CAM}} = \text{ReLU}\left(\sum_k \alpha_k A^k\right).
\end{equation}
Finally, the resulting low-resolution heatmap is upsampled via trilinear interpolation to the original input resolution and overlaid on the CT or MRI volume for visualization. This reveals whether the model's attention aligns with clinically significant regions identified by radiologists. Visualization of heatmap is provided in Fig. \ref{fig:S3} and Fig. \ref{fig:S4}.

\subsection*{Semantic-richness and representation-probing suite}

We conducted a coordinated probing suite on the CT-RATE validation set ($n\!=\!3{,}039$), comparing GreenRFM against a CLIP-style baseline trained under identical architecture and data. All probing analyses used frozen embeddings unless otherwise specified.

\textbf{Long-tail finding recognition.} Fifteen long-tail pathologies were assembled from CT-RATE annotations: pneumonia, atherosclerosis, spondylosis, pneumothorax, bronchial thickening, osteophyte, empyema, lung cyst, lung mass, pericardial calcification, pneumomediastinum, pulmonary fibrosis, tuberculosis, pleural calcification, and rib fracture. Zero-shot diagnosis was performed using positive/negative prompt pairs. 

\textbf{Within-label attribute retrieval and ranking drivers.} For each of the 18 diagnostic labels, image-to-image nearest-neighbour retrieval (cosine similarity) was performed within the positive cases for that label. Retrieved sets were scored against three LLM-derived visual attribute groups (density, morphology, distribution) extracted independently from the reports using the same LLM pipeline described above. Attribute consistency was quantified as the Top-$K$ attribute match rate and its lift above a prevalence-weighted random baseline.

To characterise the embedding dimensions driving within-label ordering, each of the 768 embedding dimensions was assigned a diagnostic-dominance score and an attribute-dominance score. The diagnostic-dominance score $r_{\mathrm{diag}}(d)$ of dimension $d$ is the maximum absolute point-biserial correlation between that dimension and any of the 18 diagnostic labels; the attribute-dominance score $r_{\mathrm{attr}}(d)$ is defined analogously over the 11 visual attribute labels:
\begin{equation}
	r_{\mathrm{diag}}(d) = \max_{k=1,\ldots,18}\bigl|r_{\mathrm{pb}}(\mathbf{z}_d,\, y_k)\bigr|, \quad
	r_{\mathrm{attr}}(d) = \max_{j=1,\ldots,11}\bigl|r_{\mathrm{pb}}(\mathbf{z}_d,\, a_j)\bigr|.
\end{equation}
A dimension is classified as \emph{attribute-dominant} if $r_{\mathrm{attr}}(d) > r_{\mathrm{diag}}(d)$, and as \emph{diagnosis-dominant} otherwise. The Spearman rank correlation ($\rho$) between the attribute-dominant subset of embedding dimensions and the within-label nearest-neighbour ordering was then computed to quantify how much fine-grained attribute structure contributes to intra-class retrieval.

\textbf{Beyond-label concept retrieval.} A vocabulary of 24 clinical concepts absent from the 18-label training set was curated across five categories: anatomical location (5), temporal change (4), post-operative/device findings (5), severity (4), and texture/appearance (6). Each concept was used as a zero-shot text query for image retrieval; performance was measured as Precision@$K$.

\textbf{Attribute probing and representation geometry.} Linear classifiers were trained on the full CT-RATE training set ($n\!=\!47{,}149$) and evaluated on the validation set. Centred kernel alignment (CKA) and mutual information estimates additionally characterised the geometry of attribute encoding. The mutual information was calculated as $ \hat{I}(X;y) = H(y) - H(y\mid X) $, with $H(y\mid X)$ estimated from calibrated classifier probabilities.

\textbf{Report-generation transfer with frozen Reg2RG encoder swap.}
For the encoder-swap analysis, the GreenRFM visual encoder was frozen and inserted into a Reg2RG-style~\cite{chen2025large} report generator. Only the projection, resampler, mask encoder, and LoRA components were trained, while the visual backbone and token embeddings were held fixed. The generator backbone is LLaMA-2-7B~\cite{touvron2023llama2} with low-rank adaptation (LoRA; $r\!=\!8$, $\alpha\!=\!32$, dropout $0.1$); visual features from the frozen encoder are projected by a linear layer into a Perceiver Resampler~\cite{alayrac2022flamingo} (6 transformer layers, 32 latent queries) before being fused into the language model.

\subsection*{Site-adaptation intervention ladder}

A five-step intervention ladder was evaluated on the AH-Abd cohort to decompose the performance gap between the universally pre-trained model and the target site. Ternary per-category labels (positive/negative/uncertain) were extracted from free-text reports with the LLM pipeline used for Merlin; uncertain labels were masked from all losses and excluded from classifier fitting. Subjects were sorted by identifier, shuffled and split 80/20 into training ($n\!=\!11{,}230$) and validation ($n\!=\!2{,}808$); the same split was used across all conditions. Performance is reported as macro-AUC over the 30 categories
with 95\% confidence intervals from $B\!=\!1{,}000$ stratified bootstrap resamples.

\textbf{Universal zero-shot baseline.} The universal GreenRFM checkpoint was applied without adaptation and re-evaluated on the held-out AH-Abd validation subset. For each category, positive- and negative-template queries were embedded in the shared 768-d latent space, and the softmax over their dot product similarities with the image embedding yielded the predicted probability.

\textbf{BN-affine covariate adaptation.} Only the affine parameters ($\gamma,\beta$) of the 20 batch-normalization layers in the R3D-18 encoder were updated; all other parameters, including BN running statistics, were frozen to preserve the source-calibrated feature
geometry. An auxiliary linear head ($\mathbb{R}^{512}\!\to\!\mathbb{R}^{30}$) provided supervision and was discarded after adaptation. Training used a masked binary cross-entropy loss,
optimized with AdamW (lr $10^{-3}$, weight decay $10^{-4}$), after which zero-shot performance was re-evaluated.

\textbf{Frozen-embedding linear probing.} Linear classifiers were fit on frozen embeddings using AH-Abd training labels, applied separately to the unadapted and BN-adapted encoders so that covariate- and label-adaptation gains could be decomposed additively.

\textbf{Site-specific two-stage retraining.} The GreenRFM pipeline was trained from scratch on AH-Abd. Stage~1 trained the visual encoder and text encoder under $\mathcal{L}_{\mathrm{mBCE}}$. Stage~2 jointly trained the full model by combining symmetric image--text contrastive learning with independent latent-classifier supervision on both modalities.

\textbf{Checkpoint-initialized fine-tuning.} The same GreenRFM pipeline was trained with weights initialized from the Merlin-pretrained checkpoint.

\subsection*{Translational downstream fine-tuning protocols}

\textbf{Long-horizon disease prediction (Merlin 5-year disease prediction subset).} A linear classification head was trained atop the frozen or fine-tuned backbone under a masked BCE objective; the per-sample mask excludes subjects with prevalent disease at baseline (raw label $\geq 2$) to prevent the task from collapsing into detection of existing disease. 

\textbf{HCC microvascular-invasion prediction.} For image-only fine-tuning, the GreenRFM backbone was adapted to the seven-channel MRI input via stem channel inflation from one to seven channels. For multimodal fine-tuning, structured clinical templates encoding laboratory values and patient demographics were tokenised by the aligned text encoder; image and text embeddings were concatenated before the final classifier head. 

\textbf{TACE response, decision curves, and survival analysis.} The GreenRFM image encoder was adapted to the four-phase CT input via stem channel inflation from one to four channels and fine-tuned for binary non-complete-response prediction, using the label definition above. Response AUC and decision-curve analysis (DCA) were evaluated on the held-out test set ($n\!=\!92$; PFS subset $n\!=\!43$); DCA was conducted over the threshold probability range $[0.05,\,0.85]$ and summarised as integrated net benefit. For survival association analysis, fixed $512$-dimensional image embeddings from the fine-tuned model were projected onto 20 principal components (PCA-20). Cox proportional-hazards models were fitted on the held-out WAW-TACE evaluation cohort using PCA-20 features, clinical scores, or their combinations as covariates. Because the image features were extracted from a fixed GreenRFM encoder without exposure to survival labels, this analysis did not leak survival labels into the learned representation. Kaplan--Meier curves were stratified by the true complete-response status; log-rank $p$-values were computed for OS and PFS on the corresponding cohorts. The four established clinical scores (HAP, mHAP-2, ALBI-TAE, 6-and-12) were obtained from the provided WAW-TACE clinical metadata and used as comparators throughout.

\section*{Data availability}
The CT-RATE dataset is available at \url{https://huggingface.co/datasets/ibrahimhamamci/CT-RATE}. The RAD-ChestCT dataset is available at \url{https://zenodo.org/records/6406114}. The Merlin dataset is available at \url{https://stanfordaimi.azurewebsites.net/datasets/60b9c7ff-877b-48ce-96c3-0194c8205c40}. The WAW-TACE dataset is available at \url{https://zenodo.org/records/12741586}. The RadGenome-ChestCT dataset is available at \url{https://huggingface.co/datasets/RadGenome/RadGenome-ChestCT}. The private datasets (AH-Chest, AH-Abd, AH-Knee, AH-Spine and XN-Liver) are not publicly available due to patient privacy restrictions but may be available from the corresponding author upon reasonable request and institutional approval.

\section*{Code availability}
The code is available on GitHub at \url{https://github.com/SadVoxel/GreenRFM}.

\bibliographystyle{unsrt}
{\small\bibliography{references}}

\newpage

\appendix
\setcounter{table}{0}
\renewcommand{\thetable}{S\arabic{table}}
\setcounter{figure}{0}
\renewcommand{\thefigure}{S\arabic{figure}}
\newcommand{\suppTableFont}{\small}
\newcommand{\suppWideTableFont}{\footnotesize}
\newcommand{\suppTableSetup}{\setlength{\tabcolsep}{4pt}\renewcommand{\arraystretch}{1.12}}
\newcolumntype{L}[1]{>{\raggedright\arraybackslash}p{#1}}
\section*{Appendix}
\setcounter{secnumdepth}{3}
\setcounter{subsection}{0}
\renewcommand{\thesubsection}{\Alph{subsection}}
\renewcommand{\thesubsubsection}{\thesubsection.\arabic{subsubsection}}
\titleformat{\subsection}
  {\bfseries\fontsize{13}{14}\selectfont}
  {\thesubsection.}
  {0.35em}
  {#1}
  []
\titleformat{\subsubsection}[runin]
  {\bfseries\fontsize{10}{11}\selectfont}
  {\thesubsubsection.}
  {0.35em}
  {#1}
  [.]

\subsection{Demographic statistics of datasets}
We summarize cohort-level demographic information for the primary pre-training, external-validation, and institutional adaptation datasets in Table \ref{tab:S1}. These details include the number of scans or studies, patient counts where available, age distributions, and sex distributions.

\begin{table}[htbp]
    \centering
    \caption{\textbf{Demographic statistics of the primary datasets used in this study.} Note: M = male, F = female. Values are presented as mean $\pm$ s.d.\ or count (\%). One AH-Chest scan had missing sex metadata.}
    \label{tab:S1}
    \suppTableFont\suppTableSetup
    \begin{tabular}{@{}llrrrr@{}}
        \toprule
        Dataset & Modality & \# Scans & \# Patients & Age (Years) & Sex (M/F) \\
        \midrule
        CT-RATE (training/validation) & Chest CT & 25,692 & 21,304 & - & $29238 / 20944$ \\
        Merlin (training/validation) & Abd CT & 25,528 & 18,321 & $53.8 \pm 19.5$ & $10254 / 8065$ \\
        RAD-ChestCT (external validation) & Chest CT & 3,630 & - & - & - \\
        AH-Chest (institutional) & Chest CT & 29,377 & - & $51.6 \pm 16.3$ & $15160 / 14216$ \\
        AH-Abd (institutional) & Abd CT & 14,038 & - & $58.6 \pm 13.5$ & $7575 / 6463$ \\
        AH-Knee (institutional) & Knee MRI & 3,662 & - & $45.3 \pm 18.3$ & $1461 / 2201$ \\
        AH-Spine (institutional) & Spine MRI & 9,203 & - & $51.3 \pm 17.3$ & $3847 / 5356$ \\
        \bottomrule
    \end{tabular}
\end{table}

\FloatBarrier
\subsection{Label distribution analysis}
The distribution of abnormality labels varies significantly across the datasets, reflecting different clinical populations and acquisition protocols. Table \ref{tab:S2} and Table \ref{tab:S3} illustrate the prevalence of each pathology in the respective datasets.

\begin{table}[htbp]
\centering
\caption{Label distribution for AH-Chest dataset. The dataset contains 29,377 samples in total. The counts and percentages for Uncertain/missing (-1), Negative (0), and Positive (1) labels are shown.}
\label{tab:S2}
\suppTableFont\suppTableSetup
\begin{tabular}{@{}lrrr@{}}
\toprule
Label & Uncertain/missing (-1) & Negative (0) & Positive (1) \\
\midrule
Medical Material & 19349 (65.86\%) & 6709 (22.84\%) & 3319 (11.30\%) \\
Arterial Wall Calcification & 25351 (86.30\%) & 3174 (10.80\%) & 852 ( 2.90\%) \\
Cardiomegaly & 22832 (77.72\%) & 5749 (19.57\%) & 796 ( 2.71\%) \\
Pericardial Effusion & 24896 (84.75\%) & 3463 (11.79\%) & 1018 ( 3.47\%) \\
Coronary Artery Wall Calcification & 25797 (87.81\%) & 2956 (10.06\%) & 624 ( 2.12\%) \\
Hiatal Hernia & 27091 (92.22\%) & 2183 ( 7.43\%) & 103 ( 0.35\%) \\
Lymphadenopathy & 195 ( 0.66\%) & 25561 (87.01\%) & 3621 (12.33\%) \\
Emphysema & 22399 (76.25\%) & 3775 (12.85\%) & 3203 (10.90\%) \\
Atelectasis & 24652 (83.92\%) & 4316 (14.69\%) & 409 ( 1.39\%) \\
Lung Nodule & 3210 (10.93\%) & 3978 (13.54\%) & 22189 (75.53\%) \\
Lung Opacity & 13026 (44.34\%) & 6437 (21.91\%) & 9914 (33.75\%) \\
Pulmonary Fibrotic Sequela & 22873 (77.86\%) & 2707 ( 9.21\%) & 3797 (12.93\%) \\
Pleural Effusion & 338 ( 1.15\%) & 26866 (91.45\%) & 2173 ( 7.40\%) \\
Mosaic Attenuation Pattern & 27800 (94.63\%) & 1546 ( 5.26\%) & 31 ( 0.11\%) \\
Peribronchial Thickening & 27089 (92.21\%) & 1931 ( 6.57\%) & 357 ( 1.22\%) \\
Consolidation & 23727 (80.77\%) & 5118 (17.42\%) & 532 ( 1.81\%) \\
Bronchiectasis & 25752 (87.66\%) & 2627 ( 8.94\%) & 998 ( 3.40\%) \\
Interlobular Septal Thickening & 26970 (91.81\%) & 2229 ( 7.59\%) & 178 ( 0.61\%) \\
\bottomrule
\end{tabular}
\end{table}

\begin{table}[htbp]
\centering
\caption{Label distribution for AH-Abd dataset. The dataset contains 14,038 samples in total. The counts and percentages for Uncertain/missing (-1), Negative (0), and Positive (1) labels are shown.}
\label{tab:S3}
\suppTableFont\suppTableSetup
\begin{tabular}{@{}lrrr@{}}
\toprule
Label & Uncertain/missing (-1) & Negative (0) & Positive (1) \\
\midrule
Submucosal Edema & 13870 (98.80\%) & 74 ( 0.53\%) & 94 ( 0.67\%) \\
Renal Hypodensities & 8213 (58.51\%) & 1874 (13.35\%) & 3951 (28.15\%) \\
Aortic Valve Calcification & 12570 (89.54\%) & 1439 (10.25\%) & 29 ( 0.21\%) \\
Coronary Calcification & 12562 (89.49\%) & 1365 ( 9.72\%) & 111 ( 0.79\%) \\
Thrombosis & 12233 (87.14\%) & 1454 (10.36\%) & 351 ( 2.50\%) \\
Metastatic Disease & 9076 (64.65\%) & 847 ( 6.03\%) & 4115 (29.31\%) \\
Pancreatic Atrophy & 7984 (56.87\%) & 5618 (40.02\%) & 436 ( 3.11\%) \\
Renal Cyst & 3596 (25.62\%) & 1505 (10.72\%) & 8937 (63.66\%) \\
Osteopenia & 13618 (97.01\%) & 395 ( 2.81\%) & 25 ( 0.18\%) \\
Surgically Absent Gallbladder & 7765 (55.31\%) & 4805 (34.23\%) & 1468 (10.46\%) \\
Atelectasis & 9214 (65.64\%) & 3982 (28.37\%) & 842 ( 6.00\%) \\
Abdominal Aortic Aneurysm & 10964 (78.10\%) & 3045 (21.69\%) & 29 ( 0.21\%) \\
Anasarca & 11587 (82.54\%) & 2432 (17.32\%) & 19 ( 0.14\%) \\
Hiatal Hernia & 11512 (82.01\%) & 2402 (17.11\%) & 124 ( 0.88\%) \\
Lymphadenopathy & 533 ( 3.80\%) & 8776 (62.52\%) & 4729 (33.69\%) \\
Prostatomegaly & 9365 (66.71\%) & 3403 (24.24\%) & 1270 ( 9.05\%) \\
Biliary Ductal Dilation & 3941 (28.07\%) & 8767 (62.45\%) & 1330 ( 9.47\%) \\
Cardiomegaly & 8926 (63.58\%) & 4980 (35.48\%) & 132 ( 0.94\%) \\
Splenomegaly & 1776 (12.65\%) & 11179 (79.63\%) & 1083 ( 7.71\%) \\
Hepatomegaly & 2311 (16.46\%) & 11445 (81.53\%) & 282 ( 2.01\%) \\
Atherosclerosis & 10938 (77.92\%) & 2564 (18.26\%) & 536 ( 3.82\%) \\
Ascites & 7435 (52.96\%) & 4517 (32.18\%) & 2086 (14.86\%) \\
Pleural Effusion & 4552 (32.43\%) & 8053 (57.37\%) & 1433 (10.21\%) \\
Hepatic Steatosis & 6801 (48.45\%) & 6232 (44.39\%) & 1005 ( 7.16\%) \\
Appendicitis & 6379 (45.44\%) & 7620 (54.28\%) & 39 ( 0.28\%) \\
Gallstones & 3355 (23.90\%) & 9166 (65.29\%) & 1517 (10.81\%) \\
Hydronephrosis & 8809 (62.75\%) & 4623 (32.93\%) & 606 ( 4.32\%) \\
Bowel Obstruction & 8956 (63.80\%) & 4943 (35.21\%) & 139 ( 0.99\%) \\
Free Air & 9157 (65.23\%) & 4788 (34.11\%) & 93 ( 0.66\%) \\
Fracture & 9062 (64.55\%) & 4610 (32.84\%) & 366 ( 2.61\%) \\
\bottomrule
\end{tabular}
\end{table}

\FloatBarrier
\subsection{Evaluation of LLM-extracted labels}

\begin{table}[htbp]
    \centering
    \caption{\textbf{Evaluation of LLM-generated diagnostic labels.} The table assesses agreement with official Merlin labels and utility for training vision encoders.}
    \label{tab:S_label_eval}
    
    \vspace{0.5em}
    \textbf{Agreement with Merlin official labels} \\
    \suppTableFont\suppTableSetup
    \begin{tabular}{@{}lrrrrr@{}}
        \toprule
        Evaluation Protocol & $\kappa$ & Acc. & Prec. & Rec. & F1 \\
        \midrule
        Strict match (including \texttt{-1})      & 0.10 & 0.29 & 0.46 & 0.66 & 0.30 \\
        Exclude uncertain labels (\texttt{-1})   & 0.94 & 0.98 & 0.95 & 0.98 & 0.96 \\
        Map uncertain labels to negative (\texttt{-1}$\to$0)      & 0.43 & 0.91 & 0.33 & 0.97 & 0.46 \\
        \bottomrule
    \end{tabular}

    \vspace{1em}
    \textbf{Vision encoder performance (Merlin test set)} \\
    \begin{tabular}{@{}lrrrrrr@{}}
    \toprule
    Training Strategy & AUC & Acc. & Prec. & F1 & Sens. & Spec. \\
    \midrule
    Uncertain labels excluded & 85.78 & 86.30 & 61.38 & 86.39 & 69.65 & 81.22 \\
    Uncertain labels mapped to negative & 84.39 & 84.27 & 59.52 & 84.66 & 70.71 & 78.02 \\
    Doubao labels, uncertain labels excluded & 87.41 & 88.30 & 70.57 & 87.85 & 69.65 & 83.18 \\
    \bottomrule
    \end{tabular}
\end{table}

\begin{table}[htbp]
    \centering
    \caption{\textbf{Local-practice reread of CT-RATE hiatal-hernia positive labels.} We sampled CT-RATE cases with positive hiatal-hernia label for focused visual rereading. Among 554 sampled CT-RATE-positive cases, Chinese radiologists reread the images according to local reporting practice and regarded 188 cases as reportable hiatal hernia, corresponding to a 33.9\% reread-positive fraction.}
    \label{tab:hiatal_hernia_reread}
    \suppTableFont\suppTableSetup
    \begin{tabular}{@{}lr@{}}
        \toprule
        Measure & Value \\
        \midrule
        CT-RATE positive cases reread & 554 \\
        Regarded as reportable hiatal hernia on reread & 188 \\
        Not regarded as reportable hiatal hernia on reread & 366 \\
        Reread-positive fraction among CT-RATE-positive cases & 33.9\% \\
        \bottomrule
    \end{tabular}
\end{table}

\begin{figure}[htbp]
    \centering
    \includegraphics[width=\textwidth]{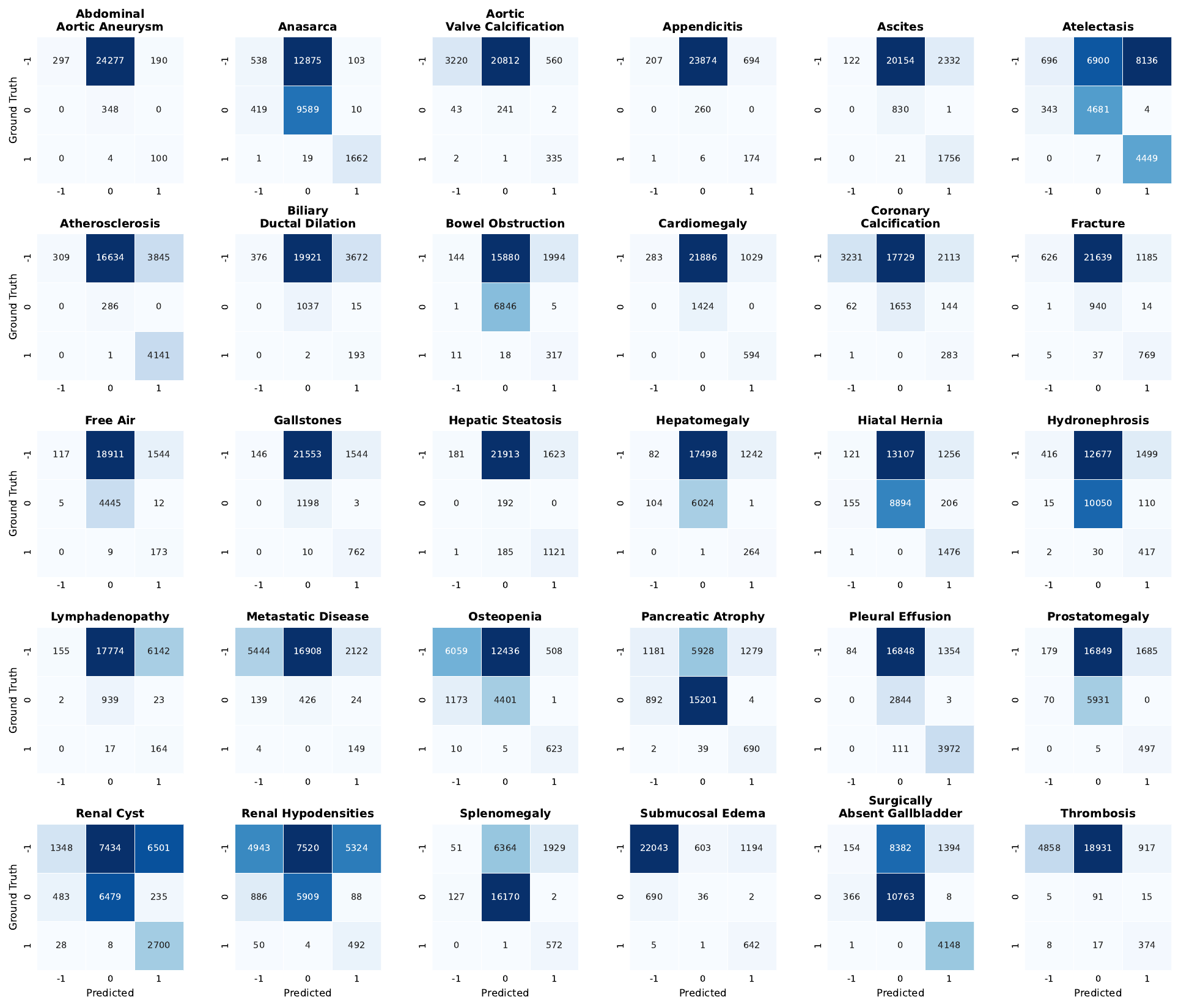}
    \caption{\textbf{Confusion matrices comparing LLM-extracted labels against Merlin official labels.} The figure displays the confusion matrices for 30 abnormality categories. The rows represent the official Merlin labels, and the columns represent the labels extracted by Doubao. The labels are defined as: 1 (present), 0 (absent), and -1 (uncertain/missing).}
    \label{fig:S1}
\end{figure}

\begin{table}[htbp]
    \centering
    
    \caption{\textbf{Board-certified expert audit of the LLM silver-label pipeline ($n=565$).} Three-class confusion matrix (rows: expert label; columns: LLM label). The largest off-diagonal mass is on the conservative axis (true-positive $\to\!-1$, $82$ samples); $3/565$ ($0.53\%$) are definite false positives.}
    \label{tab:audit_3x3}
    \suppTableFont\suppTableSetup
    \begin{tabular}{@{}lrrrr@{}}
        \toprule
         & LLM $=-1$ & LLM $=0$ & LLM $=1$ & Total \\
        \midrule
        Expert $=-1$ & 39 & 11 & 4 & 54 \\
        Expert $=0$  & 36 & 132 & 3 & 171 \\
        Expert $=1$  & 82 & 25 & 233 & 340 \\
        \midrule
        Total & 157 & 168 & 240 & 565 \\
        \bottomrule
    \end{tabular}

    \vspace{1em}
    \textbf{Aggregate agreement statistics} \\
    \begin{tabular}{@{}lrr@{}}
        \toprule
        Metric & Three-class & Binary (excl. uncertain, $n\!=\!393$) \\
        \midrule
        Raw agreement & 0.715 & 0.929 \\
        Cohen's $\kappa$ & 0.546 & 0.848 \\
        Sensitivity & --- & 0.903 \\
        Specificity & --- & 0.978 \\
        F1 & --- & 0.943 \\
        \bottomrule
    \end{tabular}
\end{table}

\begin{table}[htbp]
    \centering
    
    \caption{\textbf{Audit-informed prompt refinement.} The table reports full-audit metrics ($n\!=\!565$). Six discriminating rules derived from the linguistic risk map (hedge $\to\!1$, chronic/stable $\to\!1$, scoped negation, post-operative state, resolved/regressed, omission $\to\!-1$) raise three-class $\kappa$ from $0.546$ to $0.844$ and binary F1 from $0.943$ to $0.985$. Claude Opus with the original prompt reaches intermediate performance ($\kappa\!=\!0.654$, binary F1 $0.982$).}
    \label{tab:prompt_tuning}
    \suppTableFont\suppTableSetup
    \begin{tabular}{@{}lrrrr@{}}
        \toprule
        Configuration & 3-class agreement & 3-class $\kappa$ & Binary acc. & Binary F1 \\
        \midrule
        Doubao (primary pipeline)             & 0.715 & 0.546 & 0.929 & 0.943 \\
        Claude Opus 4.7 + original prompt      & 0.780 & 0.654 & --- & 0.982 \\
        Claude Opus 4.7 + audit-informed prompt & \textbf{0.917} & \textbf{0.844} & \textbf{0.980} & \textbf{0.985} \\
        \bottomrule
    \end{tabular}
\end{table}

\begin{table}[htbp]
    \centering
    \caption{\textbf{Uncertainty-aware training strategy comparison: Macro AUC, ECE, and Spec@Sens$_{0.95}$ with 95\% bootstrap CI (Stages~1 and~2).} Sample-level bootstrap ($n\!=\!1000$) over the Merlin validation set. Strategies are ordered by Stage-1 Macro AUC. ECE = Expected Calibration Error (lower is better); Spec@Sens$_{0.95}$ = specificity at 95\% sensitivity (higher is better).}
    \label{tab:uncertainty_strategy_ci}
    \suppWideTableFont\suppTableSetup
    \resizebox{\textwidth}{!}{%
    \begin{tabular}{@{}lrrrrrrr@{}}
        \toprule
        \multirow{2}{*}{Strategy}
          & \multicolumn{4}{c}{Stage~1}
          & \multicolumn{3}{c}{Stage~2} \\
        \cmidrule(lr){2-5}\cmidrule(lr){6-8}
         & Macro AUC & 95\% CI & ECE & Spec@Sens$_{0.95}$
         & Macro AUC & 95\% CI & ECE \\
        \midrule
        Confidence-weighted BCE           & \textbf{0.871} & \textbf{(0.861--0.881)} & 0.160 & \textbf{0.527} & \textbf{0.850} & \textbf{(0.840--0.861)} & \textbf{0.239} \\
        Ordinal cumulative-label regression              & 0.866 & (0.856--0.876) & 0.172 & 0.516 & 0.846 & (0.835--0.857) & 0.239 \\
        Masked BCE (baseline)  & 0.863 & (0.851--0.873) & \textbf{0.155} & 0.513 & 0.841 & (0.830--0.852) & 0.272 \\
        Three-class uncertainty modelling              & 0.862 & (0.851--0.872) & 0.166 & 0.522 & 0.835 & (0.824--0.846) & 0.280 \\
        Equal-probability soft labels           & 0.860 & (0.849--0.871) & 0.161 & 0.455 & 0.831 & (0.820--0.842) & 0.257 \\
        \bottomrule
    \end{tabular}}
\end{table}

\begin{table}[htbp]
    \centering
    \caption{\textbf{Noise robustness analysis --- Stage~1 and Stage~2 Macro AUC.} Stage~1 AUC is reported with 95\% bootstrap CI (sample-level bootstrap, $n\!=\!1000$). Four noise types are evaluated: symmetric label flips, uncertain-to-negative reassignment, uncertain-to-positive reassignment, and uncertain-to-random reassignment. Noise rate~0.0 (symmetric flips only) is the clean baseline.}
    \label{tab:noise_robust_stage1_ci}
    \label{tab:noise_robust_stage2}
    \suppTableFont\suppTableSetup
    \begin{tabular}{@{}lllrr@{}}
        \toprule
        Noise Type       & Rate & Stage~1 AUC & 95\% CI & Stage~2 Zero-shot AUC \\
        \midrule
        Symmetric label flip        & 0.0 & 0.856 & (0.846--0.866) & 0.835 \\
        Symmetric label flip        & 0.2 & 0.795 & (0.783--0.807) & 0.810 \\
        Symmetric label flip        & 0.4 & 0.529 & (0.512--0.543) & 0.775 \\
        Symmetric label flip        & 0.6 & 0.416 & (0.400--0.431) & 0.767 \\
        Symmetric label flip        & 0.8 & 0.201 & (0.189--0.213) & 0.789 \\
        Symmetric label flip        & 1.0 & 0.135 & (0.125--0.144) & 0.826 \\
        \addlinespace[3pt]
        Uncertain-to-negative   & 0.2 & 0.846 & (0.835--0.857) & 0.845 \\
        Uncertain-to-negative   & 0.4 & 0.838 & (0.827--0.850) & 0.832 \\
        Uncertain-to-negative   & 0.6 & 0.829 & (0.817--0.840) & 0.815 \\
        Uncertain-to-negative   & 0.8 & 0.828 & (0.817--0.839) & 0.828 \\
        Uncertain-to-negative   & 1.0 & 0.828 & (0.818--0.840) & 0.825 \\
        \addlinespace[3pt]
        Uncertain-to-positive   & 0.2 & 0.780 & (0.766--0.793) & 0.798 \\
        Uncertain-to-positive   & 0.4 & 0.765 & (0.752--0.778) & 0.807 \\
        Uncertain-to-positive   & 0.6 & 0.781 & (0.767--0.794) & 0.797 \\
        Uncertain-to-positive   & 0.8 & 0.770 & (0.757--0.783) & 0.804 \\
        Uncertain-to-positive   & 1.0 & 0.785 & (0.772--0.799) & 0.797 \\
        \addlinespace[3pt]
        Uncertain-to-random & 0.2 & 0.820 & (0.808--0.832) & 0.836 \\
        Uncertain-to-random & 0.4 & 0.795 & (0.783--0.808) & 0.832 \\
        Uncertain-to-random & 0.6 & 0.780 & (0.767--0.793) & 0.811 \\
        Uncertain-to-random & 0.8 & 0.784 & (0.771--0.796) & 0.818 \\
        Uncertain-to-random & 1.0 & 0.770 & (0.758--0.784) & 0.799 \\
        \bottomrule
    \end{tabular}
\end{table}

\FloatBarrier
\subsection{Imaging equipment information}
We summarize the imaging equipment and acquisition parameters for the datasets used in this study in Table \ref{tab:S4}.

\begin{table}[htbp]
    \centering
    \caption{\textbf{Summary of imaging equipment and acquisition parameters.} This table provides an overview of the scanner manufacturers and models used across the different datasets. For AH-Chest, AH-Abd, AH-Knee, and AH-Spine, detailed statistics are derived from DICOM metadata.}
    \label{tab:S4}
    \resizebox{\textwidth}{!}{
    \begin{tabular}{llccc}
        \toprule
        Dataset & Manufacturer & Model & Slice Thickness (mm) & Tube Voltage (kVp) \\
        \midrule
        CT-RATE & \begin{tabular}{@{}l@{}}Philips (62\%)\\Siemens (30\%)\\GE (8\%)\end{tabular} & - & 0.035 -- 6.0 & - \\
        Merlin & - & - & - & - \\
        RAD-ChestCT & \begin{tabular}{@{}l@{}}GE (57\%)\\Siemens (43\%)\end{tabular} & - & 0.6 -- 0.625 & - \\ 
        AH-Chest & \begin{tabular}{@{}l@{}}GE (80\%)\\Philips (20\%)\end{tabular} & \begin{tabular}{@{}l@{}}Optima (49\%)\\Brilliance (19\%)\\LightSpeed (18\%)\\Discovery (11\%)\end{tabular} & 0.625 -- 5.0 & 80 -- 140 \\
        AH-Abd & GE (100\%) & \begin{tabular}{@{}l@{}}Optima (45\%)\\Discovery (40\%)\\Revolution (15\%)\end{tabular} & 0.625 -- 5.0 & 100 -- 140 \\
        AH-Knee (institutional) & \begin{tabular}{@{}l@{}}Siemens (88\%)\\GE (12\%)\end{tabular} & \begin{tabular}{@{}l@{}}Avanto (78\%)\\Discovery MR750w (12\%)\\MAGNETOM Vida (10\%)\end{tabular} & -  & - \\
        AH-Spine (institutional) & \begin{tabular}{@{}l@{}}Philips (50\%)\\GE (45\%)\\Siemens (5\%) \end{tabular} & \begin{tabular}{@{}l@{}}Achieva (50\%)\\Signa Voyager (44\%)\\MAGNETOM Vida (5\%)\\Discovery MR750w (1\%)\end{tabular} & -  & - \\
        \bottomrule
    \end{tabular}
    }
\end{table}

\FloatBarrier
\subsection{Detailed results}

\begin{table}[htbp]
    \centering
    \caption{\textbf{Detailed classification performance on the AH-Chest dataset.} AUC (\%) is reported for each evaluated abnormality category.}
    \label{tab:S5}
    \suppTableFont\suppTableSetup
    \begin{tabular}{@{}lrr@{}}
    \toprule
    Abnormality & BrgSA & GreenRFM \\
    \midrule
    Atelectasis & 53.4 & 73.8 \\
    Bronchiectasis & 68.8 & 70.8 \\
    Consolidation & 72.1 & 83.3 \\
    Coronary Artery Calcification & 80.3 & 80.4 \\
    Emphysema & 52.8 & 71.3 \\
    Hiatal Hernia & 76.6 & 85.7 \\
    Interlobular Septal Thickening & 52.5 & 74.0 \\
    Lung Nodule & 54.6 & 59.7 \\
    Lung Opacity & 50.6 & 75.1 \\
    Lymphadenopathy & 52.0 & 69.8 \\
    Peribronchial Thickening & 61.1 & 81.3 \\
    Pericardial Effusion & 57.9 & 76.9 \\
    Pleural Effusion & 69.8 & 77.5 \\
    Pulmonary Fibrotic Sequelae & 54.5 & 63.9 \\
    \midrule
    Mean & 61.2 & 74.5 \\
    \bottomrule
    \end{tabular}
\end{table}

\begin{table}[htbp]
    \centering
    \caption{\textbf{Detailed classification performance on the full AH-Abd cross-site benchmark.} AUC (\%) is reported for each evaluated abnormality category.}
    \label{tab:S6}
    \suppTableFont\suppTableSetup
    \begin{tabular}{@{}lrr@{}}
    \toprule
    Abnormality & Merlin & GreenRFM \\
    \midrule
    Thrombosis & 65.0 & 71.5 \\
    Surgically absent gallbladder & 56.1 & 87.3 \\
    Submucosal edema & 70.2 & 73.3 \\
    Splenomegaly & 60.8 & 77.1 \\
    Renal hypodensities & 50.9 & 58.1 \\
    Renal cyst & 51.8 &59.1 \\
    Prostatomegaly & 47.1 &77.7 \\
    Pleural effusion & 79.9 &90.0 \\
    Pancreatic atrophy & 58.5 &77.6 \\
    Osteopenia & 52.5 &67.6 \\
    Metastatic disease & 57.5 &65.1 \\
    Lymphadenopathy & 53.4 &55.3 \\
    Hydronephrosis & 60.9 &85.5 \\
    Hiatal hernia & 59.0 &65.7 \\
    Hepatomegaly & 64.1 &76.4 \\
    Hepatic steatosis &56.0 &86.7 \\
    Gallstones &53.3 &67.5 \\
    Free air &61.0 &55.3 \\
    Fracture &58.1 &69.5 \\
    Coronary calcification &49.0 &77.6 \\
    Cardiomegaly &70.8 &86.4 \\
    Bowel obstruction &58.3 &82.5 \\
    Biliary duct dilation &56.4 &75.9 \\
    Atherosclerosis &58.9 &64.3 \\
    Atelectasis &60.1 &60.9 \\
    Ascites &72.9 &73.4 \\
    Appendicitis &61.2 &66.7 \\
    Aortic valve calcification & 44.5 &73.3 \\
    Anasarca &90.1 &97.1 \\
    Abdominal aortic aneurysm &71.5 &70.5 \\
    \midrule
    Mean & 60.3 & 73.2 \\
    \bottomrule
    \end{tabular}
\end{table}

\begin{table}[htbp]
    \centering
    \caption{\textbf{Detailed classification performance on the AH-Knee dataset.} AUC (\%) is reported for each evaluated abnormality category.}
    \label{tab:S7}
    \suppTableFont\suppTableSetup
    \begin{tabular}{@{}lrr@{}}
    \toprule
    Abnormality & MRCLIP & GreenRFM \\
    \midrule
    Medial meniscus injury & 81.6 & 85.6 \\
    Lateral meniscus injury & 72.8 & 71.6 \\
    Anterior cruciate ligament injury & 65.8 & 84.3 \\
    Posterior cruciate ligament injury & 55.5 & 72.0 \\
    Medial collateral ligament injury & 82.1 & 84.7 \\
    Lateral collateral ligament injury & 61.8 & 75.2 \\
    Articular cartilage defect, Osteochondral injury & 80.9 & 86.9 \\
    Bone marrow edema, Contusion & 77.7 & 87.3 \\
    Fracture Collapse & 79.9 & 88.4 \\
    Joint effusion or Bakers cyst & 69.2 & 52.3 \\
    Synovitis, Bursitis & 56.1 & 60.7 \\
    Postoperative changes, Internal fixation & 75.1 & 85.5 \\
    Degenerative osteoarthritis, Osteophytes & 89.0 & 93.8 \\
    \midrule
    Mean & 72.9 & 79.0 \\
    \bottomrule
    \end{tabular}
\end{table}

\begin{table}[htbp]
    \centering
    \caption{\textbf{Detailed classification performance on the AH-Spine dataset.} AUC (\%) is reported for each evaluated abnormality category.}
    \label{tab:S8}
    \suppTableFont\suppTableSetup
    \begin{tabular}{@{}lrr@{}}
    \toprule
    Abnormality & MRCLIP & GreenRFM \\
    \midrule
    Intervertebral disc degeneration & 74.8 & 86.5 \\
    Disc bulging & 56.6 & 70.2 \\
    Disc herniation, Extrusion & 61.1 & 79.2 \\
    Spondylolisthesis & 73.7 & 78.3 \\
    Spinal canal or foraminal stenosis & 57.7 & 73.2 \\
    Endplate changes, Schmorls nodes, Modic changes & 72.0 & 74.1 \\
    Compression fracture, Acute bone contusion & 91.4 & 95.8 \\
    Postoperative changes, Internal fixation, Bone graft fusion & 76.3 & 88.8 \\
    \midrule
    Mean & 70.5 & 80.8 \\
    \bottomrule
    \end{tabular}
\end{table}

\FloatBarrier
\subsection{LLM prompts for visual description extraction}
\label{supp:visual_description_prompt}
To extract visual descriptions from radiology reports for the visual-supervision analysis, we used the following system prompt. Unlike the diagnostic label extraction prompt, which explicitly models uncertainty (class -1), this prompt used a deterministic binary schema (0/1) for visual patterns. The LLM was instructed to output 1 only when a pattern was \textit{explicitly and clearly} described, and 0 otherwise. This decision relies on the distinction between observation and interpretation: while diagnostic conclusions often carry uncertainty, visual findings (e.g., ``high attenuation'') are typically factual observations. This strict binary output was used to obtain a low-noise visual supervisory signal and reduce ambiguity from underspecified descriptions.

\begin{tcolorbox}[colback=gray!10,colframe=gray!75,title=System Prompt: Visual Description Extraction,fontupper=\small\ttfamily,breakable]
\begin{verbatim}
You are an experienced chest radiologist. Review each non-contrast 
chest CT report and determine whether the following visual patterns 
are explicitly described. If a pattern is clearly present, output 1; 
if it is not mentioned or the description is inconclusive, output 0. 
Multiple patterns can be present simultaneously.

Visual patterns to evaluate (binary flags in this exact order):
- Density patterns -
1. High-attenuation focus (density_high)
2. Low-attenuation focus (density_low)
3. Mixed-attenuation focus (density_mixed)

- Morphology patterns -
4. Nodular opacity (morphology_nodular)
5. Patchy opacity (morphology_patchy)
6. Linear/stripe-like opacity (morphology_linear)
7. Reticular/network-like opacity (morphology_reticular)

- Distribution patterns -
8. Focal distribution (distribution_focal)
9. Diffuse distribution (distribution_diffuse)
10. Bilateral symmetric distribution 
    (distribution_bilateral_symmetric)
11. Unilateral distribution (distribution_unilateral)

Output format: exactly eleven comma-separated binary digits 
(0 or 1) matching the order above, with no additional text 
or punctuation. 
Example: 0,1,0,0,0,1,0,1,0,0,0
\end{verbatim}
\end{tcolorbox}

\FloatBarrier
\subsection{LLM prompts and label-verification workflow for diagnostic label extraction}
\label{supp:diagnostic_label_prompts}
To support reproducibility, we provide the exact system prompts and label-verification workflow used to extract structured diagnostic labels from radiology reports.

\subsubsection{Prompt for 30-class diagnostic label extraction (Merlin/AH-Abd)}
\begin{tcolorbox}[colback=gray!10,colframe=gray!75,title=System Prompt: 30-Class Diagnostic Label Extraction,fontupper=\small\ttfamily,breakable]
\begin{verbatim}
You are a board-certified radiologist specializing in thoracic imaging.

Task
Classify the chest/abdomen CT report for the following 30 findings,
in the exact order listed:
1. submucosal_edema
2. renal_hypodensities
3. aortic_valve_calcification
4. coronary_calcification
5. thrombosis
6. metastatic_disease
7. pancreatic_atrophy
8. renal_cyst
9. osteopenia
10. surgically_absent_gallbladder
11. atelectasis
12. abdominal_aortic_aneurysm
13. anasarca
14. hiatal_hernia
15. lymphadenopathy
16. prostatomegaly
17. biliary_ductal_dilation
18. cardiomegaly
19. splenomegaly
20. hepatomegaly
21. atherosclerosis
22. ascites
23. pleural_effusion
24. hepatic_steatosis
25. appendicitis
26. gallstones
27. hydronephrosis
28. bowel_obstruction
29. free_air
30. fracture

Labeling rules
- Output a single line of 30 comma-separated values.
- Use 1 if the finding is explicitly present.
- Use 0 if the finding is explicitly ruled out or absent.
- Use -1 if the report is ambiguous, uncertain, or lacks information.

Only return the comma-separated numbers.
Example:
0,1,-1,-1,1,-1,-1,-1,-1,-1,0,0,0,0,1,0,0,0,-1,0,0,1,0,0,0,0,0,0,0,0
\end{verbatim}
\end{tcolorbox}

\subsubsection{Prompt for 18-class diagnostic label extraction (CT-RATE/AH-Chest)}
\begin{tcolorbox}[colback=gray!10,colframe=gray!75,title=System Prompt: 18-Class Diagnostic Label Extraction,fontupper=\small\ttfamily,breakable]
\begin{verbatim}
You are a board-certified radiologist specializing in thoracic imaging.

Task
Classify the chest/abdomen CT report for the following 18 findings,
in the exact order listed:
1. medical material
2. arterial wall calcification
3. cardiomegaly
4. pericardial effusion
5. coronary artery wall calcification
6. hiatal hernia
7. lymphadenopathy
8. emphysema
9. atelectasis
10. lung nodule
11. lung opacity
12. pulmonary fibrotic sequela
13. pleural effusion
14. mosaic attenuation pattern
15. peribronchial thickening
16. consolidation
17. bronchiectasis
18. interlobular septal thickening

Labeling rules
- Output a single line of 18 comma-separated values.
- Use 1 if the finding is explicitly present.
- Use 0 if the finding is explicitly ruled out or absent.
- Use -1 if the report is ambiguous, uncertain, or lacks information.

Only return the comma-separated numbers.
Example:
0,1,-1,-1,1,-1,-1,-1,-1,-1,0,0,0,0,1,0,0,0
\end{verbatim}
\end{tcolorbox}

\subsubsection{Label extraction and verification workflow}

\begin{figure}[!b]
    \centering
    \begin{tikzpicture}[node distance=2cm, scale=0.8, transform shape]
        \tikzstyle{startstop} = [rectangle, rounded corners, minimum width=3cm, minimum height=1cm,text centered, draw=black, fill=red!10]
        \tikzstyle{io} = [trapezium, trapezium left angle=70, trapezium right angle=110, minimum width=3cm, minimum height=1cm, text centered, draw=black, fill=blue!10]
        \tikzstyle{process} = [rectangle, minimum width=3cm, minimum height=1cm, text centered, draw=black, fill=orange!10]
        \tikzstyle{decision} = [diamond, minimum width=3cm, minimum height=1cm, text centered, draw=black, fill=green!10]
        \tikzstyle{arrow} = [thick,->,>=stealth]

        \node (start) [startstop] {Input: Radiology Report};
        \node (pro1) [process, below of=start] {LLM Extraction};
        \node (dec1) [decision, below of=pro1, yshift=-1.5cm] {Format Valid?};
        \node (pro2a) [io, below of=dec1, yshift=-1.0cm] {Output: Labels};
        \node (dec2) [decision, right of=dec1, xshift=3.5cm] {Retry $<$ Max?};
        \node (pro2b) [process, above of=dec2, yshift=1.5cm] {Exponential Backoff};
        \node (stop) [startstop, right of=dec2, xshift=2.5cm] {Log for Manual Review};

        \draw [arrow] (start) -- (pro1);
        \draw [arrow] (pro1) -- (dec1);
        \draw [arrow] (dec1) -- node[anchor=east] {Yes} (pro2a);
        \draw [arrow] (dec1) -- node[anchor=south] {No} (dec2);
        \draw [arrow] (dec2) -- node[anchor=west] {Yes} (pro2b);
        \draw [arrow] (pro2b) -- (pro1);
        \draw [arrow] (dec2) -- node[anchor=south] {No} (stop);
    \end{tikzpicture}
    \caption{\textbf{Schematic of the LLM-based label extraction and verification pipeline.} The workflow accepts only syntactically valid label sequences. Invalid outputs trigger a retry mechanism to maximize data utilization while maintaining structural integrity.}
    \label{fig:llm_workflow}
\end{figure}

To improve the reliability of the extracted labels, we implement a verification loop (Fig. \ref{fig:llm_workflow}). The process involves:
\begin{enumerate}[label=\arabic*.]
    \item \textbf{Extraction}: The LLM is prompted to generate a structured string of comma-separated integers.
    \item \textbf{Format verification}: The output is parsed to verify that it contains exactly $N$ tokens (18 for CT-RATE, 30 for Merlin) and that every token is a valid class identifier $\{-1, 0, 1\}$.
    \item \textbf{Error handling}: If the output fails verification, the request is retried with exponential backoff up to 3 times. This filters out hallucinated formats or incomplete generations. Samples that fail all retry attempts are logged for manual review to protect data integrity.
\end{enumerate}

\FloatBarrier
\subsection{Grad-CAM visualizations}

We use Grad-CAM to visualize the image regions activated by the corresponding text prompts in the model. For visualization, each CAM was clipped using a fixed display threshold to suppress background activations, min-max normalized to $[0,1]$ before color mapping, and then overlaid on the original image using alpha blending; the transparency coefficient ($\alpha$) was adjusted to preserve anatomical context while keeping high-activation regions visible. 

\begin{figure}[!h]
    \centering
    \includegraphics[width=\textwidth]{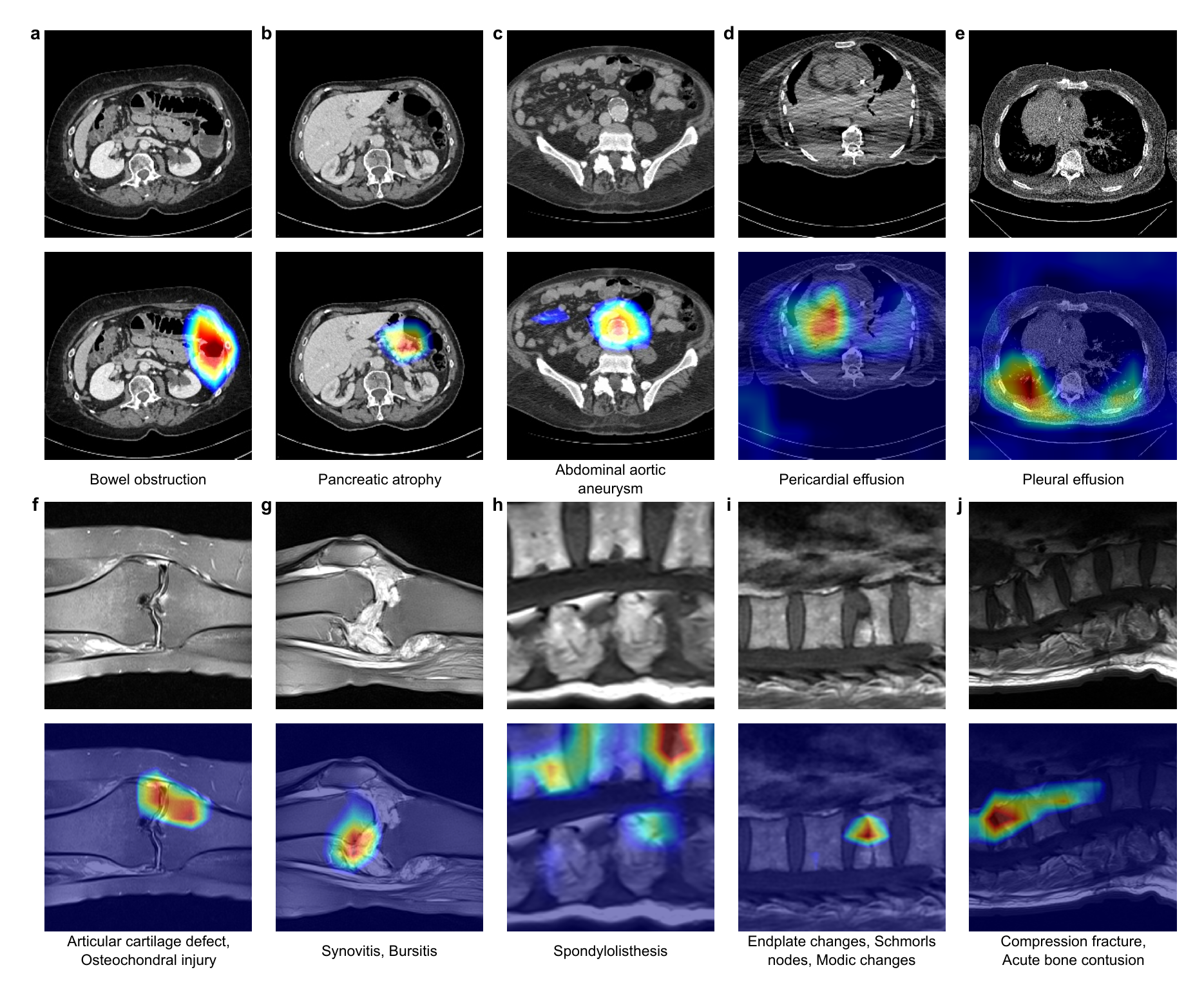}
    \caption{\textbf{Visualization of correct pathology localization using Grad-CAM heatmaps.} The figure displays representative cases where the model successfully identified the region of interest. The top rows show the original cross-sectional imaging (CT or MRI), and the bottom rows show the corresponding class activation maps superimposed on the original images, with red regions indicating high diagnostic weight. 
    (a) Bowel obstruction. 
    (b) Pancreatic atrophy. 
    (c) Abdominal aortic aneurysm. 
    (d) Pericardial effusion. 
    (e) Pleural effusion. 
    (f) Articular cartilage defect, osteochondral injury. 
    (g) Synovitis, bursitis. 
    (h) Spondylolisthesis. 
    (i) Endplate changes, Schmorl's nodes, and Modic changes. 
    (j) Compression fracture, acute bone contusion.} 
    \label{fig:S3}
\end{figure}

\begin{figure}[!h]
    \centering
    \includegraphics[width=\textwidth]{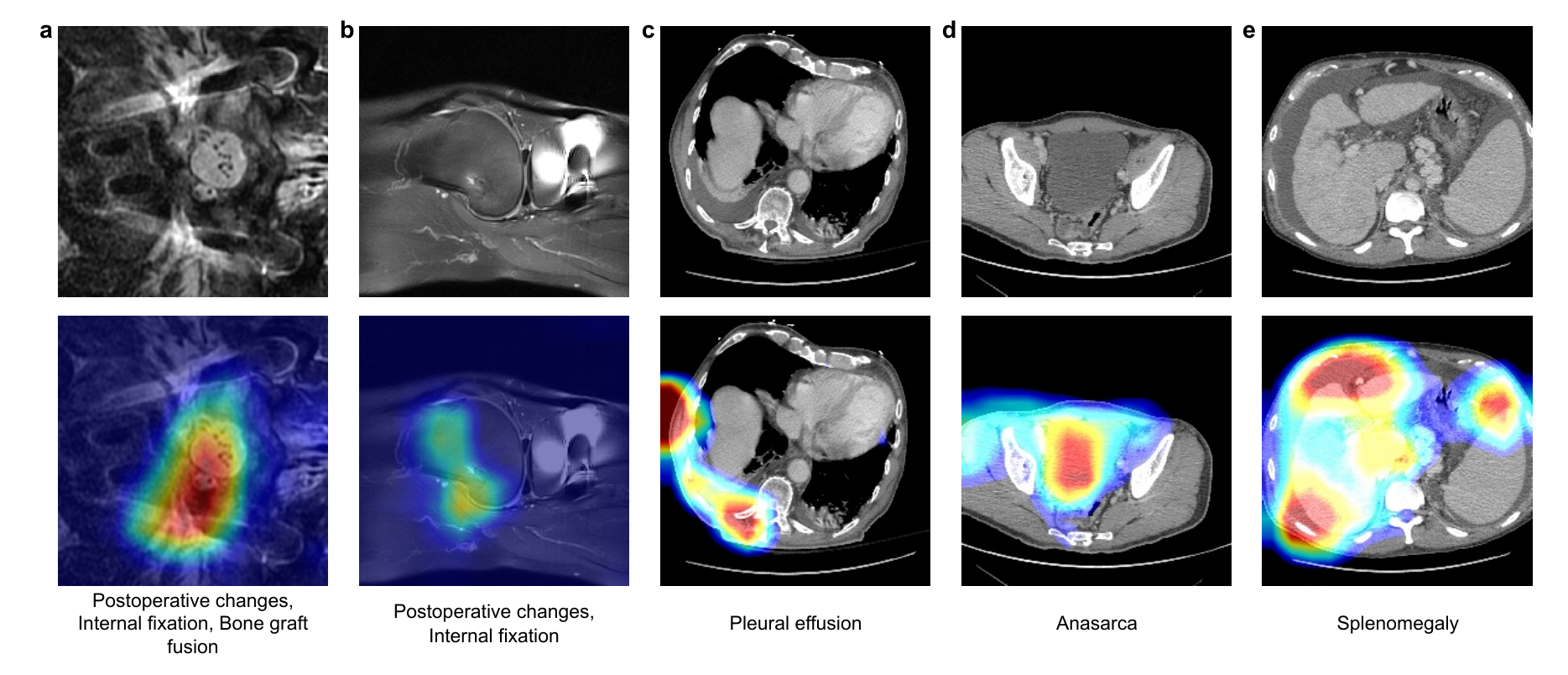}
    \caption{\textbf{Failure modes and misinterpretation of Grad-CAM heatmaps.} The figure highlights cases where the model failed to localize the primary pathology or focused on anatomically irrelevant regions. 
    (a-b) Postoperative changes (Knee \& Spine): The model consistently failed to activate regions with internal fixation or bone grafts. 
    (c) Ascites/Pleural Effusion confusion: The model displayed ambiguity in distinguishing fluid compartments, likely due to the similar radiodensity of pleural and peritoneal fluids. 
    (d) Anasarca: In cases of anasarca, the model often disproportionately directs high attention to the bladder, misinterpreting high-contrast fluid accumulation as the target feature. 
    (e) Splenomegaly: In cases of splenomegaly, the model often simultaneously focuses on the liver; this might suggest the model relies on comparative organ sizing or hepatosplenomegaly co-occurrence patterns. }
    \label{fig:S4}
\end{figure}

\FloatBarrier
\subsection{Inference-time computational cost}
\label{supp:inf_cost}

This appendix supplements the inference cost measured on a single NVIDIA RTX~3090 (24~GB VRAM).
Latency was measured after 10 warm-up passes using 50 timed forward passes at batch size 1. 
CPU pre-processing costs (HU windowing, trilinear resampling, and GPU transfer) are benchmarked separately after 15 warm-up passes using 30 timed passes, and combined with inference latency to form the end-to-end estimate reported in Table~\ref{tab:e2e_cost}.
We report directly measured latency, throughput, pre-processing time, and peak VRAM rather than extrapolating site-independent carbon estimates from short inference runs, because carbon accounting depends on local power mix, utilization, and deployment scheduling.

\begin{table}[htbp]
\centering
\caption{\textbf{Inference-cost benchmark across three spatial resolutions.}
Latency (mean\,$\pm$\,s.d., 50 timed passes), peak VRAM, and throughput for image encoders benchmarked on a single NVIDIA RTX~3090 (24~GB VRAM), batch size~1, \texttt{torch.inference\_mode()}.
$\dagger$~CT-CLIP requires $F\bmod 10=0$, $H/W\bmod 20=0$; inputs not satisfying this constraint are zero-padded to the nearest valid size ($192^2 \times 96 \to 200^2 \times 100$; $224^2 \times 112 \to 240^2 \times 120$).}
\label{tab:inf_cost}
\suppWideTableFont\suppTableSetup
\resizebox{\textwidth}{!}{
\begin{tabular}{@{}lrrrrrrrrr@{}}
\toprule
& \multicolumn{3}{c}{\textbf{Latency (ms)}} & \multicolumn{3}{c}{\textbf{Peak VRAM (GB)}} & \multicolumn{3}{c}{\textbf{Throughput (vol/s)}} \\
\cmidrule(lr){2-4}\cmidrule(lr){5-7}\cmidrule(lr){8-10}
Model & $192^2 \times 96$ & $480^2 \times 240$ & $224^2{\times}112$ & $192^2 \times 96$ & $480^2 \times 240$ & $224^2{\times}112$ & $192^2 \times 96$ & $480^2 \times 240$ & $224^2{\times}112$ \\
\midrule
GreenRFM         & $29.0{\pm}0.3$ & $427.6{\pm}5.0$ & $46.8{\pm}0.6$ & 0.67 & 8.58 & 0.99 & 34.4 & 2.3 & 21.4 \\
GreenRFM-L  & $\mathbf{5.3{\pm}0.1}$ & $\mathbf{60.9{\pm}0.6}$ & $\mathbf{7.9{\pm}0.1}$ & \textbf{0.23} & \textbf{1.36} & \textbf{0.25} & \textbf{187.2} & \textbf{16.4} & \textbf{127.2} \\
CT-CLIP$^\dagger$    & $8.3{\pm}1.8$ & $109.9{\pm}0.3$ & $11.1{\pm}0.2$ & 0.28 & 2.25 & 0.37 & 119.9 & 9.1 & 89.8 \\
BrgSA               & $39.5{\pm}0.3$ & $4086{\pm}170$ & $83.3{\pm}1.3$ & 0.46 & 2.17 & 0.53 & 25.3 & 0.2 & 12.0 \\
\bottomrule
\end{tabular}
}
\end{table}

\paragraph{Pre-processing footprint}
The shared CPU pre-processing pipeline (applied identically to all models) consists of three steps: HU windowing of the raw $300{\times}512{\times}512$ float32 volume to $[-1000,+1000]$ or $[-1000,+200]$ followed by normalization to $[-1,1]$; trilinear resampling to the target resolution via NumPy axis-wise \texttt{linspace} indexing; and GPU transfer via \texttt{torch.from\_numpy().to(device)}. 
Measured pre-processing latencies (mean~$\pm$~s.d., 30 timed passes after 15 warm-up passes) are: $261\pm7$~ms at $96{\times}192{\times}192$; $1{,}176\pm6$~ms at $240{\times}480{\times}480$; and $295\pm2$~ms at $112{\times}224{\times}224$.
HU windowing is approximately constant ($\approx140$~ms) because it always operates on the original $300{\times}512{\times}512$ volume; resampling dominates and scales with output voxel count.

\begin{table}[htbp]
\centering
\caption{\textbf{End-to-end latency and VRAM at each model's native input resolution.}
Inference latency and peak VRAM are taken from Table~\ref{tab:inf_cost}. Values are rounded to the nearest millisecond.}
\label{tab:e2e_cost}
\suppTableFont\suppTableSetup
\begin{tabular}{@{}lrrrrr@{}}
\toprule
Model & Native resolution & Pre-proc. (ms) & Inference (ms) & VRAM (GB) & \textbf{E2E (ms)} \\
\midrule
GreenRFM-L (lightweight) & $96{\times}192{\times}192$ & $261\pm7$ & $5\pm0$ & \textbf{0.23} & $\mathbf{\sim266}$ \\
GreenRFM         & $96{\times}192{\times}192$ & $261\pm7$ & $29\pm0$ & 0.67 & $\mathbf{\sim290}$ \\
BrgSA              & $112{\times}224{\times}224$ & $295\pm2$  & $83\pm1$ & 0.53 & $\mathbf{\sim378}$ \\
CT-CLIP            & $240{\times}480{\times}480$ & $1{,}176\pm6$ & $110\pm0$ & 2.25 & $\mathbf{\sim1{,}286}$ \\
\bottomrule
\end{tabular}
\end{table}

\paragraph{Native-resolution summary}
At each model's native input resolution, GreenRFM-L achieves the lowest inference latency (5.3~ms) and VRAM (0.23~GB). GreenRFM at native resolution requires 29.0~ms and 0.67~GB.
CT-CLIP at its native $240{\times}480{\times}480$ input requires 109.9~ms and 2.25~GB; BrgSA at $112{\times}224{\times}224$ requires 83.3~ms and 0.53~GB.
At the largest tested resolution ($240{\times}480{\times}480$), BrgSA's full 3D self-attention over $\approx\!27{,}000$ tokens causes latency to increase to over 4~seconds, whereas GreenRFM-L remains at 60.9~ms.
Including the pre-processing pipeline, end-to-end per-volume latency at native resolution is approximately 266~ms (GreenRFM-L), 290~ms (GreenRFM), 378~ms (BrgSA), and 1,286~ms (CT-CLIP).

\FloatBarrier
\subsection{Robustness, representation diagnostics, retrieval, and generative transfer}
\label{supp:repr_diag}

\subsubsection{Within-label retrieval: attribute consistency and ranking drivers}
\label{supp:within_label}

\textbf{Setup.}  For each of the 18 CT-RATE diagnostic labels, positive-class cases ($n_{\text{pos}}$ per label) were retrieved by cosine similarity on image embeddings from GreenRFM and a CLIP-style baseline. Top-$K$ neighbours were scored against three visual-attribute groups extracted by LLM from the same reports (11 attributes total): density (high / low / mixed), morphology (nodular / patchy / linear / reticular), distribution (focal / diffuse / bilateral\_symmetric / unilateral).  Match rate is the fraction of Top-$K$ neighbours sharing the query's majority attribute; random baseline equals the per-attribute prevalence in the positive-class subset.

\begin{table}[htbp]
\centering
\caption{\textbf{Within-label retrieval attribute consistency (averaged over 18 diagnoses).}  GreenRFM and CLIP-style baseline Top-$K$ attribute match rates.}
\label{tab:within_label_retrieval}
\suppTableFont\suppTableSetup
\begin{tabular}{@{}llrrrr@{}}
\toprule
Attribute group & Method & $K=5$ & $K=10$ & $K=20$ & $K=50$ \\
\midrule
\multirow{2}{*}{Density}
  & GreenRFM   & \textbf{0.838} & \textbf{0.817} & \textbf{0.805} & \textbf{0.797} \\
  & CLIP-Baseline & 0.828 & 0.812 & 0.801 & 0.796 \\
\addlinespace[2pt]
\multirow{2}{*}{Morphology}
  & GreenRFM   & \textbf{0.781} & \textbf{0.751} & \textbf{0.733} & \textbf{0.717} \\
  & CLIP-Baseline & 0.759 & 0.733 & 0.718 & 0.707 \\
\addlinespace[2pt]
\multirow{2}{*}{Distribution}
  & GreenRFM   & \textbf{0.724} & \textbf{0.690} & \textbf{0.672} & \textbf{0.657} \\
  & CLIP-Baseline & 0.707 & 0.680 & 0.663 & 0.651 \\
\bottomrule
\end{tabular}
\end{table}

\FloatBarrier
\subsubsection{Ranking-driver analysis.}  For each of the 768 embedding dimensions, the diagnostic-dominance score $r_{\mathrm{diag}}(d)$ was defined as the maximum absolute point-biserial correlation between dimension $d$ and any of the 18 diagnostic labels; the attribute-dominance score $r_{\mathrm{attr}}(d)$ was defined analogously over the 11 visual attributes.  Dimensions with $r_{\mathrm{attr}}(d)>r_{\mathrm{diag}}(d)$ were labelled \emph{attribute-dominant}; the remainder were labelled \emph{diagnostic-dominant}.  GreenRFM assigns $88/768$ ($11.5\%$) dimensions to attribute-dominant roles versus $72/768$ ($9.4\%$) for the baseline.  Spearman rank correlation between each dimension subset and within-label case ordering is reported in Tab.~\ref{tab:ranking_driver}.

\begin{table}[htbp]
\centering
\caption{\textbf{Ranking-driver analysis: Spearman $\rho$ between dimension subsets and within-label ordering.}  Bold marks entries where GreenRFM exceeds baseline by $\geq 0.05$.}
\label{tab:ranking_driver}
\suppWideTableFont\suppTableSetup
\resizebox{\textwidth}{!}{
\begin{tabular}{@{}lrrrrr@{}}
\toprule
Pathology & $n_{\text{pos}}$ & GreenRFM diag-$\rho$ & GreenRFM attr-$\rho$ & Baseline diag-$\rho$ & Baseline attr-$\rho$ \\
\midrule
Medical material                  & 313  & 0.997 & 0.817 & 0.998 & 0.810 \\
Arterial wall calcification       & 867  & 0.996 & 0.833 & 0.999 & 0.819 \\
Cardiomegaly                      & 325  & 0.997 & 0.869 & 0.998 & 0.850 \\
Pericardial effusion              & 226  & 0.998 & 0.823 & 0.999 & 0.857 \\
Coronary artery wall calcification& 765  & 0.996 & 0.840 & 0.999 & 0.823 \\
Hiatal hernia                     & 417  & 0.995 & 0.820 & 0.997 & 0.736 \\
Lymphadenopathy                   & 789  & 0.995 & \textbf{0.773} & 0.997 & 0.676 \\
Emphysema                         & 600  & 0.997 & \textbf{0.789} & 0.998 & 0.724 \\
Atelectasis                       & 713  & 0.997 & \textbf{0.780} & 0.998 & 0.677 \\
Lung nodule                       & 1361 & 0.997 & \textbf{0.808} & 0.998 & 0.688 \\
Lung opacity                      & 1184 & 0.995 & \textbf{0.793} & 0.997 & 0.738 \\
Pulmonary fibrotic sequela        & 831  & 0.996 & \textbf{0.793} & 0.998 & 0.718 \\
Pleural effusion                  & 376  & 0.998 & 0.869 & 0.998 & 0.869 \\
Mosaic attenuation pattern        & 253  & 0.998 & \textbf{0.846} & 0.998 & 0.762 \\
Peribronchial thickening          & 355  & 0.996 & \textbf{0.807} & 0.998 & 0.768 \\
Consolidation                     & 581  & 0.997 & 0.836 & 0.997 & 0.831 \\
Bronchiectasis                    & 330  & 0.997 & \textbf{0.807} & 0.998 & 0.761 \\
Interlobular septal thickening    & 249  & 0.997 & 0.846 & 0.996 & 0.853 \\
\midrule
\textbf{Mean}                     & ---  & \textbf{0.997} & \textbf{0.822} & \textbf{0.998} & \textbf{0.769} \\
\bottomrule
\end{tabular}
}
\end{table}

\FloatBarrier
\subsubsection{Beyond-label concept retrieval}
\label{supp:beyond_label}

\textbf{Setup.}  24 clinical concepts absent from the 18 training-label set were compiled across five categories (anatomical location: 5; temporal change: 4; post-operative findings: 5; severity: 4; texture/appearance: 6). 
Concept text was encoded and ranked against all $n\!=\!3{,}039$ image embeddings; Precision@$K$ measures the fraction of true positives in the Top-$K$.

\begin{table}[htbp]
\centering
\caption{\textbf{Beyond-label concept retrieval: Precision@$K$ by category (mean across concepts in each category).}  Bold marks the better method at each $K$. }
\label{tab:beyond_label_retrieval}
\suppTableFont\suppTableSetup
\begin{tabular}{@{}llrrrrr@{}}
\toprule
Category & Method & P@5 & P@10 & P@20 & P@50 & P@100  \\
\midrule
\multirow{2}{*}{Anatomical location}
  & GreenRFM      & \textbf{0.080} & 0.040 & \textbf{0.060} & \textbf{0.044} & \textbf{0.038} \\
  & CLIP-Baseline & 0.040 & 0.040 & 0.020 & 0.020 & 0.026 \\
\addlinespace[2pt]
\multirow{2}{*}{Temporal change}
  & GreenRFM      & \textbf{0.100} & \textbf{0.175} & \textbf{0.138} & \textbf{0.145} & \textbf{0.128} \\
  & CLIP-Baseline & 0.050 & 0.075 & 0.088 & 0.065 & 0.068 \\
\addlinespace[2pt]
\multirow{2}{*}{Post-operative findings}
  & GreenRFM      & 0.000 & 0.050 & 0.038 & 0.015 & 0.023 \\
  & CLIP-Baseline & \textbf{0.200} & \textbf{0.200} & \textbf{0.163} & \textbf{0.125} & \textbf{0.120} \\
\addlinespace[2pt]
\multirow{2}{*}{Severity}
  & GreenRFM      & 0.000 & 0.000 & 0.000 & 0.005 & 0.010 \\
  & CLIP-Baseline & \textbf{0.250} & \textbf{0.250} & \textbf{0.200} & \textbf{0.150} & \textbf{0.118} \\
\addlinespace[2pt]
\multirow{2}{*}{Texture/appearance}
  & GreenRFM      & 0.160 & 0.120 & 0.090 & 0.076 & 0.072 \\
  & CLIP-Baseline & \textbf{0.280} & \textbf{0.220} & \textbf{0.220} & \textbf{0.220} & \textbf{0.224} \\
\bottomrule
\end{tabular}
\end{table}

\FloatBarrier
\subsubsection{Attribute probing: linear probe, CKA, and mutual information}
\label{supp:attr_probe}

\textbf{Setup.}  For each of the 18 diagnostic labels and 11 visual attributes, a linear layer was trained on GreenRFM and CLIP-Baseline embeddings of the full CT-RATE \emph{training} set ($n\!=\!47{,}149$) and evaluated on the \emph{validation} set ($n\!=\!3{,}039$). The results are summarized in Table~\ref{tab:attribute_probing} and Table~\ref{tab:attr_probe_perlabel}.

\begin{table}[htbp]
\centering
\caption{\textbf{Attribute-probing summary (train$\to$val, full training set).}  Mean AUC across diagnostic labels and visual attributes for GreenRFM and the CLIP-Baseline.  The attribute AUC gain ($+0.062$) exceeds the diagnostic AUC gain ($+0.046$). We report the attribute mean excluding \texttt{density\_mixed}, as it has too few positive validation cases.}
\label{tab:attribute_probing}
\suppTableFont\suppTableSetup
\begin{tabularx}{\linewidth}{@{}>{\raggedright\arraybackslash}p{0.22\linewidth}>{\raggedright\arraybackslash}Xrrr@{}}
\toprule
Analysis & Metric & GreenRFM & CLIP-Baseline & $\Delta$ \\
\midrule
\emph{Linear probe} & Diagnostic AUC (mean, 18 labels)        & \textbf{0.861} & 0.815 & $+0.046$ \\
 & Attribute AUC (excl.\ \texttt{density\_mixed}) & \textbf{0.757} & 0.696 & $+0.061$ \\
\addlinespace[4pt]
\emph{CKA} 
&  CKA(embed, diagnostic labels)  & \textbf{0.456} & 0.406 & $+0.050$ \\
& CKA(embed, visual attributes)  & 0.162 & \textbf{0.168} & $-0.006$ \\
& Attribute/diagnostic ratio  & 0.355 & \textbf{0.414} & --- \\
\addlinespace[4pt]
\emph{MI} 
& MI(embed; diagnostic labels) [nats]  & \textbf{0.163} & 0.147 & $+0.016$ \\
& MI(embed; visual attributes) [nats]  & \textbf{0.079} & 0.078 & $+0.001$ \\
\addlinespace[4pt]
\emph{Reference} & CKA(random-feature, diagnostic labels) & \multicolumn{3}{c@{}}{0.040} \\
& CKA(random-feature, visual attributes) & \multicolumn{3}{c@{}}{0.027} \\
\bottomrule
\end{tabularx}
\end{table}

\begin{table}[htbp]
\centering
\caption{\textbf{Per-attribute visual AUC from linear probing (train$\to$val).}}  
\label{tab:attr_probe_perlabel}
\suppTableFont\suppTableSetup
\begin{tabular}{@{}lrrrr@{}}
\toprule
Attribute & $n_{\text{pos}}$ & GreenRFM AUC & CLIP-Baseline AUC & $\Delta$ \\
\midrule
morphology\_patchy              & 413  & \textbf{0.905} & 0.880 & $+0.025$ \\
distribution\_diffuse           & 1113 & \textbf{0.866} & 0.819 & $+0.047$ \\
distribution\_unilateral        & 67   & \textbf{0.811} & 0.685 & $+0.126$ \\
morphology\_reticular           & 220  & \textbf{0.774} & 0.704 & $+0.070$ \\
distribution\_bilateral\_symmetric & 1588 & \textbf{0.751} & 0.708 & $+0.043$ \\
morphology\_nodular             & 1509 & \textbf{0.727} & 0.651 & $+0.076$ \\
morphology\_linear              & 493  & \textbf{0.711} & 0.622 & $+0.089$ \\
distribution\_focal             & 1160 & \textbf{0.700} & 0.605 & $+0.095$ \\
density\_low                    & 309  & \textbf{0.672} & 0.654 & $+0.018$ \\
density\_high                   & 199  & \textbf{0.653} & 0.632 & $+0.021$ \\
\midrule
\textbf{Mean (excl.\ density\_mixed)} & & \textbf{0.757} & 0.696 & $+0.061$ \\
\bottomrule
\end{tabular}
\end{table}

\FloatBarrier
\subsection{Translational downstream: methodological detail}
\label{supp:downstream_detail}

\FloatBarrier
\subsubsection{HCC MVI clinical text input format.}\ For the HCC MVI analysis, the text encoder receives a structured natural-language string constructed per patient from the clinical data table. The string encodes: (1) age and sex; (2) key laboratory values (albumin, total bilirubin, ALT, AST, PT-INR); and (3) hepatic function indices (ALBI grade, FIB-4 index). Representative examples are shown below:

\begin{quote}
\suppWideTableFont\ttfamily\suppTableSetup
\noindent\hspace*{-0.05\linewidth}%
\begin{tabularx}{\linewidth}{@{}L{0.03\linewidth} >{\raggedright\arraybackslash}X@{}}
\toprule
\textbf{Case} & \textbf{Clinical text} \\
\midrule
1 & 52-year-old male patient. Lab values: albumin 44.0\,g/L, total bilirubin 13.3\,\textmu mol/L, ALT 29\,U/L, AST 30\,U/L, PT-INR 1.07. ALBI grade 1, FIB-4 index 1.35. \\
\addlinespace[3pt]
2 & 32-year-old female patient. Lab values: albumin 38.0\,g/L, total bilirubin 95.8\,\textmu mol/L, ALT 197\,U/L, AST 185\,U/L, PT-INR 0.92. ALBI grade 2, FIB-4 index 2.72. \\
\addlinespace[3pt]
3 & 77-year-old male patient. Lab values: albumin 46.9\,g/L, total bilirubin 21.9\,\textmu mol/L, ALT 31\,U/L, AST 31\,U/L, PT-INR 1.01. ALBI grade 1, FIB-4 index 13.26. \\
\addlinespace[3pt]
4 & 51-year-old male patient. Lab values: albumin 43.2\,g/L, total bilirubin 15.1\,\textmu mol/L, ALT 15\,U/L, AST 18\,U/L, PT-INR 1.01. ALBI grade 1, FIB-4 index 0.86. \\
\bottomrule
\end{tabularx}
\end{quote}

\noindent Each string is tokenised with \texttt{BertTokenizer} and fed into CXR-BERT-specialized; the \texttt{[CLS]} hidden state (768\,d) is projected by a linear layer and concatenated with the image embedding before the classification head.

\FloatBarrier
\subsubsection{WAW-TACE survival modelling detail.}\ Cox proportional-hazards modelling was performed on the held-out test set ($n\!=\!92$; PFS subset $n\!=\!43$). GreenRFM features were projected to $20$ components via PCA before entering the Cox model. GreenRFM feature C-indices were $0.717$ (OS) and $0.758$ (PFS), both exceeding all four clinical scores (HAP, mHAP-2, ALBI-TAE, 6-and-12). Adding the numeric 6-and-12 score directly to the Cox model raised C-indices to $0.754$ (OS, $95\%$ CI $[0.698, 0.806]$) and $0.815$ (PFS, $95\%$ CI $[0.719, 0.899]$). The PFS bootstrap CI was non-overlapping against every individual clinical baseline; for OS, the CI was non-overlapping against HAP, ALBI-TAE, and 6-and-12, with only a slight overlap against mHAP-2(0.003). Full C-index results for all GreenRFM combinations are reported in Table~\ref{tab:tace_cox_all}.

\begin{table}[htbp]
    \centering
    
    \caption{\textbf{WAW-TACE Cox proportional-hazards C-index for all GreenRFM combinations (OS and PFS).}
    \textbf{Top panel:} Overall survival (OS, $n\!=\!92$); strongest clinical baseline is mHAP-2 ($C\!=\!0.630$).
    \textbf{Bottom panel:} Progression-free survival (PFS, $n\!=\!43$); strongest clinical baseline is 6-and-12 ($C\!=\!0.559$).
    $p$-values are from one-sided permutation tests (10{,}000 permutations) against the respective strongest clinical baseline.
    All GreenRFM combinations are higher than all four clinical baselines in these one-sided permutation tests ($p\!<\!0.05$).
    Significance codes: $*$ $p\!<\!0.05$; $**$ $p\!<\!0.01$; $***$ $p\!<\!0.001$.}
    \label{tab:tace_cox_all}
    \suppTableFont\suppTableSetup
    \begin{tabular}{@{}llrr@{}}
    \toprule
    Endpoint & Model & C-index & $p$ vs.\ strongest baseline \\
    \midrule
    \multirow{5}{*}{\shortstack[l]{OS ($n\!=\!92$)\\baseline: mHAP-2\\$C\!=\!0.630$}}
      & GreenRFM (PCA)           & $0.717$ & $0.045$ $*$  \\
      & GreenRFM $+$ HAP         & $0.723$ & $0.016$ $*$  \\
      & GreenRFM $+$ mHAP-2      & $0.726$ & $0.007$ $**$ \\
      & GreenRFM $+$ ALBI-TAE    & $0.726$ & $0.010$ $*$  \\
      & GreenRFM $+$ 6-and-12    & $\mathbf{0.754}$ & $0.005$ $**$ \\
    \addlinespace[4pt]
    \multirow{5}{*}{\shortstack[l]{PFS ($n\!=\!43$)\\baseline: 6-and-12\\$C\!=\!0.559$}}
      & GreenRFM (PCA)           & $0.758$ & $0.016$ $*$   \\
      & GreenRFM $+$ HAP         & $0.764$ & $0.013$ $*$   \\
      & GreenRFM $+$ ALBI-TAE    & $0.764$ & $0.016$ $*$   \\
      & GreenRFM $+$ mHAP-2      & $0.760$ & $0.012$ $*$   \\
      & GreenRFM $+$ 6-and-12    & $\mathbf{0.815}$ & $0.001$ $***$ \\
    \bottomrule
    \end{tabular}
\end{table}

\FloatBarrier

\textbf{WAW-TACE clinical text input format.}\ For the TACE-response prediction analysis, the text encoder receives a structured string encoding tumour imaging characteristics and TACE-specific prognostic scores. Each string encodes: (1) diagnostic context (HCC scheduled for first TACE); (2) age and sex; (3) hepatic aetiology (e.g., HCV, alcoholic cirrhosis); (4) laboratory values (albumin, bilirubin, AFP, INR, ALT, creatinine); (5) tumour imaging findings (lesion count, largest lesion size and lobe, LI-RADS grade); (6) clinical staging (BCLC stage, Child-Pugh class); and (7) TACE-specific prognostic scores (HAP score, mHAP-II score). Representative examples:

\begin{quote}
\suppWideTableFont\ttfamily\suppTableSetup
\noindent\hspace*{-0.05\linewidth}%
\begin{tabularx}{1.12\linewidth}{@{}L{0.03\linewidth} >{\raggedright\arraybackslash}X@{}}
\toprule
\textbf{Case} & \textbf{Clinical text} \\
\midrule
2 & HCC patient scheduled for first TACE treatment. 63-year-old male. Etiology: HCV. Lab values: albumin 3.8\,g/L, bilirubin 2.5\,mg/dL, AFP 28.7\,ng/mL, INR 1.50, ALT 61\,IU/L, creatinine 0.7\,mg/dL. 2 lesion(s), largest 27\,mm in left lobe, LI-RADS~5. BCLC stage B, Child-Pugh class A. HAP score 1, mHAP-II 2. \\
\addlinespace[3pt]
3 & HCC patient scheduled for first TACE treatment. 76-year-old female. Etiology: HCV. Lab values: albumin 3.2\,g/L, bilirubin 1.2\,mg/dL, AFP 31300.0\,ng/mL, INR 1.09, ALT 42\,IU/L, creatinine 0.9\,mg/dL. 1 lesion(s), largest 78\,mm in right lobe, LI-RADS~5. BCLC stage A, Child-Pugh class A. HAP score 4, mHAP-II 4. \\
\addlinespace[3pt]
5 & HCC patient scheduled for first TACE treatment. 80-year-old female. Etiology: HCV. Lab values: albumin 3.7\,g/L, bilirubin 1.2\,mg/dL, AFP 31300.0\,ng/mL, INR 1.12, ALT 46\,IU/L, creatinine 0.8\,mg/dL. 1 lesion(s), largest 45\,mm in left lobe, LI-RADS~5. BCLC stage A, Child-Pugh class A. HAP score 2, mHAP-II 2. \\
\addlinespace[3pt]
7 & HCC patient scheduled for first TACE treatment. 58-year-old male. Etiology: alcoholic cirrhosis. Lab values: albumin 4.2\,g/L, bilirubin 0.7\,mg/dL, AFP 5.2\,ng/mL, INR 1.14, ALT 31\,IU/L, creatinine 0.5\,mg/dL. 3 lesion(s), largest 40\,mm in left lobe, LI-RADS~5. BCLC stage B, Child-Pugh class A. HAP score 0, mHAP-II 1. \\
\bottomrule
\end{tabularx}
\end{quote}

The same BertTokenizer $\to$ CXR-BERT $\to$ \texttt{[CLS]} projection pipeline is used for text-based TACE-response models. In the survival Cox analysis above, clinical scores such as 6-and-12 are used as numeric covariates rather than being passed through the text encoder.

\FloatBarrier
\subsection{Additional supervision-control analyses}
\label{supp:additional_supervision_controls}

\subsubsection{Hierarchical visual-description supervision}
The main ablation uses an 11-attribute visual-description target (density, morphology, and distribution; Supplementary Appendix~\ref{supp:visual_description_prompt}). To test whether this control was too coarse, we also evaluated a stronger hierarchical description target. For each of the 18 CT-RATE findings, extracted descriptors were assigned to anatomical location (9 categories), morphology or appearance (12), severity or extent (4), and distribution or laterality (7), producing $18{\times}32=576$ attribute targets. A second variant added an explicit 18-dimensional diagnostic-category branch, yielding 594 targets. All variants used the same CT-RATE train/validation split, R3D-18 image encoder, CXR-BERT text encoder, and two-stage training protocol as the main ablation.

A manual check of 100 sampled attribute extractions gave a macro precision-proxy AUC of $0.8633$ across audited attributes, with per-attribute values ranging from $0.7128$ to $0.9909$. The hierarchical attribute-only target reached AUC $0.8121$, similar to the original 11-attribute visual-description control. Adding explicit diagnostic-category identity raised AUC to $0.8352$, but remained below diagnostic-label supervision ($0.848$; Table~\ref{tab:hierarchical_description}). These results suggest that descriptive attributes alone did not close the gap to diagnosis-aligned supervision for the diagnosis-centered task studied here.

\begin{table}[htbp]
\centering
\caption{\textbf{Stronger hierarchical visual-description supervision.} AUC is the CT-RATE validation zero-shot macro AUC for each hierarchical-description variant. GreenRFM-L denotes the lightweight model variant.}
\label{tab:hierarchical_description}
\suppTableFont\suppTableSetup
\begin{tabular}{@{}L{4.0cm}cL{4.0cm}c@{}}
\toprule
Variant & Labels & Model variant & Zero-shot AUC \\
\midrule
Attribute-only & 576 & GreenRFM & 0.8121 \\
Attribute-only & 576 & GreenRFM-L & 0.8002 \\
Attribute + diagnostic category & 594 & GreenRFM & \textbf{0.8352} \\
Attribute + diagnostic category & 594 & GreenRFM-L & 0.8135 \\
\bottomrule
\end{tabular}
\end{table}

\FloatBarrier
\subsubsection{Stage-2 partial-freeze sensitivity}
We also evaluated supervised-first partial-freeze variants during Stage~2 alignment. The full GreenRFM alignment stage updates both encoders and the alignment heads. In the sensitivity analysis, the image encoder, the text encoder, or both encoders were frozen during Stage~2, leaving the remaining trainable components to perform alignment. All variants used the same Stage~1 checkpoint, training data, and zero-shot CT-RATE evaluation protocol.

Freezing either encoder reduced AUC modestly relative to the fully trainable Stage~2 model, and freezing both encoders while training only the two linear alignment heads gave AUC $83.5\%$ (Table~\ref{tab:partial_freeze_stage2}). This pattern supports the ordering used by GreenRFM: Stage~1 establishes useful unimodal diagnostic structure, and Stage~2 benefits from allowing that structure to adapt during final alignment.

\begin{table}[htbp]
\centering
\caption{\textbf{Stage-2 partial-freeze sensitivity.} CT-RATE zero-shot AUC for supervised-first variants in which one or both encoders are frozen during Stage~2 alignment.}
\label{tab:partial_freeze_stage2}
\suppTableFont\suppTableSetup
\begin{tabular}{@{}L{5.4cm}L{5.4cm}c@{}}
\toprule
Stage~2 setting & Trainable components & CT-RATE AUC \\
\midrule
Full GreenRFM & Image encoder, text encoder, projection heads, shared classifier & \textbf{84.8\%} \\
Freeze image encoder & Text encoder, projection heads, shared classifier & 84.3\% \\
Freeze text encoder & Image encoder, projection heads, shared classifier & 83.5\% \\
Freeze both encoders & Projection heads and shared classifier only & 83.5\% \\
\bottomrule
\end{tabular}
\end{table}

\end{document}